%% file: LMASE-AAAI.tex
\def\year{2022}\relax
\documentclass[letterpaper]{article} 
\usepackage{aaai22}  
\usepackage{times}  
\usepackage{helvet}  
\usepackage{courier}  
\usepackage[hyphens]{url}  
\usepackage{graphicx} 
\usepackage{natbib}  
\usepackage{caption} 
\DeclareCaptionStyle{ruled}{labelfont=normalfont,labelsep=colon,strut=off} 
\usepackage{algorithm}
\usepackage{algorithmic}

\usepackage{newfloat}
\usepackage{listings}

\usepackage{latexsym}
\usepackage{booktabs}
\usepackage{amsmath}
\usepackage{xcolor}
\usepackage{adjustbox}
\usepackage{caption}
\usepackage{longtable}

\floatstyle{ruled}
\newfloat{listing}{tb}{lst}{}
\floatname{listing}{Listing}
\newcommand{\cb}[2]{\colorbox{blue!#1}{\vphantom{Q}#2}}
\input{tables}
\input{doc_viz_whales}
\input{doc_viz_elvis}
\input{doc_viz_cycling}
\input{doc_viz_italy}

\newdimen\origiwspc%
\origiwspc=\fontdimen2\font

\title{Play the Shannon Game With Language Models: \\ A Human-Free Approach to Summary Evaluation}
\author {
    Nicholas Egan,
    Oleg Vasilyev,
    John Bohannon
}
\affiliations {
    Primer AI\\
    \{negan, oleg, bohannon\}@primer.ai
}

\begin{document}

\maketitle

\begin{abstract}
The goal of a summary is to concisely state the most important information in a document. With this principle in mind, we introduce new reference-free summary evaluation metrics that use a pretrained language model to estimate the information content shared between a document and its summary. These metrics are a modern take on the Shannon Game, a method for summary quality scoring proposed decades ago, where we replace human annotators with language models. We also view these metrics as an extension of BLANC, a recently proposed approach to summary quality measurement based on the performance of a language model with and without the help of a summary. Using transformer based language models, we empirically verify that our metrics achieve state-of-the-art correlation with human judgement of the summary quality dimensions of both coherence and relevance, as well as competitive correlation with human judgement of consistency and fluency.
\end{abstract}

\section{Introduction}
With the ever-expanding development of new summarization algorithms in the NLP community, metrics that reliably measure summary quality are more important than ever. And yet, the most popular method for summary quality estimation remains the ROUGE \citep{lin-2004-rouge} family of metrics, which require human written reference summaries for comparison and measure summary quality through simple token overlap, ignoring the syntax and semantics governing the way humans use language.

The goal of a summary is to concisely state the most important information conveyed by a document. Examining summarization through this lens, one should be able to determine summary quality by measuring how much information from the document is represented in the summary. Put another way, when comparing alternative summaries of similar length, the information we gain from reading the original document should be minimal given the best summary.

The idea of measuring this difference in information content was proposed as the Shannon Game by \citet{hovy-lin-1998-automated}: they assign 3 humans the task of guessing a document letter by letter, where the first human is allowed to look at the document, the second human is allowed to look at a summary of the document, and the third human is given nothing at all. By measuring how many tries it takes the second human to guess the document compared to the other humans, you can evaluate how much information about the document is communicated in the summary, and therefore measure how good the summary is.

\paragraph{Contributions} This paper proposes a new summarization evaluation metric, the \textit{Shannon Score}, that performs the Shannon Game with a language model such as GPT-2 \citep{gpt2}. By using a language model to autoregressively generate a document both with and without a summary as a prompt, we measure the information provided by the summary. One can view this method as a more theoretically driven extension to the recently proposed BLANC metric \citep{blanc}, which measures the accuracy of unmasking document tokens with and without a summary. In addition to the Shannon Score, we also propose a variant we call \textit{Information Difference}.

To understand the empirical performance of this method as a summary evaluation technique, we performed experiments to correlate our metrics against human judgement. We found that our metrics perform strongly on the SummEval benchmark \citep{summeval}, achieving state-of-the-art correlation with human judgement of summary coherence and relevance, and competitive correlation with human judgement of summary consistency and fluency.

\section{Methods}
\subsection{Computing Information}

Language models are probability distributions over documents, giving us $p(\mathcal{D})$ for some document $\mathcal{D}$. Autoregressive language models do this by predicting next token probabilities given prior tokens, modeling 
\[p(x_t | x_1,\ldots,x_{t-1})\]
where our input document is broken into tokens $\{x_1,\ldots,x_n\}$.
The Shannon information content, or surprisal, of event $E$ with probability $p(E)$ of happening is defined as $ I(E) = -\log p(E) $, so we can compute the information of a document according to our language model as
\[ 
\begin{split}
I(\mathcal{D}) = &-\log p(x_1) -\log p(x_2|x_1) - \ldots \\
&- \log p(x_n | x_1, x_2, \ldots x_{n-1})
\end{split}
\]


\subsection{Conditional Information}

Suppose we had a conditional language model $p(\mathcal{D} | \mathcal{S})$ that gives us a probability distribution of documents that could correspond to a given summary $\mathcal{S}$. Using this conditional language model, we could compute the conditional information content $I(\mathcal{D}|\mathcal{S})$ as the amount of information we gain from the document $\mathcal{D}$ if we are already given the information of summary $\mathcal{S}$. 

If $\mathcal{S}$ is a satisfactory summary of $\mathcal{D}$, then $I(\mathcal{D} | \mathcal{S}) < I(\mathcal{D})$, as documents that have little to do with the summary should be much less likely than documents that are relevant to the summary after conditioning the language model. If the summary fluently describes people, ideas, or relationships that appear in the document, then that should decrease the information one learns from subsequently reading the document.

Thus we can define an \textit{Information Difference} metric of summary quality as:
\[ ID(\mathcal{D}, \mathcal{S}) = I(\mathcal{D}) - I(\mathcal{D} | \mathcal{S}) \]
The Information Difference tells us the change in document information between using the summary and not using the summary, and it is equivalent to the log likelihood ratio between the document and the document given the summary. While it is unbounded, it should be positive unless a summary does such a bad job that it makes the document more confusing to read.

Considering the fact that the summary that best preserves the information of a document is the document itself, we can view $I(\mathcal{D}|\mathcal{D})$ as a lower bound on $I(\mathcal{D}|\mathcal{S})$. Since this idea of having a third evaluator who has the document itself as help is inspired by the Shannon Game, we can compute the \textit{Shannon Score} metric as:
\[ s(\mathcal{D}, \mathcal{S}) = \frac{I(\mathcal{D}) - I(\mathcal{D}|\mathcal{S})}{I(\mathcal{D}) - I(\mathcal{D}|\mathcal{D})} \]
The Shannon Score gives us the ratio between how helpful the summary was and how helpful the document itself was. While this formula in theory is unbounded, it usually should be in the range $0$ to $1$, unless the summary makes the document more confusing or somehow explains the document better than the document itself.

\subsection{Approximating Conditional Information}
\label{section:approx-cond-info}
To the extent of our knowledge, there is no easy way to exactly condition a pretrained language model such as GPT-2 on a summary, even though there has been work on conditioning language models on fixed control codes \citep{keskarCTRL2019}, bags of words, or discriminators \citep{Dathathri2020Plug}. 
We also have a strong motivation not to train such a model because we want our method to be universal and robust, while summarization datasets are much smaller and more restricted in domain than the massive datasets that modern language models require. 

We approximate $p(\mathcal{D}|\mathcal{S})$ by computing the probability that $\mathcal{D}$ is generated when we provide $\mathcal{S}$ as a prompt to a language model. We intuitively justify this idea by the fact that in real-world documents the most important information is often summarized at the top as an introduction, and then described in more detail in body paragraphs. This setup resembles the BLANC-help metric \citep{blanc}, which measures language model token unmasking accuracy for a document when a summary is prepended. An alternative setup would be to finetune a language model on the summary which was also explored by \citet{blanc}, but we don't explore that method in this paper. We use the GPT-2 small language model \cite{gpt2} for our experiments, but investigate the use of other language models in section \ref{section:lm-choice}.

An issue we run into when computing information with GPT-2 is that the model can only be given a maximum of 1024 tokens, making many documents too large to fit in at once. To get around this, we approximate document information with an independence assumption between sentences in the document, meaning that only the preceding tokens within a sentence are provided when generating the next token in the sentence. In section \ref{section:upstream-sentences}, we investigate the effects of prompting the language model with additional upstream sentences of context.

\section{Understanding Our Metrics}

\subsection{Information Visualization}
\whalesfigure

A toy illustration of our methodology is shown in Figure \ref{fig:whales-trunc}. We picked a document excerpt in the CNN/DailyMail \cite{cnndm} dataset and paired it with two abstractive summaries we wrote. While both of these summaries are grammatically correct and mostly consist of words from the document, one of the summaries is of high quality and the other is of low quality. The figure shows the information content of each token in the document as estimated by GPT-2 in 4 scenarios: $I(\mathcal{D})$ (the document on its own), $I(\mathcal{D}|\mathcal{D})$ (the document given the document), $I(\mathcal{D}|\mathcal{S})$ (the document given a summary) for the high quality summary, and $I(\mathcal{D}|\mathcal{S})$ for the low quality summary. A darker background color denotes higher information according to the model.

As you can see, the model gained less information from words like ``gray'' and ``Varvara'' after seeing those words in the high quality summary. We can also see that words like ``Pacific'' and ``journey,'' which do not appear in the high quality summary, became more likely to appear in the document due to their association with concepts in the summary.
The low quality summary may have helped the model predict words like ``CNN,'' but it is unhelpful for words like ``mammal'' and ``website'' that are confusingly used in the summary.
Very little information was gained from reading a document that was already read, except for the first token or two for each sentence. This is an artifact of our autoregressive language modeling setup, so measuring $I(\mathcal{D}|\mathcal{D})$ is useful for normalizing our Shannon Scores.

We used a truncated document and toy summaries here to demonstrate the Shannon Score in a concise way, but we included visualizations of real, full-length documents and summaries from the SummEval dataset in the appendix.

\subsection{Baseline Validation}
\begin{figure}[h!]
\centering
\includegraphics{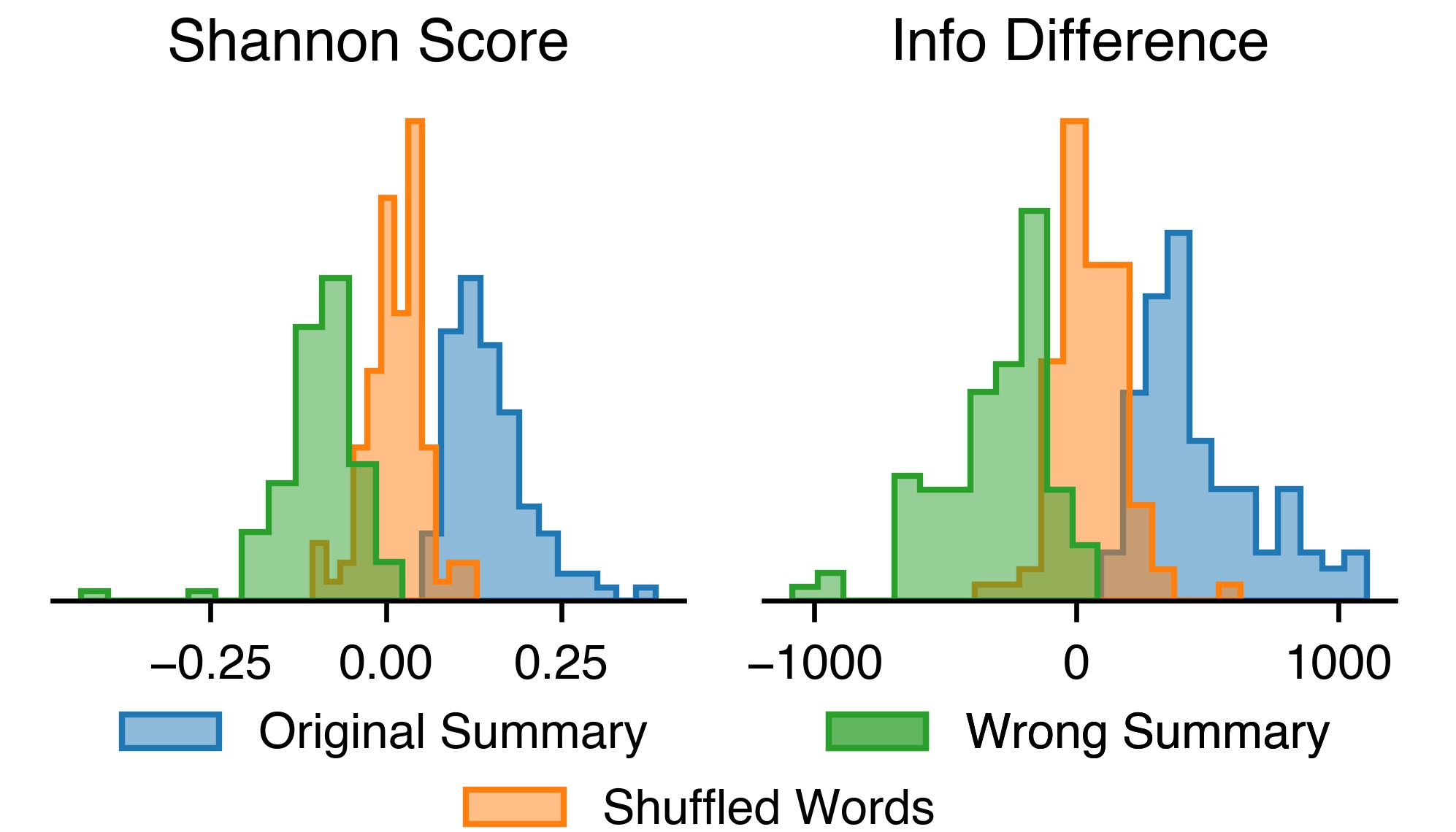}
\caption{Distributions of Shannon Score and Information Difference on 100 summaries from the CNN/DailyMail dataset. Three different summaries are used: the original human written reference summary (in blue), the original summary with words scrambled (in orange), and a reference summary for a different document in the dataset (in green).}
\label{fig:dists}
\end{figure}

As a simple validation of our information-based metrics, we sampled 100 documents with their corresponding reference summaries from the CNN/DailyMail dataset \citep{cnndm}, and created two ``bad'' summaries per document: a version of the reference summary with all the words randomly shuffled, and a reference summary for a different document in the dataset. 

Figure \ref{fig:dists} shows the distributions of the Shannon Score and Information Difference for these three summaries. As expected, the original summaries have the highest scores, followed by shuffled summaries and wrong summaries. It is good to see that there is full separation between original summaries and wrong summaries for both metrics. The fact that the original summaries and shuffled summaries are almost completely separated demonstrates the importance of syntax to our metrics, a quality that metrics like the Jensen-Shannon divergence \citep{louis-nenkova-2009-automatically} and ROUGE-1 \citep{lin-2004-rouge} lack.

We also verified that there are no documents for which the shuffled summary or wrong summary score better than the original summary for either of the metrics. 
Despite the fact that the Shannon Score has no lower bound, we can see that it doesn't go far below zero for even the most unreasonable of summaries. And despite the fact that the Shannon Score has no upper bound, even high quality human reference summaries are unable to achieve a score above 0.4.

\section{Evaluation of Our Metrics}

\subsection{SummEval}
\label{section:summeval}
\summevaltablefour

The SummEval \citep{summeval} benchmark was established as a comprehensive evaluation tool for summary evaluation metrics. It consists of 100 English-language documents from the CNN/DailyMail dataset, each paired with system summaries from 17 different summarization systems: 3 extractive models, 13 abstractive models, and a lead-3 baseline. All models were published in 2017 or later. Each of these 1700 system summaries were scored by a panel of 3 experts in the field of summarization on the qualities of coherence (the collective quality of all sentences), consistency (the factual alignment between the summary and document), fluency (the quality of individual sentences), and relevance (selection of important content from the source). The experts achieved an inter-annotator agreement kappa coefficient of 0.7127.

\citet{summeval} scored each summary using these evaluation metrics: ROUGE \cite{lin-2004-rouge}, ROUGE-WE \cite{ng-abrecht-2015-better}, $S^3$ \cite{peyrard-etal-2017-learning}, BertScore \cite{bert-score}, MoverScore \cite{zhao-etal-2019-moverscore}, Sentence Mover's Similarity (SMS) \cite{clark-etal-2019-sentence}, SummaQA \cite{scialom-etal-2019-answers}, BLANC \cite{blanc}, SUPERT \cite{gao-etal-2020-supert}, BLEU \cite{bleu}, CHRF \cite{popovic-2015-chrf}, METEOR \cite{lavie-agarwal-2007-meteor}, and CIDEr \cite{vedantam2015cider}. They also measure the \citet{grusky-etal-2018-newsroom} statistics of summary length, extractive fragment coverage (coverage), compression ratio, average length of extractive fragments (density), proportion of $n$-grams in summary that aren't in the document (novel $n$), and $n$-grams repeated in summary (repeat $n$).

Table \ref{tab:summeval4} shows the correlation between expert annotations and the automated evaluation metrics. Following \citet{summeval}, we use Kendall tau-b system-level correlation for comparison. Our metrics of Shannon Score and Information Difference are the only metrics to be in the top 5 for every category of summary quality. Additionally, our metrics achieve state-of-the-art performance for the qualities of coherence and relevance.

\subsection{Coverage}
\ductablesystems

The coverage score \citep{lin-hovy-2003-automatic} is a human evaluation method that measures a system summary's recall of semantic units that appeared in a reference summary, weighed by how well the system summary was able to capture each semantic unit as judged by the human labeler. The 2001 and 2002 Document Understanding Conferences (DUC) provide datasets of English-language system and reference summaries for news documents with human coverage labels, on both single-document and multi-document levels. 

Table \ref{tab:coverage-systems} shows the correlation of various metrics to these coverage scores for the single-document summaries. System-level Spearman correlation is used following \citet{louis-nenkova-2013-automatically}. The reference-free metrics perform similarly, except for Jensen-Shannon Divergence \cite{louis-nenkova-2009-automatically} which performs particularly well on DUC 2001 and Info Diff which performs particularly poorly on DUC 2002. The metrics using references benefit from the bias that the coverage itself was measured with respect to the reference summary, so as expected, they have higher correlations with the coverage than the reference-free metrics for this dataset. A fair comparison would involve a coverage measured with respect to the document itself. One can also see that most metrics perform better on DUC 2002 than DUC 2001: this was  also observed by \citet{sun-nenkova-2019-feasibility}, who suggested that this can be explained by the fact that DUC 2001 systems are more similar to each other and worse than DUC 2002 systems on average.

\subsection{Metric Biases}
\biastable

To understand the biases of our metrics, we measured the correlation between our metrics and the SummEval statistics describing summaries described in section \ref{section:summeval} across the 1700 SummEval summaries. For comparison, we also correlated the expert summary quality judgements with the statistics. These correlations are shown in table \ref{tab:biases}.

Both of our metrics have significant positive correlation with summary length, which is expected since longer summaries can contain more information. Our metrics have bias against more abstractive summaries (based on novel $n$-gram, coverage, and density), but we are generally less biased against abstractive summaries than humans judging consistency are: we suspect this is because abstractive summaries are more likely to hallucinate factual errors. The Shannon Score is biased against highly compressed summaries, which is not shared by Information Difference.

\section{Metric Variations}

\subsection{Choice of Language Model or Model Size}
\label{section:lm-choice}

\begin{figure}[ht!]
\begin{minipage}{\linewidth}
\lmcomparisontable
\vskip 10mm
\centering
\includegraphics{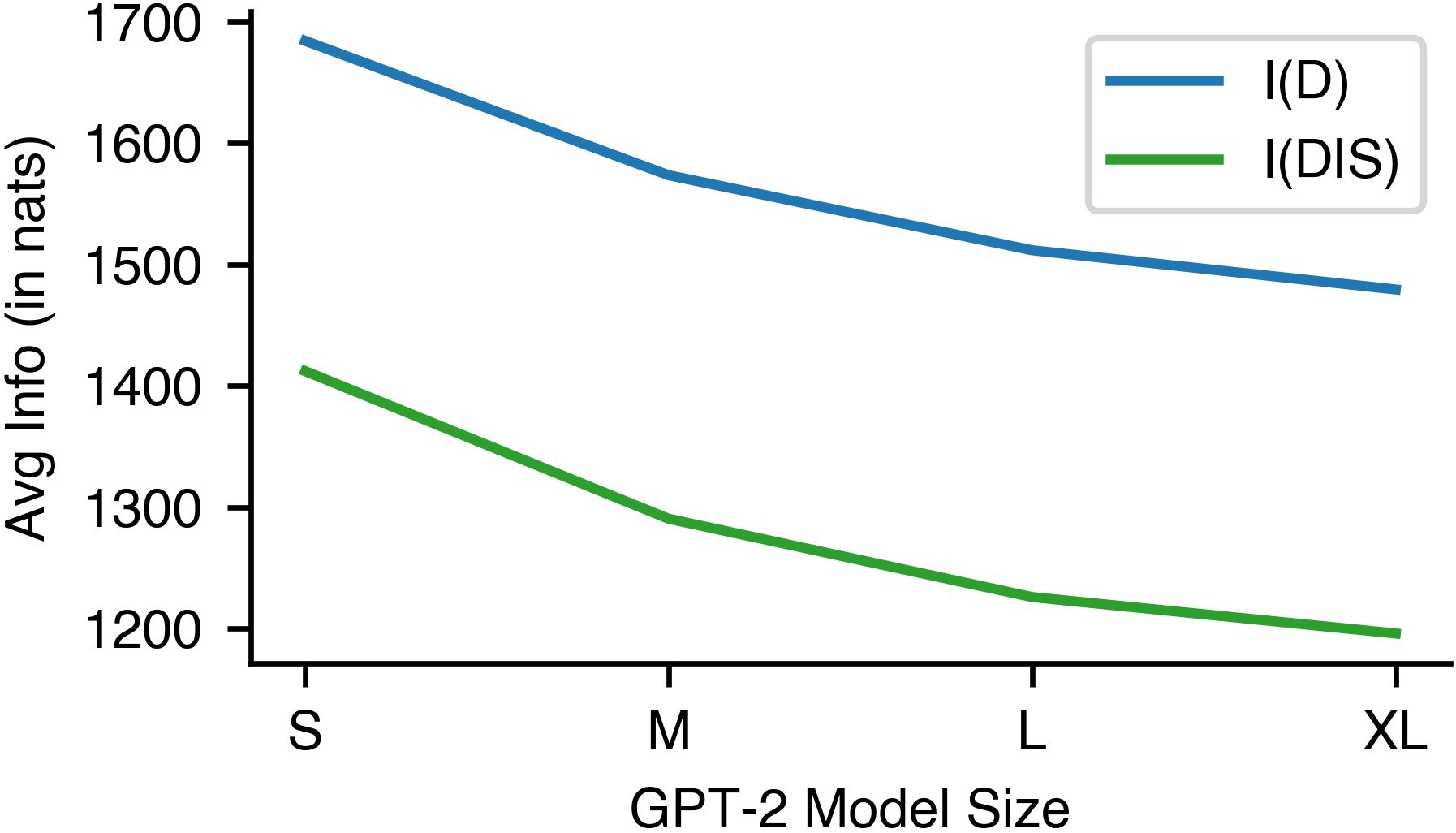}
\captionof{figure}{The average document information and document information given summary as estimated by different sizes of GPT-2 for the SummEval dataset.}
\label{fig:size-info}
\end{minipage}
\end{figure}

In the previous sections, we used GPT-2 small as our language model of choice when computing the Shannon Score and Information Difference. To understand how well our method generalizes to other language models, we computed the Shannon Score and Information Difference metrics using the three other GPT-2 sizes (medium, large, and extra-large), and three other language models with autoregressive pretraining objectives: GPT \citep{gpt1}, XLNet \citep{xlnet}, and Transformer-XL \cite{dai-etal-2019-transformer}.

Table \ref{tab:lms-comparison} shows the system-level Kendall tau-b correlation between our metrics and the SummEval quality judgements from section \ref{section:summeval} for each language model. The language models perform quite similarly overall, suggesting that the choice of language model is not overly important when using the Shannon Score or Information Difference. The exception is the low correlation of GPT, particularly on the coherence and relevance qualities: we suspect this is because GPT was trained on the BooksCorpus dataset \citep{Zhu2015AligningBA}, which is less diverse than the datasets used for the other language models.

It is also interesting to see that bigger GPT-2 models do not necessarily perform better. Figure \ref{fig:size-info} shows the relationship between model size and average document info with and without the help of a summary. We can see that as the model gets larger, both average $I(\mathcal{D})$ and average $I(\mathcal{D}|\mathcal{S})$ decrease together. Larger models should be better at autoregressive token prediction, as reflected in the plot of $I(\mathcal{D})$, but it is interesting to see that $I(\mathcal{D}|\mathcal{S})$ decreases at around the same rate. We suspect this is because larger models may not be more suitable at utilizing a summary to predict a document under our setup.

\subsection{Upstream Sentences}
\label{section:upstream-sentences}

\begin{figure}[ht!]
\begin{minipage}{\linewidth}
\upstreamtable
\vskip 10mm
\centering
\includegraphics{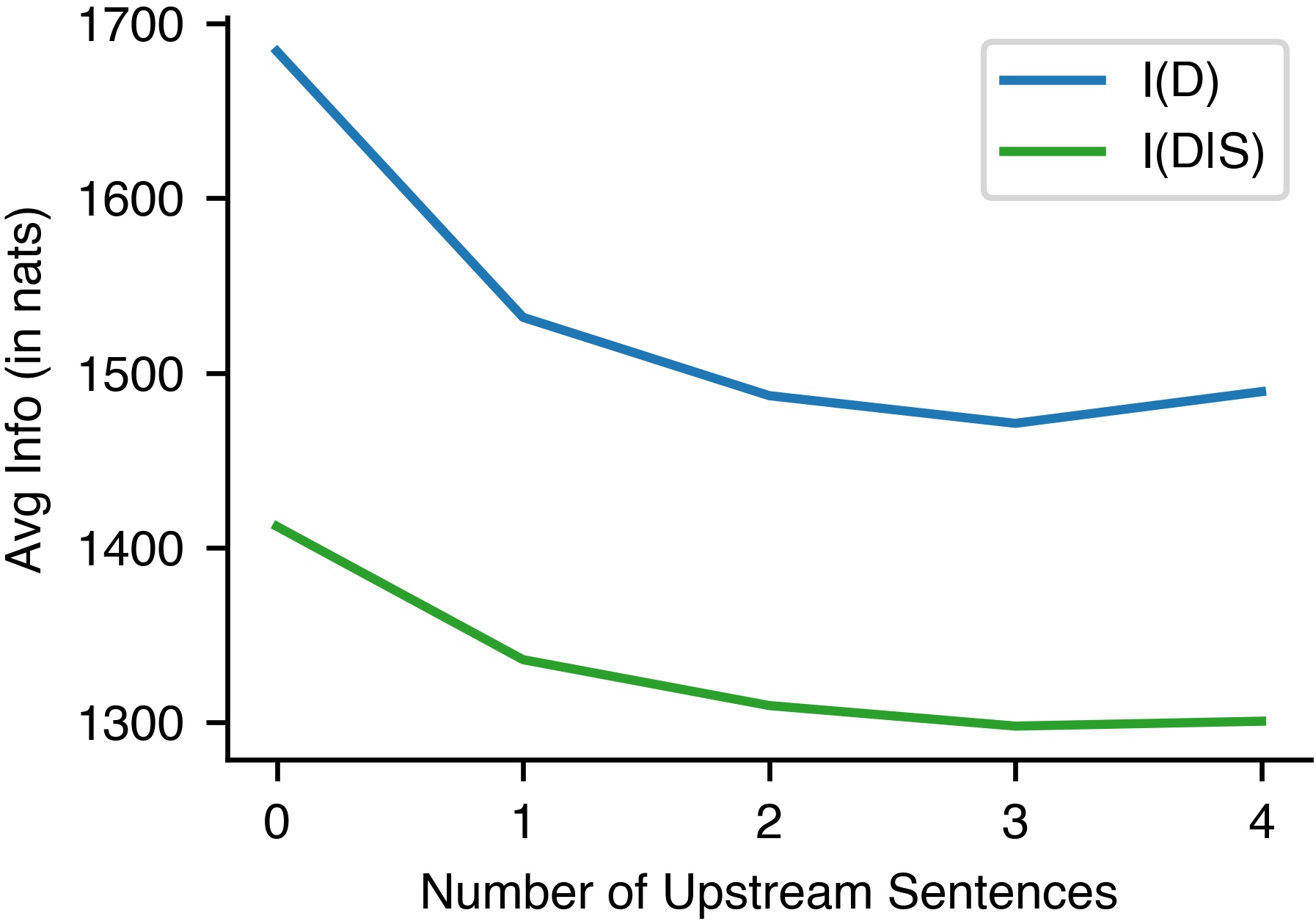}
\captionof{figure}{The average document information and document information given summary when prompting GPT-2 with different amounts of upstream sentences for the SummEval dataset.}
\label{fig:upstream-info}
\end{minipage}
\end{figure}

As described in section \ref{section:approx-cond-info}, we are making an independence assumption between sentences in a document when estimating $I(\mathcal{D})$, $I(\mathcal{D}|\mathcal{S})$, and $I(\mathcal{D}|\mathcal{D})$ by feeding each sentence into the model individually. 
We could alternatively assume that each sentence in the document is dependent on the $k$ previous sentences, where $k=0$ refers to our current approach and $k=\infty$ (or the maximum number of sentences in a document) drops the sentence independence assumption altogether. One could reason that this would better allow us to quantify the information in a document, which may lead to a more effective metric.

As shown in table \ref{tab:upstream}, using $k>0$ leads to an improvement in consistency at the expense of the other summary dimensions, and increasing $k$ beyond 1 does not yield any significant gains in performance. Figure \ref{fig:upstream-info} shows that increasing $k$ from 0 is more helpful at decreasing $I(\mathcal{D})$ than it is at decreasing $I(\mathcal{D}|\mathcal{S})$. We could draw a similar conclusion as we did in section \ref{section:lm-choice} that increasing $k$ is helpful for autoregressive token prediction, but it doesn't help our model with utilizing a summary to predict a document in our setup. 

\subsection{BLANC-Shannon}
Our metrics bear similarity to the BLANC-help metric \cite{blanc, vasilyev2020sensitivity}, which measures the accuracy of the BERT language model on the task of guessing masked tokens with and without a summary prepended to a document. The BLANC score is measured as a boost in unmasking accuracy $ a_{help} - a_{base} $ when masking various sets of $M$ evenly spaced tokens, where $a_{help}$ is the accuracy when the summary is provided as help and $a_{base}$ is the accuracy when no help is provided. Our metrics differ from BLANC in that we measure information instead of raw accuracy, we generate documents autoregressively instead of masking, and we typically use GPT-2 instead of BERT.

To study the utility of measuring document information as opposed to raw accuracy counts, we define BLANC-Shannon to be the boost in accuracy when generating document tokens given the summary. On the SummEval benchmark, BLANC-Shannon achieves Kendall tau-b system-level correlations of 0.3676, 0.6765, 0.5092, and 0.5588 for the expert annotations of coherence, consistency, fluency, and relevance respectively. These scores are an improvement on the consistency dimension over the Shannon Score and Information Difference metrics at the expense of every other dimension. We can only hypothesize that accuracy may be more sensitive to wrongly generated tokens and hence to consistency, but it would be interesting to compare BLANC-Shannon to the other metrics on an even larger dataset than SummEval.

\section{Related Work}
\paragraph{The Shannon Game}
The Shannon Game \citep{hovy-lin-1998-automated} was proposed over two decades ago as a way to use humans to measure the information retention between document and summary. In the original formulation, humans need to guess a document letter by letter given the summary, document, or nothing, and they measure the total number of guesses that were required to reconstruct the document. The authors ran a small-scale experiment where they conducted this game using human subjects, and they found a clear order of magnitude difference between the number of guesses each human required, as expected. However, they also found that reconstructing the original document with no help (the task of human 3) was extremely time-consuming, sometimes taking over 3 hours, making the Shannon Game prohibitively expensive as a human evaluation method.

\paragraph{Automated Summary Evaluation}
The most popular automatic summarization evaluation method is the ROUGE family of metrics \citep{lin-2004-rouge, lin-och-2004-automatic}, which measure word overlap between the system summary and one or more reference summaries. The two biggest problems we see with ROUGE as a metric are 1) that it relies on human written reference summaries, and 2) that it measures simple word overlap, which means that a perfectly paraphrased version of the reference summary would score poorly.

Many solutions have been proposed to remedy issue \#2 without solving issue \#1, such as BERTScore \citep{bert-score}, MoverScore \citep{zhao-etal-2019-moverscore}, Sentence Mover Similarity \citep{clark-etal-2019-sentence}, Word Mover Similarity \citep{pmlr-v37-kusnerb15}, and ROUGE-WE \citep{ng-abrecht-2015-better}. All of these metrics involve the idea of using soft overlap or embedding/token distance between the system and reference summaries. \citet{louis-nenkova-2009-automatically} suggested measuring the Jensen-Shannon divergence between word distributions used in the system summary and original document, which suffers from issue \#2 while fixing issue \#1.  \citet{sun-nenkova-2019-feasibility} and \citet{gao-etal-2020-supert} perform reference-free summary evaluation using language model word embeddings with promising results. Other have used question generation and question answering models to evaluate summaries \citep{scialom-etal-2019-answers, Ping2018SemanticQA}, but we argue that these metrics are only as good as the datasets the models were trained on, and may have problems generalizing. Beyond summarization, there have been many metrics proposed for Natural Language Generation more generally \citep{sai2020survey}.

Our methods are most similar to BLANC \cite{blanc, vasilyev2020sensitivity}, which measures the accuracy boost of BERT \citep{devlin-etal-2019-bert} on the Cloze task \citep{Wilson1953Cloze} when a summary is prepended to a document or the model is finetuned on the summary. 
This paper contributes to the study of BLANC-like metrics by extending them to new language models, giving them a theoretical motivation, and performing more robust experiments to better understand their behavior.
The information-theoretic motivations of our metrics are similar to that of \citet{peyrard-2019-simple} who formally defined some metrics based on distributions of semantic units, which contrasts with our use of pretrained language models.

\section{Conclusion}
In this work, we successfully show that a universal language model performing the basic language modeling task is an effective reference-free evaluator of summary quality. This work extends the Shannon Game from using humans as evaluators to using machines, and extends the work on BLANC-like metrics to new language models and theoretical interpretations. 
We experimentally showed that our metrics strongly correlate with expert judgement of summary quality, and hope that they will serve as useful tools for the future development of summarization models.
As next steps, it would be interesting to see if our metrics are useful for summarization model training, or evaluation in tasks beyond standard summarization, such as paraphrasing or query-focused summarization. Our code is available on GitHub.\footnote{github.com/primerai/blanc/tree/master/shannon}

\bibliography{custom,anthology}

\appendix
\include{appendix}

\end{document}

%% file: tables.tex
\newcommand{\summevaltablefour}{\begin{table}[h!]
\centering
\tabcolsep=0.11cm
\begin{tabular}{@{}lcccc@{}}
\toprule
\textbf{Metric} & \textbf{Coher.} & \textbf{Consi.} & \textbf{Fluen.} & \textbf{Relev.} \\ \midrule
Shannon\^ & \textbf{0.4118} & \textbf{0.6324} & \textbf{0.5240} & \textbf{0.6029} \\
Info Diff\^ & \textbf{0.4706} & \textbf{0.6324} & \textbf{0.5683} & \textbf{0.6618} \\ \midrule
rouge-1 & 0.2500 & 0.5294 & \textbf{0.5240} & 0.4118 \\
rouge-2 & 0.1618 & 0.5882 & 0.4797 & 0.2941 \\
rouge-3 & 0.2206 & \textbf{0.7059} & 0.5092 & 0.3529 \\
rouge-4 & 0.3088 & 0.5882 & \textbf{0.5535} & 0.4118 \\
rouge-L & 0.0735 & 0.1471 & 0.2583 & 0.2353 \\
rouge-su* & 0.1912 & 0.2941 & 0.4354 & 0.3235 \\
rouge-w & 0.0000 & 0.3971 & 0.3764 & 0.1618 \\
rouge-we-1 & 0.2647 & 0.4559 & 0.5092 & \textbf{0.4265} \\
rouge-we-2 & -0.0147 & 0.5000 & 0.3026 & 0.1176 \\
rouge-we-3 & 0.0294 & 0.3676 & 0.3026 & 0.1912 \\
$S^3$-pyr & -0.0294 & 0.5147 & 0.3173 & 0.1324 \\
$S^3$-resp & -0.0147 & 0.5000 & 0.3321 & 0.1471 \\
BertScore-p & 0.0588 & -0.1912 & 0.0074 & 0.1618 \\
BertScore-r & 0.1471 & \textbf{0.6618} & 0.4945 & 0.3088 \\
BertScore-f & 0.2059 & 0.0441 & 0.2435 & \textbf{0.4265} \\
MoverScore & 0.1912 & -0.0294 & 0.2583 & 0.2941 \\
SMS & 0.1618 & 0.5588 & 0.3616 & 0.2353 \\
SummaQA\^ & 0.1176 & 0.6029 & 0.4059 & 0.2206 \\
BLANC\^ & 0.0735 & 0.5588 & 0.3616 & 0.2647 \\
SuPERT\^ & 0.1029 & 0.5882 & 0.4207 & 0.2353 \\
BLEU & 0.1176 & 0.0735 & 0.3321 & 0.2206 \\
CHRF & \textbf{0.3971} & 0.5294 & 0.4649 & \textbf{0.5882} \\
CIDEr & 0.1176 & -0.1912 & -0.0221 & 0.1912 \\
METEOR & 0.2353 & \textbf{0.6324} & \textbf{0.6126} & \textbf{0.4265} \\ \midrule
Length\^ & -0.0294 & 0.4265 & 0.2583 & 0.1618 \\
Novel 1\^ & 0.1471 & -0.2206 & -0.1402 & 0.1029 \\
Novel 2\^ & 0.0294 & -0.5441 & -0.3469 & -0.1029 \\
Novel 3\^ & 0.0294 & -0.5735 & -0.3469 & -0.1324 \\
Repeat 1\^ & \textbf{-0.3824} & 0.1029 & -0.0664 & \textbf{-0.3676} \\
Repeat 2\^ & \textbf{-0.3824} & -0.0147 & -0.2435 & -0.4559 \\
Repeat 3\^ & -0.2206 & 0.1471 & -0.0221 & -0.2647 \\
Coverage\^ & -0.1324 & 0.3529 & 0.1550 & -0.0294 \\
Compress\^ & 0.1176 & -0.4265 & -0.2288 & -0.0147 \\
Density\^ & 0.1618 & \textbf{0.6471} & 0.3911 & 0.2941 \\ \bottomrule
\end{tabular}
\caption{Kendall tau-b system-level correlation between expert annotations of coherence, consistency, fluency, and relevance and various automated metrics, adapted from \citet{summeval}. \^{}~denotes reference-free metrics. The five highest correlations per column are in bold, with ties for consistency and relevance. Coefficients with a magnitude above 0.36 are significant at the $\alpha=0.05$ level.}
\label{tab:summeval4}
\end{table}
}

\newcommand{\biastable}{
\begin{table*}[ht]
\centering
\begin{tabular}{@{}lcccccc@{}}
\toprule
\textbf{Metric} & \textbf{Info Diff} & \textbf{Shannon Score} & \textbf{Coherence} & \textbf{Consistency} & \textbf{Fluency} & \textbf{Relevance} \\ \midrule
Length & 0.5425 & 0.4291 & 0.0615 & 0.0886 & -0.0105 & 0.2054 \\
Novel 1 & -0.1140 & -0.0962 & 0.1340 & -0.2719 & -0.1924 & 0.0267 \\
Novel 2 & -0.2935 & -0.2849 & -0.0248 & -0.3693 & -0.2674 & -0.0733 \\
Novel 3 & -0.3324 & -0.3297 & -0.0781 & -0.3840 & -0.2755 & -0.1035 \\
Coverage & 0.2163 & 0.1896 & 0.0144 & 0.3369 & 0.2431 & 0.0688 \\
Compression & -0.0879 & -0.6086 & -0.0041 & -0.0697 & -0.0084 & -0.1155 \\
Density & 0.4591 & 0.4517 & 0.1991 & 0.4035 & 0.2738 & 0.2019 \\ \bottomrule
\end{tabular}
\caption{Spearman correlation of our metrics and human judged quality metrics with various statistics describing summaries across the 1700 SummEval summaries. }
\label{tab:biases}
\end{table*}
}

\newcommand{\ductablesystems}{
\begin{table}[h!]
\centering
\begin{tabular}{@{}lcc@{}}
\toprule
\textbf{Metric} & \textbf{DUC 2001} & \textbf{DUC 2002} \\ \midrule
Shannon Score & 0.2909 & 0.5714 \\
Info Diff & 0.3000 & 0.4835 \\
Jensen-Shannon & 0.4455 & 0.5440 \\
BLANC-help & 0.2727 & 0.5769 \\ \midrule
ROUGE-1 & 0.9636 & 0.9066 \\
ROUGE-2 & 0.8273 & 0.9121 \\
ROUGE-L & 0.7455 & 0.9176 \\
ROUGE-Lsum & 0.9455 & 0.9066 \\
BERTScore-P & 0.4636 & 0.5989 \\
BERTScore-R & 0.8545 & 0.9451 \\
BERTScore-F1 & 0.6091 & 0.7308 \\ \bottomrule
\end{tabular}
\caption{System-level Spearman correlation of various summary quality metrics with human-judged coverage scores on the DUC 2001 and 2002 single-document summary datasets. The last seven metrics make use of reference summaries, while the first four metrics have to rely only on the original document itself. DUC 2001 coefficients above 0.60 and DUC 2002 coefficients above 0.55 are significant at the $\alpha=0.05$ level.}
\label{tab:coverage-systems}
\end{table}
}

\newcommand{\lmcomparisontable}{
\centering
\begin{tabular}{@{}lcccc@{}}
\toprule
\textbf{Model} & \textbf{Coher.} & \textbf{Consi.} & \textbf{Fluen.} & \textbf{Relev.} \\ \midrule
\multicolumn{5}{c}{\textbf{Shannon Score}} \\
GPT-2 S & \textbf{0.4118} & \textbf{0.6324} & \textbf{0.5240} & \textbf{0.6029} \\
GPT-2 M & 0.3529 & \textbf{0.6618} & 0.4945 & 0.5441 \\
GPT-2 L & 0.3676 & \textbf{0.6471} & 0.5092 & 0.5588 \\
GPT-2 XL & 0.3824 & \textbf{0.6324} & 0.4945 & 0.5735 \\
GPT & 0.0294 & 0.5147 & 0.3469 & 0.1912 \\
XLNet & \textbf{0.4265} & 0.5882 & 0.4945 & \textbf{0.6471} \\
TransfoXL & 0.3529 & 0.5441 & 0.4502 & 0.5441 \\ \midrule
\multicolumn{5}{c}{\textbf{Information Difference}} \\
GPT-2 S & \textbf{0.4706} & \textbf{0.6324} & \textbf{0.5683} & \textbf{0.6618} \\
GPT-2 M & 0.3971 & \textbf{0.6765} & 0.5092 & 0.5882 \\
GPT-2 L & 0.3824 & \textbf{0.6324} & 0.4945 & 0.5735 \\
GPT-2 XL & 0.3971 & \textbf{0.6471} & 0.5092 & 0.5882 \\
GPT & 0.0441 & 0.5294 & 0.3616 & 0.2059 \\
XLNet & 0.4559 & 0.5882 & 0.5240 & \textbf{0.6765} \\
TransfoXL & 0.3529 & 0.5441 & 0.4502 & 0.5441 \\ \bottomrule
\end{tabular}
\captionof{table}{Kendall tau-b system-level correlations between expert annotations of coherence, consistency, fluency, and relevance and our Shannon Score and Information Difference metrics with the choice of different language models on the SummEval dataset. Scores at least as high as GPT-2 S are bold. Coefficients above 0.36 are significant at the $\alpha=0.05$ level.}
\label{tab:lms-comparison}
}

\newcommand{\upstreamtable}{
\centering
\begin{tabular}{@{}lcccc@{}}
\toprule
\textbf{k} & \textbf{Coher.} & \textbf{Consi.} & \textbf{Fluen.} & \textbf{Relev.} \\ \midrule
\multicolumn{5}{c}{\textbf{Shannon Score}} \\
0 & \textbf{0.4118} & \textbf{0.6324} & \textbf{0.5240} & \textbf{0.6029} \\
1 & 0.3529 & \textbf{0.6618} & 0.4945 & 0.5441 \\
2 & 0.3235 & \textbf{0.6618} & 0.4945 & 0.5147 \\
3 & 0.3235 & \textbf{0.6618} & 0.4945 & 0.5147 \\
4 & 0.3235 & \textbf{0.6618} & 0.4945 & 0.5147 \\ \midrule
\multicolumn{5}{c}{\textbf{Information Difference}} \\
0 & \textbf{0.4706} & \textbf{0.6324} & \textbf{0.5683} & \textbf{0.6618} \\
1 & 0.3529 & \textbf{0.6618} & 0.4945 & 0.5441 \\
2 & 0.3382 & \textbf{0.6765} & 0.5092 & 0.5294 \\
3 & 0.3235 & \textbf{0.6618} & 0.4945 & 0.5147 \\
4 & 0.3382 & \textbf{0.6765} & 0.5092 & 0.5294 \\ \bottomrule
\end{tabular}
\captionof{table}{Kendall tau-b system-level correlations between expert annotations of coherence, consistency, fluency, and relevance and our Shannon Score and Information Difference metrics with different choices of $k$ (the number of upstream sentences to provide the model) on the SummEval dataset. Scores at least as high as those of $k=0$ are bold. Coefficients above 0.36 are significant at the $\alpha=0.05$ level.}
\label{tab:upstream}
}

\newcommand{\lmcomputetable}{
\begin{table}[ht]
\centering
\begin{tabular}{@{}llcc@{}}
\toprule
\textbf{Model} & \textbf{HF Name} & \textbf{Params} & \textbf{sec} \\ \midrule
GPT-2 S & gpt2 & 124M & 13 \\
GPT-2 M & gpt2-medium & 355M & 28 \\
GPT-2 L & gpt2-large & 774M & 43 \\
GPT-2 XL & gpt2-xl & 1558M & 59 \\
GPT & openai-gpt & 117M & 15 \\
XLNet & xlnet-base-cased & 117M & 40 \\
TransfoXL & transfo-xl-wt103 & 285M & 49 \\ \bottomrule
\end{tabular}
\caption{Additional information about each language model used in our experiments: HuggingFace model hub name, number of model parameters, and number of seconds it takes to compute the Shannon Score on average for SummEval without batching.}
\label{tab:lm-compute}
\end{table}
}

%% file: doc_viz_whales.tex
\newcommand{\whalesbase}{
\cb{8}{(} \cb{3}{CNN} \cb{0}{)} \cb{13}{A} \cb{8}{North} \cb{15}{Pacific} \cb{14}{gray} \cb{1}{whale} \cb{2}{has} \cb{13}{earned} \cb{2}{a} \cb{5}{spot} \cb{2}{in} \cb{0}{the} \cb{8}{record} \cb{0}{books} \cb{3}{after} \cb{10}{completing} \cb{3}{the} \cb{2}{longest} \cb{8}{migration} \cb{1}{of} \cb{3}{a} \cb{6}{mammal} \cb{4}{ever} \cb{1}{recorded} \cb{2}{.} \cb{5}{The} \cb{18}{whale} \cb{5}{,} \cb{4}{named} \cb{14}{Var} \cb{7}{v} \cb{2}{ara} \cb{0}{,} \cb{9}{sw} \cb{0}{am} \cb{9}{nearly} \cb{9}{14} \cb{2}{,} \cb{0}{000} \cb{1}{miles} \cb{2}{(} \cb{4}{22} \cb{0}{,} \cb{4}{500} \cb{1}{kilometers} \cb{5}{),} \cb{5}{according} \cb{0}{to} \cb{3}{a} \cb{9}{release} \cb{0}{from} \cb{13}{Oregon} \cb{2}{State} \cb{0}{University} \cb{5}{,} \cb{9}{whose} \cb{5}{scientists} \cb{9}{helped} \cb{7}{conduct} \cb{0}{the} \cb{7}{whale} \cb{6}{-} \cb{8}{tracking} \cb{3}{study} \cb{0}{.} \cb{16}{Var} \cb{11}{v} \cb{0}{ara} \cb{4}{,} \cb{7}{which} \cb{2}{is} \cb{15}{Russian} \cb{0}{for} \cb{1}{"} \cb{16}{Bar} \cb{7}{bara} \cb{2}{,"} \cb{14}{left} \cb{2}{her} \cb{12}{primary} \cb{18}{feeding} \cb{5}{ground} \cb{12}{off} \cb{10}{Russia} \cb{0}{'s} \cb{9}{S} \cb{3}{akh} \cb{0}{alin} \cb{5}{Island} \cb{5}{to} \cb{12}{cross} \cb{1}{the} \cb{30}{Pacific} \cb{1}{Ocean} \cb{4}{and} \cb{11}{down} \cb{1}{the} \cb{7}{West} \cb{0}{Coast} \cb{2}{of} \cb{1}{the} \cb{0}{United} \cb{0}{States} \cb{4}{to} \cb{11}{B} \cb{0}{aja} \cb{6}{,} \cb{2}{Mexico} \cb{0}{.} \cb{16}{Var} \cb{11}{v} \cb{0}{ara} \cb{4}{'s} \cb{11}{journey} \cb{20}{surpassed} \cb{7}{a} \cb{9}{record} \cb{16}{listed} \cb{5}{on} \cb{1}{the} \cb{6}{Guinness} \cb{21}{Worlds} \cb{4}{Records} \cb{3}{website} \cb{2}{.} \cb{7}{It} \cb{12}{said} \cb{4}{the} \cb{11}{previous} \cb{8}{record} \cb{2}{was} \cb{4}{set} \cb{2}{by} \cb{4}{a} \cb{15}{hump} \cb{0}{back} \cb{0}{whale} \cb{3}{that} \cb{6}{sw} \cb{0}{am} \cb{8}{a} \cb{7}{mere} \cb{5}{10} \cb{5}{,} \cb{15}{190} \cb{11}{-} \cb{4}{mile} \cb{11}{round} \cb{0}{trip} }

\newcommand{\whalesgood}{
\cb{8}{(} \cb{8}{CNN} \cb{0}{)} \cb{10}{A} \cb{8}{North} \cb{7}{Pacific} \cb{6}{gray} \cb{0}{whale} \cb{3}{has} \cb{12}{earned} \cb{2}{a} \cb{4}{spot} \cb{2}{in} \cb{0}{the} \cb{7}{record} \cb{1}{books} \cb{3}{after} \cb{10}{completing} \cb{4}{the} \cb{1}{longest} \cb{8}{migration} \cb{1}{of} \cb{4}{a} \cb{5}{mammal} \cb{3}{ever} \cb{2}{recorded} \cb{2}{.} \cb{5}{The} \cb{5}{whale} \cb{4}{,} \cb{3}{named} \cb{1}{Var} \cb{0}{v} \cb{0}{ara} \cb{2}{,} \cb{7}{sw} \cb{0}{am} \cb{8}{nearly} \cb{8}{14} \cb{2}{,} \cb{0}{000} \cb{1}{miles} \cb{2}{(} \cb{4}{22} \cb{0}{,} \cb{4}{500} \cb{1}{kilometers} \cb{3}{),} \cb{6}{according} \cb{0}{to} \cb{3}{a} \cb{8}{release} \cb{0}{from} \cb{15}{Oregon} \cb{1}{State} \cb{0}{University} \cb{5}{,} \cb{9}{whose} \cb{5}{scientists} \cb{8}{helped} \cb{9}{conduct} \cb{0}{the} \cb{7}{whale} \cb{5}{-} \cb{7}{tracking} \cb{3}{study} \cb{0}{.} \cb{16}{Var} \cb{0}{v} \cb{0}{ara} \cb{3}{,} \cb{4}{which} \cb{2}{is} \cb{14}{Russian} \cb{0}{for} \cb{1}{"} \cb{16}{Bar} \cb{7}{bara} \cb{1}{,"} \cb{10}{left} \cb{4}{her} \cb{12}{primary} \cb{10}{feeding} \cb{6}{ground} \cb{9}{off} \cb{6}{Russia} \cb{0}{'s} \cb{8}{S} \cb{4}{akh} \cb{0}{alin} \cb{4}{Island} \cb{6}{to} \cb{8}{cross} \cb{0}{the} \cb{25}{Pacific} \cb{0}{Ocean} \cb{4}{and} \cb{12}{down} \cb{1}{the} \cb{7}{West} \cb{0}{Coast} \cb{1}{of} \cb{1}{the} \cb{0}{United} \cb{0}{States} \cb{5}{to} \cb{11}{B} \cb{0}{aja} \cb{7}{,} \cb{2}{Mexico} \cb{0}{.} \cb{16}{Var} \cb{0}{v} \cb{0}{ara} \cb{5}{'s} \cb{4}{journey} \cb{15}{surpassed} \cb{7}{a} \cb{3}{record} \cb{14}{listed} \cb{5}{on} \cb{0}{the} \cb{4}{Guinness} \cb{21}{Worlds} \cb{6}{Records} \cb{3}{website} \cb{1}{.} \cb{7}{It} \cb{13}{said} \cb{4}{the} \cb{10}{previous} \cb{0}{record} \cb{1}{was} \cb{1}{set} \cb{2}{by} \cb{3}{a} \cb{7}{hump} \cb{0}{back} \cb{0}{whale} \cb{3}{that} \cb{6}{sw} \cb{0}{am} \cb{8}{a} \cb{6}{mere} \cb{5}{10} \cb{4}{,} \cb{15}{190} \cb{10}{-} \cb{3}{mile} \cb{11}{round} \cb{0}{trip} }

\newcommand{\whalesbad}{
\cb{8}{(} \cb{3}{CNN} \cb{0}{)} \cb{11}{A} \cb{10}{North} \cb{4}{Pacific} \cb{14}{gray} \cb{0}{whale} \cb{3}{has} \cb{13}{earned} \cb{2}{a} \cb{4}{spot} \cb{2}{in} \cb{1}{the} \cb{9}{record} \cb{0}{books} \cb{3}{after} \cb{10}{completing} \cb{3}{the} \cb{3}{longest} \cb{9}{migration} \cb{1}{of} \cb{3}{a} \cb{5}{mammal} \cb{4}{ever} \cb{1}{recorded} \cb{2}{.} \cb{5}{The} \cb{5}{whale} \cb{4}{,} \cb{4}{named} \cb{16}{Var} \cb{8}{v} \cb{3}{ara} \cb{1}{,} \cb{9}{sw} \cb{0}{am} \cb{10}{nearly} \cb{8}{14} \cb{1}{,} \cb{0}{000} \cb{0}{miles} \cb{3}{(} \cb{4}{22} \cb{0}{,} \cb{4}{500} \cb{1}{kilometers} \cb{4}{),} \cb{5}{according} \cb{0}{to} \cb{3}{a} \cb{8}{release} \cb{0}{from} \cb{13}{Oregon} \cb{1}{State} \cb{0}{University} \cb{5}{,} \cb{9}{whose} \cb{6}{scientists} \cb{9}{helped} \cb{9}{conduct} \cb{0}{the} \cb{5}{whale} \cb{6}{-} \cb{9}{tracking} \cb{4}{study} \cb{0}{.} \cb{16}{Var} \cb{10}{v} \cb{0}{ara} \cb{4}{,} \cb{5}{which} \cb{2}{is} \cb{13}{Russian} \cb{0}{for} \cb{0}{"} \cb{13}{Bar} \cb{6}{bara} \cb{1}{,"} \cb{12}{left} \cb{6}{her} \cb{12}{primary} \cb{15}{feeding} \cb{6}{ground} \cb{10}{off} \cb{9}{Russia} \cb{0}{'s} \cb{9}{S} \cb{2}{akh} \cb{0}{alin} \cb{3}{Island} \cb{5}{to} \cb{11}{cross} \cb{1}{the} \cb{26}{Pacific} \cb{2}{Ocean} \cb{4}{and} \cb{12}{down} \cb{1}{the} \cb{6}{West} \cb{0}{Coast} \cb{1}{of} \cb{1}{the} \cb{0}{United} \cb{0}{States} \cb{5}{to} \cb{8}{B} \cb{0}{aja} \cb{2}{,} \cb{5}{Mexico} \cb{0}{.} \cb{16}{Var} \cb{10}{v} \cb{0}{ara} \cb{5}{'s} \cb{12}{journey} \cb{20}{surpassed} \cb{6}{a} \cb{6}{record} \cb{14}{listed} \cb{5}{on} \cb{1}{the} \cb{5}{Guinness} \cb{17}{Worlds} \cb{5}{Records} \cb{3}{website} \cb{2}{.} \cb{7}{It} \cb{6}{said} \cb{3}{the} \cb{9}{previous} \cb{9}{record} \cb{3}{was} \cb{2}{set} \cb{2}{by} \cb{3}{a} \cb{8}{hump} \cb{0}{back} \cb{4}{whale} \cb{3}{that} \cb{8}{sw} \cb{0}{am} \cb{8}{a} \cb{9}{mere} \cb{5}{10} \cb{5}{,} \cb{15}{190} \cb{12}{-} \cb{4}{mile} \cb{10}{round} \cb{1}{trip} }

\newcommand{\whalesfull}{
\cb{8}{(} \cb{3}{CNN} \cb{0}{)} \cb{2}{A} \cb{0}{North} \cb{0}{Pacific} \cb{0}{gray} \cb{0}{whale} \cb{0}{has} \cb{0}{earned} \cb{0}{a} \cb{0}{spot} \cb{0}{in} \cb{0}{the} \cb{0}{record} \cb{0}{books} \cb{0}{after} \cb{0}{completing} \cb{0}{the} \cb{0}{longest} \cb{0}{migration} \cb{0}{of} \cb{0}{a} \cb{0}{mammal} \cb{0}{ever} \cb{0}{recorded} \cb{0}{.} \cb{5}{The} \cb{3}{whale} \cb{2}{,} \cb{0}{named} \cb{0}{Var} \cb{0}{v} \cb{0}{ara} \cb{0}{,} \cb{0}{sw} \cb{0}{am} \cb{0}{nearly} \cb{0}{14} \cb{0}{,} \cb{0}{000} \cb{0}{miles} \cb{0}{(} \cb{0}{22} \cb{0}{,} \cb{0}{500} \cb{0}{kilometers} \cb{0}{),} \cb{0}{according} \cb{0}{to} \cb{0}{a} \cb{0}{release} \cb{0}{from} \cb{0}{Oregon} \cb{0}{State} \cb{0}{University} \cb{0}{,} \cb{0}{whose} \cb{0}{scientists} \cb{0}{helped} \cb{0}{conduct} \cb{0}{the} \cb{0}{whale} \cb{0}{-} \cb{0}{tracking} \cb{0}{study} \cb{0}{.} \cb{16}{Var} \cb{0}{v} \cb{0}{ara} \cb{2}{,} \cb{0}{which} \cb{0}{is} \cb{0}{Russian} \cb{0}{for} \cb{0}{"} \cb{0}{Bar} \cb{0}{bara} \cb{0}{,"} \cb{0}{left} \cb{0}{her} \cb{0}{primary} \cb{0}{feeding} \cb{0}{ground} \cb{0}{off} \cb{0}{Russia} \cb{0}{'s} \cb{0}{S} \cb{0}{akh} \cb{0}{alin} \cb{0}{Island} \cb{0}{to} \cb{0}{cross} \cb{0}{the} \cb{0}{Pacific} \cb{0}{Ocean} \cb{0}{and} \cb{0}{down} \cb{0}{the} \cb{0}{West} \cb{0}{Coast} \cb{0}{of} \cb{0}{the} \cb{0}{United} \cb{0}{States} \cb{0}{to} \cb{0}{B} \cb{0}{aja} \cb{0}{,} \cb{0}{Mexico} \cb{0}{.} \cb{16}{Var} \cb{0}{v} \cb{0}{ara} \cb{2}{'s} \cb{2}{journey} \cb{0}{surpassed} \cb{0}{a} \cb{0}{record} \cb{0}{listed} \cb{0}{on} \cb{0}{the} \cb{0}{Guinness} \cb{2}{Worlds} \cb{0}{Records} \cb{0}{website} \cb{0}{.} \cb{7}{It} \cb{2}{said} \cb{0}{the} \cb{0}{previous} \cb{0}{record} \cb{0}{was} \cb{0}{set} \cb{0}{by} \cb{0}{a} \cb{0}{hump} \cb{0}{back} \cb{0}{whale} \cb{0}{that} \cb{0}{sw} \cb{0}{am} \cb{0}{a} \cb{0}{mere} \cb{0}{10} \cb{0}{,} \cb{0}{190} \cb{0}{-} \cb{0}{mile} \cb{0}{round} \cb{0}{trip} }

\newcommand{\whaleshigh}{Varvara the gray whale traveled from Russia to Mexico, a swim of record breaking length.}

\newcommand{\whaleslow}{The round humpback has told CNN mammals that Baja was a previous Pacific website for "Guinness."}

\newcommand{\whalesfigure}{
\begin{figure*}[h!]
\small
\setlength\fboxsep{1pt}
\begin{tabular}{@{}p{0.48\linewidth}p{0.48\linewidth}@{}}
$I(\mathcal{D}) = 580$ & $I(\mathcal{D}|\mathcal{D})=52$ \\
\noalign{\vskip 2mm}
\fontdimen2\font=0pt
\whalesbase & \whalesfull \\
\noalign{\vskip 1mm}
\midrule
\noalign{\vskip 1mm}
\fontdimen2\font=\origiwspc
$I(\mathcal{D}|\mathcal{S})=482$ for this high quality summary: & $I(\mathcal{D}|\mathcal{S})=540$ for this low quality summary: \\
\noalign{\vskip 1mm}
\whaleshigh & \whaleslow \\
\noalign{\vskip 2mm}
\fontdimen2\font=0pt
\whalesgood & \whalesbad \\
\end{tabular}
\caption{A comparison of token-wise information content within a document as estimated by GPT-2 in 4 scenarios: the document on its own, the document given the document, the document given a high quality summary, and the document given a low quality summary. Tokens with a darker background color have more information.}
\label{fig:whales-trunc}
\end{figure*}
}

%% file: doc_viz_elvis.tex
\newcommand{\elvisbase}{
\cb{17}{Pr} \cb{8}{isc} \cb{0}{illa} \cb{15}{Pres} \cb{1}{ley} \cb{10}{will} \cb{9}{serve} \cb{0}{as} \cb{4}{a} \cb{10}{witness} \cb{4}{at} \cb{1}{the} \cb{9}{first} \cb{19}{wedding} \cb{7}{to} \cb{3}{be} \cb{1}{held} \cb{3}{at} \cb{11}{an} \cb{10}{all} \cb{0}{-} \cb{8}{new} \cb{10}{chapel} \cb{8}{of} \cb{15}{love} \cb{2}{in} \cb{11}{Las} \cb{0}{Vegas} \cb{2}{.} \cb{6}{The} \cb{20}{69} \cb{0}{-} \cb{0}{year} \cb{0}{-} \cb{0}{old} \cb{23}{collaborated} \cb{0}{with} \cb{16}{NBC} \cb{2}{'s} \cb{21}{ł} \cb{15}{Today} \cb{5}{show} \cb{3}{to} \cb{10}{launch} \cb{2}{a} \cb{12}{contest} \cb{3}{for} \cb{10}{one} \cb{26}{Elvis} \cb{5}{-} \cb{15}{obs} \cb{0}{essed} \cb{10}{couple} \cb{2}{to} \cb{4}{win} \cb{4}{the} \cb{9}{'} \cb{15}{ultimate} \cb{12}{wedding} \cb{9}{'.} \cb{6}{The} \cb{18}{winning} \cb{13}{duo} \cb{10}{-} \cb{16}{announced} \cb{15}{next} \cb{7}{Monday} \cb{1}{-} \cb{3}{will} \cb{15}{tie} \cb{2}{the} \cb{6}{knot} \cb{3}{at} \cb{22}{Elvis} \cb{1}{Pres} \cb{0}{ley} \cb{0}{'s} \cb{15}{Grac} \cb{0}{eland} \cb{18}{Wedding} \cb{10}{Chapel} \cb{15}{inside} \cb{2}{the} \cb{10}{West} \cb{8}{gate} \cb{6}{Hotel} \cb{4}{on} \cb{6}{Thursday} \cb{1}{,} \cb{4}{April} \cb{6}{23} \cb{1}{.} \cb{12}{No} \cb{10}{vel} \cb{16}{idea} \cb{5}{:} \cb{16}{Pr} \cb{6}{isc} \cb{0}{illa} \cb{17}{Pres} \cb{1}{ley} \cb{9}{will} \cb{11}{serve} \cb{0}{as} \cb{4}{a} \cb{11}{witness} \cb{5}{at} \cb{1}{the} \cb{9}{first} \cb{17}{wedding} \cb{6}{to} \cb{4}{be} \cb{0}{held} \cb{3}{at} \cb{10}{an} \cb{10}{all} \cb{9}{new} \cb{11}{chapel} \cb{8}{of} \cb{15}{love} \cb{2}{in} \cb{11}{Las} \cb{0}{Vegas} \cb{21}{West} \cb{13}{gate} \cb{3}{,} \cb{17}{formerly} \cb{3}{the} \cb{6}{Las} \cb{0}{Vegas} \cb{4}{Hilton} \cb{5}{,} \cb{13}{is} \cb{13}{where} \cb{12}{Elvis} \cb{11}{performed} \cb{13}{more} \cb{0}{than} \cb{11}{8} \cb{9}{30} \cb{15}{sold} \cb{0}{-} \cb{0}{out} \cb{3}{shows} \cb{3}{.} \cb{18}{Along} \cb{0}{with} \cb{3}{the} \cb{18}{singer} \cb{2}{'s} \cb{12}{former} \cb{10}{wife} \cb{11}{in} \cb{2}{the} \cb{10}{audience} \cb{0}{,} \cb{4}{the} \cb{18}{winning} \cb{10}{couple} \cb{7}{will} \cb{24}{} \cb{30}{win} \cb{2}{a} \cb{9}{free} \cb{8}{wedding} \cb{7}{reception} \cb{4}{and} \cb{13}{hotel} \cb{7}{suite} \cb{4}{for} \cb{10}{two} \cb{7}{nights} \cb{2}{.} \cb{11}{To} \cb{13}{top} \cb{0}{it} \cb{0}{off} \cb{0}{,} \cb{20}{air} \cb{12}{f} \cb{0}{ares} \cb{6}{and} \cb{21}{concert} \cb{1}{tickets} \cb{11}{to} \cb{3}{the} \cb{21}{Elvis} \cb{11}{Experience} \cb{18}{theater} \cb{11}{show} \cb{6}{will} \cb{5}{also} \cb{0}{be} \cb{17}{thrown} \cb{4}{in} \cb{3}{.} \cb{13}{While} \cb{20}{Pr} \cb{5}{isc} \cb{0}{illa} \cb{14}{agreed} \cb{1}{to} \cb{8}{make} \cb{6}{an} \cb{5}{appearance} \cb{6}{,} \cb{4}{the} \cb{11}{woman} \cb{5}{who} \cb{17}{wed} \cb{18}{Elvis} \cb{7}{in} \cb{7}{1967} \cb{9}{made} \cb{10}{one} \cb{9}{thing} \cb{0}{clear} \cb{12}{before} \cb{15}{unveiling} \cb{2}{the} \cb{12}{latest} \cb{10}{wedding} \cb{12}{chapel} \cb{6}{to} \cb{13}{bear} \cb{3}{his} \cb{0}{name} \cb{1}{:} \cb{8}{No} \cb{15}{imperson} \cb{2}{ators} \cb{1}{.} \cb{10}{'} \cb{7}{This} \cb{1}{is} \cb{10}{all} \cb{18}{first} \cb{3}{-} \cb{2}{class} \cb{6}{,'} \cb{5}{she} \cb{5}{told} \cb{3}{the} \cb{11}{Associated} \cb{0}{Press} \cb{14}{recently} \cb{0}{.} \cb{10}{'} \cb{7}{This} \cb{1}{is} \cb{4}{not} \cb{2}{a} \cb{6}{joke} \cb{3}{.} \cb{6}{The} \cb{18}{wedding} \cb{9}{chapel} \cb{4}{is} \cb{7}{not} \cb{4}{a} \cb{14}{joke} \cb{18}{.'} \cb{6}{The} \cb{18}{actress} \cb{5}{and} \cb{14}{business} \cb{9}{magn} \cb{0}{ate} \cb{4}{has} \cb{2}{been} \cb{9}{involved} \cb{0}{in} \cb{3}{the} \cb{20}{look} \cb{2}{and} \cb{2}{feel} \cb{0}{of} \cb{2}{the} \cb{24}{chapel} \cb{8}{that} \cb{16}{officials} \cb{2}{say} \cb{2}{is} \cb{6}{part} \cb{0}{of} \cb{2}{the} \cb{11}{first} \cb{13}{permanent} \cb{24}{Grac} \cb{1}{eland} \cb{14}{exhibit} \cb{7}{of} \cb{3}{the} \cb{18}{singer} \cb{1}{'s} \cb{18}{artifacts} \cb{13}{outside} \cb{7}{his} \cb{11}{iconic} \cb{12}{Memphis} \cb{7}{,} \cb{1}{Tennessee} \cb{2}{,} \cb{0}{home} \cb{0}{.} \cb{9}{C} \cb{9}{ou} \cb{1}{ples} \cb{13}{wanting} \cb{0}{to} \cb{6}{be} \cb{9}{the} \cb{5}{first} \cb{21}{} \cb{18}{to} \cb{15}{} \cb{23}{wed} \cb{8}{at} \cb{3}{the} \cb{23}{Elvis} \cb{1}{Pres} \cb{0}{ley} \cb{6}{'s} \cb{16}{Grac} \cb{1}{eland} \cb{13}{Wedding} \cb{12}{Chapel} \cb{10}{must} \cb{7}{submit} \cb{3}{a} \cb{11}{video} \cb{7}{or} \cb{7}{photos} \cb{12}{along} \cb{0}{with} \cb{6}{an} \cb{7}{explanation} \cb{11}{detailing} \cb{6}{why} \cb{1}{they} \cb{13}{deserve} \cb{3}{the} \cb{16}{prize} \cb{0}{.} \cb{13}{Ch} \cb{8}{ap} \cb{2}{el} \cb{11}{of} \cb{16}{love} \cb{5}{:} \cb{5}{The} \cb{17}{winning} \cb{12}{duo} \cb{11}{-} \cb{17}{announced} \cb{13}{next} \cb{8}{Monday} \cb{2}{-} \cb{3}{will} \cb{10}{get} \cb{5}{married} \cb{2}{in} \cb{4}{the} \cb{14}{brand} \cb{1}{new} \cb{21}{Elvis} \cb{1}{Pres} \cb{0}{ley} \cb{7}{'s} \cb{15}{Grac} \cb{0}{eland} \cb{17}{Wedding} \cb{6}{Chapel} \cb{8}{(} \cb{15}{artist} \cb{2}{'s} \cb{2}{rendering} \cb{12}{above} \cb{2}{)} \cb{4}{at} \cb{2}{the} \cb{11}{West} \cb{8}{gate} \cb{4}{Hotel} \cb{5}{on} \cb{6}{Thursday} \cb{1}{,} \cb{4}{April} \cb{6}{23} \cb{28}{Flash} \cb{2}{back} \cb{3}{:} \cb{8}{In} \cb{6}{this} \cb{8}{May} \cb{7}{1} \cb{0}{,} \cb{9}{1967} \cb{1}{,} \cb{3}{file} \cb{0}{photo} \cb{0}{,} \cb{6}{singer} \cb{4}{Elvis} \cb{0}{Pres} \cb{0}{ley} \cb{4}{,} \cb{15}{32} \cb{0}{,} \cb{3}{and} \cb{2}{his} \cb{6}{bride} \cb{2}{,} \cb{14}{21} \cb{6}{,} \cb{13}{the} \cb{8}{former} \cb{16}{Pr} \cb{1}{isc} \cb{0}{illa} \cb{14}{Be} \cb{9}{aul} \cb{0}{ieu} \cb{2}{,} \cb{11}{appear} \cb{3}{at} \cb{1}{the} \cb{11}{Al} \cb{7}{addin} \cb{3}{Hotel} \cb{1}{in} \cb{6}{Las} \cb{0}{Vegas} \cb{3}{,} \cb{7}{after} \cb{3}{their} \cb{2}{wedding} }

\newcommand{\elvisgood}{
\cb{17}{Pr} \cb{0}{isc} \cb{0}{illa} \cb{0}{Pres} \cb{0}{ley} \cb{0}{will} \cb{0}{serve} \cb{0}{as} \cb{0}{a} \cb{0}{witness} \cb{0}{at} \cb{0}{the} \cb{0}{first} \cb{0}{wedding} \cb{0}{to} \cb{0}{be} \cb{0}{held} \cb{0}{at} \cb{22}{an} \cb{9}{all} \cb{0}{-} \cb{5}{new} \cb{15}{chapel} \cb{8}{of} \cb{16}{love} \cb{3}{in} \cb{2}{Las} \cb{0}{Vegas} \cb{3}{.} \cb{6}{The} \cb{13}{69} \cb{0}{-} \cb{0}{year} \cb{0}{-} \cb{0}{old} \cb{2}{collaborated} \cb{0}{with} \cb{0}{NBC} \cb{0}{'s} \cb{18}{ł} \cb{6}{Today} \cb{0}{show} \cb{0}{to} \cb{0}{launch} \cb{0}{a} \cb{0}{contest} \cb{0}{for} \cb{0}{one} \cb{0}{Elvis} \cb{0}{-} \cb{0}{obs} \cb{0}{essed} \cb{0}{couple} \cb{0}{to} \cb{0}{win} \cb{0}{the} \cb{0}{'} \cb{0}{ultimate} \cb{0}{wedding} \cb{8}{'.} \cb{6}{The} \cb{8}{winning} \cb{6}{duo} \cb{11}{-} \cb{17}{announced} \cb{14}{next} \cb{5}{Monday} \cb{0}{-} \cb{0}{will} \cb{5}{tie} \cb{0}{the} \cb{0}{knot} \cb{1}{at} \cb{1}{Elvis} \cb{0}{Pres} \cb{0}{ley} \cb{0}{'s} \cb{0}{Grac} \cb{0}{eland} \cb{0}{Wedding} \cb{0}{Chapel} \cb{1}{inside} \cb{0}{the} \cb{0}{West} \cb{0}{gate} \cb{0}{Hotel} \cb{0}{on} \cb{0}{Thursday} \cb{0}{,} \cb{0}{April} \cb{0}{23} \cb{3}{.} \cb{12}{No} \cb{14}{vel} \cb{18}{idea} \cb{4}{:} \cb{9}{Pr} \cb{0}{isc} \cb{0}{illa} \cb{0}{Pres} \cb{0}{ley} \cb{1}{will} \cb{1}{serve} \cb{0}{as} \cb{0}{a} \cb{0}{witness} \cb{0}{at} \cb{0}{the} \cb{0}{first} \cb{0}{wedding} \cb{0}{to} \cb{0}{be} \cb{0}{held} \cb{0}{at} \cb{20}{an} \cb{9}{all} \cb{8}{new} \cb{15}{chapel} \cb{8}{of} \cb{17}{love} \cb{3}{in} \cb{1}{Las} \cb{0}{Vegas} \cb{18}{West} \cb{0}{gate} \cb{5}{,} \cb{16}{formerly} \cb{3}{the} \cb{7}{Las} \cb{0}{Vegas} \cb{4}{Hilton} \cb{5}{,} \cb{10}{is} \cb{12}{where} \cb{2}{Elvis} \cb{12}{performed} \cb{12}{more} \cb{0}{than} \cb{11}{8} \cb{9}{30} \cb{15}{sold} \cb{1}{-} \cb{0}{out} \cb{5}{shows} \cb{4}{.} \cb{18}{Along} \cb{0}{with} \cb{3}{the} \cb{10}{singer} \cb{3}{'s} \cb{10}{former} \cb{6}{wife} \cb{12}{in} \cb{3}{the} \cb{6}{audience} \cb{0}{,} \cb{6}{the} \cb{10}{winning} \cb{1}{couple} \cb{0}{will} \cb{22}{} \cb{25}{win} \cb{2}{a} \cb{7}{free} \cb{5}{wedding} \cb{11}{reception} \cb{4}{and} \cb{12}{hotel} \cb{5}{suite} \cb{5}{for} \cb{8}{two} \cb{6}{nights} \cb{2}{.} \cb{11}{To} \cb{14}{top} \cb{1}{it} \cb{0}{off} \cb{0}{,} \cb{22}{air} \cb{13}{f} \cb{1}{ares} \cb{6}{and} \cb{13}{concert} \cb{0}{tickets} \cb{7}{to} \cb{1}{the} \cb{7}{Elvis} \cb{14}{Experience} \cb{17}{theater} \cb{10}{show} \cb{3}{will} \cb{3}{also} \cb{0}{be} \cb{17}{thrown} \cb{2}{in} \cb{3}{.} \cb{13}{While} \cb{12}{Pr} \cb{0}{isc} \cb{0}{illa} \cb{18}{agreed} \cb{0}{to} \cb{9}{make} \cb{6}{an} \cb{2}{appearance} \cb{4}{,} \cb{4}{the} \cb{12}{woman} \cb{3}{who} \cb{12}{wed} \cb{1}{Elvis} \cb{5}{in} \cb{7}{1967} \cb{10}{made} \cb{10}{one} \cb{11}{thing} \cb{0}{clear} \cb{11}{before} \cb{14}{unveiling} \cb{2}{the} \cb{12}{latest} \cb{5}{wedding} \cb{11}{chapel} \cb{6}{to} \cb{18}{bear} \cb{5}{his} \cb{0}{name} \cb{1}{:} \cb{10}{No} \cb{13}{imperson} \cb{3}{ators} \cb{2}{.} \cb{10}{'} \cb{8}{This} \cb{2}{is} \cb{9}{all} \cb{16}{first} \cb{3}{-} \cb{1}{class} \cb{8}{,'} \cb{8}{she} \cb{2}{told} \cb{3}{the} \cb{7}{Associated} \cb{0}{Press} \cb{13}{recently} \cb{1}{.} \cb{10}{'} \cb{8}{This} \cb{2}{is} \cb{6}{not} \cb{2}{a} \cb{7}{joke} \cb{3}{.} \cb{6}{The} \cb{7}{wedding} \cb{6}{chapel} \cb{3}{is} \cb{10}{not} \cb{5}{a} \cb{16}{joke} \cb{18}{.'} \cb{6}{The} \cb{15}{actress} \cb{4}{and} \cb{13}{business} \cb{10}{magn} \cb{0}{ate} \cb{4}{has} \cb{2}{been} \cb{7}{involved} \cb{0}{in} \cb{2}{the} \cb{18}{look} \cb{1}{and} \cb{3}{feel} \cb{0}{of} \cb{2}{the} \cb{12}{chapel} \cb{11}{that} \cb{17}{officials} \cb{2}{say} \cb{2}{is} \cb{8}{part} \cb{0}{of} \cb{2}{the} \cb{11}{first} \cb{12}{permanent} \cb{6}{Grac} \cb{0}{eland} \cb{16}{exhibit} \cb{6}{of} \cb{3}{the} \cb{11}{singer} \cb{1}{'s} \cb{18}{artifacts} \cb{15}{outside} \cb{6}{his} \cb{9}{iconic} \cb{12}{Memphis} \cb{7}{,} \cb{1}{Tennessee} \cb{2}{,} \cb{0}{home} \cb{0}{.} \cb{9}{C} \cb{6}{ou} \cb{0}{ples} \cb{14}{wanting} \cb{0}{to} \cb{6}{be} \cb{4}{the} \cb{1}{first} \cb{20}{} \cb{15}{to} \cb{17}{} \cb{16}{wed} \cb{3}{at} \cb{3}{the} \cb{5}{Elvis} \cb{0}{Pres} \cb{0}{ley} \cb{1}{'s} \cb{1}{Grac} \cb{0}{eland} \cb{1}{Wedding} \cb{0}{Chapel} \cb{7}{must} \cb{4}{submit} \cb{2}{a} \cb{8}{video} \cb{5}{or} \cb{6}{photos} \cb{12}{along} \cb{0}{with} \cb{5}{an} \cb{9}{explanation} \cb{11}{detailing} \cb{6}{why} \cb{0}{they} \cb{13}{deserve} \cb{2}{the} \cb{14}{prize} \cb{1}{.} \cb{13}{Ch} \cb{9}{ap} \cb{0}{el} \cb{11}{of} \cb{16}{love} \cb{6}{:} \cb{5}{The} \cb{11}{winning} \cb{6}{duo} \cb{11}{-} \cb{18}{announced} \cb{14}{next} \cb{5}{Monday} \cb{1}{-} \cb{0}{will} \cb{10}{get} \cb{1}{married} \cb{3}{in} \cb{3}{the} \cb{14}{brand} \cb{0}{new} \cb{5}{Elvis} \cb{0}{Pres} \cb{0}{ley} \cb{2}{'s} \cb{3}{Grac} \cb{0}{eland} \cb{2}{Wedding} \cb{0}{Chapel} \cb{7}{(} \cb{16}{artist} \cb{1}{'s} \cb{1}{rendering} \cb{13}{above} \cb{0}{)} \cb{5}{at} \cb{2}{the} \cb{2}{West} \cb{0}{gate} \cb{0}{Hotel} \cb{2}{on} \cb{0}{Thursday} \cb{0}{,} \cb{0}{April} \cb{0}{23} \cb{27}{Flash} \cb{1}{back} \cb{1}{:} \cb{9}{In} \cb{5}{this} \cb{8}{May} \cb{6}{1} \cb{0}{,} \cb{9}{1967} \cb{1}{,} \cb{3}{file} \cb{0}{photo} \cb{0}{,} \cb{6}{singer} \cb{3}{Elvis} \cb{0}{Pres} \cb{0}{ley} \cb{4}{,} \cb{14}{32} \cb{0}{,} \cb{3}{and} \cb{2}{his} \cb{5}{bride} \cb{2}{,} \cb{13}{21} \cb{6}{,} \cb{13}{the} \cb{8}{former} \cb{16}{Pr} \cb{0}{isc} \cb{0}{illa} \cb{20}{Be} \cb{10}{aul} \cb{0}{ieu} \cb{1}{,} \cb{11}{appear} \cb{3}{at} \cb{1}{the} \cb{11}{Al} \cb{7}{addin} \cb{4}{Hotel} \cb{1}{in} \cb{3}{Las} \cb{0}{Vegas} \cb{4}{,} \cb{7}{after} \cb{4}{their} \cb{2}{wedding} }

\newcommand{\elvisbad}{
\cb{17}{Pr} \cb{5}{isc} \cb{0}{illa} \cb{4}{Pres} \cb{0}{ley} \cb{7}{will} \cb{7}{serve} \cb{0}{as} \cb{4}{a} \cb{11}{witness} \cb{5}{at} \cb{0}{the} \cb{13}{first} \cb{4}{wedding} \cb{6}{to} \cb{4}{be} \cb{0}{held} \cb{2}{at} \cb{12}{an} \cb{8}{all} \cb{0}{-} \cb{6}{new} \cb{9}{chapel} \cb{9}{of} \cb{15}{love} \cb{3}{in} \cb{9}{Las} \cb{0}{Vegas} \cb{2}{.} \cb{6}{The} \cb{14}{69} \cb{0}{-} \cb{0}{year} \cb{0}{-} \cb{0}{old} \cb{24}{collaborated} \cb{0}{with} \cb{16}{NBC} \cb{3}{'s} \cb{19}{ł} \cb{15}{Today} \cb{5}{show} \cb{3}{to} \cb{8}{launch} \cb{3}{a} \cb{11}{contest} \cb{3}{for} \cb{11}{one} \cb{17}{Elvis} \cb{6}{-} \cb{14}{obs} \cb{0}{essed} \cb{3}{couple} \cb{2}{to} \cb{3}{win} \cb{4}{the} \cb{8}{'} \cb{13}{ultimate} \cb{5}{wedding} \cb{9}{'.} \cb{6}{The} \cb{6}{winning} \cb{11}{duo} \cb{11}{-} \cb{15}{announced} \cb{15}{next} \cb{7}{Monday} \cb{1}{-} \cb{1}{will} \cb{2}{tie} \cb{0}{the} \cb{0}{knot} \cb{0}{at} \cb{0}{Elvis} \cb{0}{Pres} \cb{0}{ley} \cb{0}{'s} \cb{0}{Grac} \cb{0}{eland} \cb{0}{Wedding} \cb{0}{Chapel} \cb{0}{inside} \cb{0}{the} \cb{0}{West} \cb{0}{gate} \cb{0}{Hotel} \cb{0}{on} \cb{0}{Thursday} \cb{0}{,} \cb{0}{April} \cb{0}{23} \cb{6}{.} \cb{12}{No} \cb{15}{vel} \cb{19}{idea} \cb{4}{:} \cb{14}{Pr} \cb{1}{isc} \cb{0}{illa} \cb{6}{Pres} \cb{0}{ley} \cb{6}{will} \cb{10}{serve} \cb{0}{as} \cb{4}{a} \cb{11}{witness} \cb{4}{at} \cb{0}{the} \cb{12}{first} \cb{3}{wedding} \cb{6}{to} \cb{4}{be} \cb{1}{held} \cb{2}{at} \cb{12}{an} \cb{8}{all} \cb{9}{new} \cb{10}{chapel} \cb{9}{of} \cb{16}{love} \cb{3}{in} \cb{8}{Las} \cb{0}{Vegas} \cb{19}{West} \cb{1}{gate} \cb{4}{,} \cb{18}{formerly} \cb{3}{the} \cb{7}{Las} \cb{0}{Vegas} \cb{4}{Hilton} \cb{4}{,} \cb{10}{is} \cb{10}{where} \cb{2}{Elvis} \cb{10}{performed} \cb{13}{more} \cb{0}{than} \cb{11}{8} \cb{8}{30} \cb{15}{sold} \cb{0}{-} \cb{0}{out} \cb{4}{shows} \cb{4}{.} \cb{18}{Along} \cb{0}{with} \cb{2}{the} \cb{13}{singer} \cb{3}{'s} \cb{10}{former} \cb{6}{wife} \cb{12}{in} \cb{2}{the} \cb{8}{audience} \cb{0}{,} \cb{3}{the} \cb{7}{winning} \cb{0}{couple} \cb{1}{will} \cb{23}{} \cb{26}{win} \cb{1}{a} \cb{8}{free} \cb{6}{wedding} \cb{10}{reception} \cb{4}{and} \cb{12}{hotel} \cb{5}{suite} \cb{4}{for} \cb{9}{two} \cb{6}{nights} \cb{2}{.} \cb{11}{To} \cb{12}{top} \cb{1}{it} \cb{1}{off} \cb{0}{,} \cb{20}{air} \cb{13}{f} \cb{0}{ares} \cb{5}{and} \cb{13}{concert} \cb{0}{tickets} \cb{7}{to} \cb{1}{the} \cb{10}{Elvis} \cb{12}{Experience} \cb{18}{theater} \cb{10}{show} \cb{3}{will} \cb{2}{also} \cb{0}{be} \cb{16}{thrown} \cb{2}{in} \cb{2}{.} \cb{13}{While} \cb{16}{Pr} \cb{1}{isc} \cb{0}{illa} \cb{16}{agreed} \cb{1}{to} \cb{8}{make} \cb{7}{an} \cb{3}{appearance} \cb{4}{,} \cb{4}{the} \cb{13}{woman} \cb{3}{who} \cb{11}{wed} \cb{2}{Elvis} \cb{5}{in} \cb{8}{1967} \cb{10}{made} \cb{10}{one} \cb{10}{thing} \cb{0}{clear} \cb{11}{before} \cb{14}{unveiling} \cb{2}{the} \cb{14}{latest} \cb{5}{wedding} \cb{10}{chapel} \cb{6}{to} \cb{17}{bear} \cb{6}{his} \cb{0}{name} \cb{1}{:} \cb{11}{No} \cb{15}{imperson} \cb{3}{ators} \cb{2}{.} \cb{10}{'} \cb{7}{This} \cb{1}{is} \cb{9}{all} \cb{16}{first} \cb{4}{-} \cb{1}{class} \cb{8}{,'} \cb{2}{she} \cb{6}{told} \cb{3}{the} \cb{5}{Associated} \cb{0}{Press} \cb{15}{recently} \cb{0}{.} \cb{10}{'} \cb{7}{This} \cb{1}{is} \cb{6}{not} \cb{3}{a} \cb{9}{joke} \cb{2}{.} \cb{6}{The} \cb{6}{wedding} \cb{7}{chapel} \cb{3}{is} \cb{9}{not} \cb{5}{a} \cb{16}{joke} \cb{17}{.'} \cb{6}{The} \cb{15}{actress} \cb{4}{and} \cb{13}{business} \cb{9}{magn} \cb{0}{ate} \cb{4}{has} \cb{2}{been} \cb{7}{involved} \cb{0}{in} \cb{2}{the} \cb{18}{look} \cb{2}{and} \cb{3}{feel} \cb{0}{of} \cb{2}{the} \cb{9}{chapel} \cb{11}{that} \cb{16}{officials} \cb{2}{say} \cb{2}{is} \cb{7}{part} \cb{0}{of} \cb{1}{the} \cb{11}{first} \cb{12}{permanent} \cb{6}{Grac} \cb{0}{eland} \cb{15}{exhibit} \cb{6}{of} \cb{3}{the} \cb{15}{singer} \cb{1}{'s} \cb{19}{artifacts} \cb{14}{outside} \cb{6}{his} \cb{9}{iconic} \cb{12}{Memphis} \cb{8}{,} \cb{1}{Tennessee} \cb{2}{,} \cb{0}{home} \cb{0}{.} \cb{9}{C} \cb{4}{ou} \cb{0}{ples} \cb{14}{wanting} \cb{0}{to} \cb{6}{be} \cb{6}{the} \cb{2}{first} \cb{23}{} \cb{17}{to} \cb{18}{} \cb{18}{wed} \cb{5}{at} \cb{2}{the} \cb{8}{Elvis} \cb{0}{Pres} \cb{0}{ley} \cb{2}{'s} \cb{2}{Grac} \cb{0}{eland} \cb{0}{Wedding} \cb{0}{Chapel} \cb{9}{must} \cb{6}{submit} \cb{2}{a} \cb{9}{video} \cb{5}{or} \cb{7}{photos} \cb{11}{along} \cb{0}{with} \cb{6}{an} \cb{9}{explanation} \cb{11}{detailing} \cb{6}{why} \cb{0}{they} \cb{14}{deserve} \cb{2}{the} \cb{6}{prize} \cb{0}{.} \cb{13}{Ch} \cb{8}{ap} \cb{2}{el} \cb{10}{of} \cb{16}{love} \cb{5}{:} \cb{2}{The} \cb{5}{winning} \cb{12}{duo} \cb{12}{-} \cb{16}{announced} \cb{15}{next} \cb{7}{Monday} \cb{1}{-} \cb{1}{will} \cb{10}{get} \cb{2}{married} \cb{3}{in} \cb{3}{the} \cb{16}{brand} \cb{0}{new} \cb{7}{Elvis} \cb{0}{Pres} \cb{0}{ley} \cb{2}{'s} \cb{4}{Grac} \cb{0}{eland} \cb{3}{Wedding} \cb{0}{Chapel} \cb{7}{(} \cb{14}{artist} \cb{1}{'s} \cb{1}{rendering} \cb{14}{above} \cb{0}{)} \cb{3}{at} \cb{2}{the} \cb{2}{West} \cb{0}{gate} \cb{0}{Hotel} \cb{1}{on} \cb{0}{Thursday} \cb{0}{,} \cb{0}{April} \cb{0}{23} \cb{25}{Flash} \cb{2}{back} \cb{1}{:} \cb{8}{In} \cb{3}{this} \cb{8}{May} \cb{6}{1} \cb{0}{,} \cb{9}{1967} \cb{1}{,} \cb{3}{file} \cb{0}{photo} \cb{0}{,} \cb{6}{singer} \cb{3}{Elvis} \cb{0}{Pres} \cb{0}{ley} \cb{4}{,} \cb{14}{32} \cb{0}{,} \cb{3}{and} \cb{1}{his} \cb{5}{bride} \cb{1}{,} \cb{14}{21} \cb{4}{,} \cb{13}{the} \cb{8}{former} \cb{17}{Pr} \cb{1}{isc} \cb{0}{illa} \cb{14}{Be} \cb{10}{aul} \cb{0}{ieu} \cb{1}{,} \cb{11}{appear} \cb{3}{at} \cb{1}{the} \cb{11}{Al} \cb{7}{addin} \cb{4}{Hotel} \cb{1}{in} \cb{6}{Las} \cb{0}{Vegas} \cb{3}{,} \cb{8}{after} \cb{4}{their} \cb{1}{wedding} }

\newcommand{\elvisfull}{
\cb{17}{Pr} \cb{0}{isc} \cb{0}{illa} \cb{0}{Pres} \cb{0}{ley} \cb{0}{will} \cb{0}{serve} \cb{0}{as} \cb{0}{a} \cb{0}{witness} \cb{0}{at} \cb{0}{the} \cb{0}{first} \cb{0}{wedding} \cb{0}{to} \cb{0}{be} \cb{0}{held} \cb{0}{at} \cb{0}{an} \cb{0}{all} \cb{0}{-} \cb{0}{new} \cb{0}{chapel} \cb{0}{of} \cb{0}{love} \cb{0}{in} \cb{0}{Las} \cb{0}{Vegas} \cb{0}{.} \cb{6}{The} \cb{2}{69} \cb{0}{-} \cb{0}{year} \cb{0}{-} \cb{0}{old} \cb{0}{collaborated} \cb{0}{with} \cb{0}{NBC} \cb{0}{'s} \cb{2}{ł} \cb{0}{Today} \cb{0}{show} \cb{0}{to} \cb{0}{launch} \cb{0}{a} \cb{0}{contest} \cb{0}{for} \cb{0}{one} \cb{0}{Elvis} \cb{0}{-} \cb{0}{obs} \cb{0}{essed} \cb{0}{couple} \cb{0}{to} \cb{0}{win} \cb{0}{the} \cb{0}{'} \cb{0}{ultimate} \cb{0}{wedding} \cb{0}{'.} \cb{6}{The} \cb{7}{winning} \cb{0}{duo} \cb{0}{-} \cb{0}{announced} \cb{0}{next} \cb{0}{Monday} \cb{0}{-} \cb{0}{will} \cb{0}{tie} \cb{0}{the} \cb{0}{knot} \cb{0}{at} \cb{0}{Elvis} \cb{0}{Pres} \cb{0}{ley} \cb{0}{'s} \cb{0}{Grac} \cb{0}{eland} \cb{0}{Wedding} \cb{0}{Chapel} \cb{0}{inside} \cb{0}{the} \cb{0}{West} \cb{0}{gate} \cb{0}{Hotel} \cb{0}{on} \cb{0}{Thursday} \cb{0}{,} \cb{0}{April} \cb{0}{23} \cb{0}{.} \cb{12}{No} \cb{0}{vel} \cb{0}{idea} \cb{0}{:} \cb{4}{Pr} \cb{0}{isc} \cb{0}{illa} \cb{0}{Pres} \cb{0}{ley} \cb{0}{will} \cb{0}{serve} \cb{0}{as} \cb{0}{a} \cb{0}{witness} \cb{0}{at} \cb{0}{the} \cb{0}{first} \cb{0}{wedding} \cb{0}{to} \cb{0}{be} \cb{0}{held} \cb{0}{at} \cb{0}{an} \cb{0}{all} \cb{0}{new} \cb{0}{chapel} \cb{0}{of} \cb{0}{love} \cb{0}{in} \cb{0}{Las} \cb{0}{Vegas} \cb{0}{West} \cb{0}{gate} \cb{0}{,} \cb{0}{formerly} \cb{0}{the} \cb{0}{Las} \cb{0}{Vegas} \cb{0}{Hilton} \cb{0}{,} \cb{0}{is} \cb{0}{where} \cb{0}{Elvis} \cb{0}{performed} \cb{0}{more} \cb{0}{than} \cb{0}{8} \cb{0}{30} \cb{0}{sold} \cb{0}{-} \cb{0}{out} \cb{0}{shows} \cb{0}{.} \cb{18}{Along} \cb{0}{with} \cb{0}{the} \cb{0}{singer} \cb{0}{'s} \cb{0}{former} \cb{0}{wife} \cb{0}{in} \cb{0}{the} \cb{0}{audience} \cb{0}{,} \cb{0}{the} \cb{0}{winning} \cb{0}{couple} \cb{0}{will} \cb{5}{} \cb{3}{win} \cb{0}{a} \cb{0}{free} \cb{0}{wedding} \cb{0}{reception} \cb{0}{and} \cb{0}{hotel} \cb{0}{suite} \cb{0}{for} \cb{0}{two} \cb{0}{nights} \cb{0}{.} \cb{11}{To} \cb{6}{top} \cb{0}{it} \cb{0}{off} \cb{0}{,} \cb{1}{air} \cb{0}{f} \cb{0}{ares} \cb{0}{and} \cb{0}{concert} \cb{0}{tickets} \cb{0}{to} \cb{0}{the} \cb{0}{Elvis} \cb{0}{Experience} \cb{0}{theater} \cb{0}{show} \cb{0}{will} \cb{0}{also} \cb{0}{be} \cb{0}{thrown} \cb{0}{in} \cb{0}{.} \cb{13}{While} \cb{2}{Pr} \cb{0}{isc} \cb{0}{illa} \cb{4}{agreed} \cb{0}{to} \cb{0}{make} \cb{0}{an} \cb{0}{appearance} \cb{0}{,} \cb{0}{the} \cb{0}{woman} \cb{0}{who} \cb{0}{wed} \cb{0}{Elvis} \cb{0}{in} \cb{0}{1967} \cb{0}{made} \cb{0}{one} \cb{0}{thing} \cb{0}{clear} \cb{0}{before} \cb{0}{unveiling} \cb{0}{the} \cb{0}{latest} \cb{0}{wedding} \cb{0}{chapel} \cb{0}{to} \cb{0}{bear} \cb{0}{his} \cb{0}{name} \cb{0}{:} \cb{0}{No} \cb{0}{imperson} \cb{0}{ators} \cb{0}{.} \cb{10}{'} \cb{5}{This} \cb{0}{is} \cb{4}{all} \cb{1}{first} \cb{0}{-} \cb{0}{class} \cb{8}{,'} \cb{0}{she} \cb{0}{told} \cb{0}{the} \cb{2}{Associated} \cb{0}{Press} \cb{0}{recently} \cb{0}{.} \cb{10}{'} \cb{17}{This} \cb{0}{is} \cb{1}{not} \cb{1}{a} \cb{0}{joke} \cb{1}{.} \cb{6}{The} \cb{5}{wedding} \cb{0}{chapel} \cb{0}{is} \cb{0}{not} \cb{0}{a} \cb{0}{joke} \cb{1}{.'} \cb{6}{The} \cb{11}{actress} \cb{0}{and} \cb{0}{business} \cb{0}{magn} \cb{0}{ate} \cb{0}{has} \cb{0}{been} \cb{0}{involved} \cb{0}{in} \cb{0}{the} \cb{0}{look} \cb{0}{and} \cb{0}{feel} \cb{0}{of} \cb{0}{the} \cb{0}{chapel} \cb{0}{that} \cb{0}{officials} \cb{0}{say} \cb{0}{is} \cb{0}{part} \cb{0}{of} \cb{0}{the} \cb{0}{first} \cb{0}{permanent} \cb{0}{Grac} \cb{0}{eland} \cb{0}{exhibit} \cb{0}{of} \cb{0}{the} \cb{0}{singer} \cb{0}{'s} \cb{0}{artifacts} \cb{0}{outside} \cb{0}{his} \cb{0}{iconic} \cb{0}{Memphis} \cb{0}{,} \cb{0}{Tennessee} \cb{0}{,} \cb{0}{home} \cb{0}{.} \cb{9}{C} \cb{0}{ou} \cb{0}{ples} \cb{2}{wanting} \cb{0}{to} \cb{0}{be} \cb{0}{the} \cb{0}{first} \cb{1}{} \cb{0}{to} \cb{0}{} \cb{0}{wed} \cb{0}{at} \cb{0}{the} \cb{0}{Elvis} \cb{0}{Pres} \cb{0}{ley} \cb{0}{'s} \cb{0}{Grac} \cb{0}{eland} \cb{0}{Wedding} \cb{0}{Chapel} \cb{0}{must} \cb{0}{submit} \cb{0}{a} \cb{0}{video} \cb{0}{or} \cb{0}{photos} \cb{0}{along} \cb{0}{with} \cb{0}{an} \cb{0}{explanation} \cb{0}{detailing} \cb{0}{why} \cb{0}{they} \cb{0}{deserve} \cb{0}{the} \cb{0}{prize} \cb{0}{.} \cb{13}{Ch} \cb{0}{ap} \cb{0}{el} \cb{0}{of} \cb{0}{love} \cb{0}{:} \cb{1}{The} \cb{0}{winning} \cb{0}{duo} \cb{0}{-} \cb{0}{announced} \cb{0}{next} \cb{0}{Monday} \cb{0}{-} \cb{0}{will} \cb{0}{get} \cb{0}{married} \cb{0}{in} \cb{0}{the} \cb{0}{brand} \cb{0}{new} \cb{0}{Elvis} \cb{0}{Pres} \cb{0}{ley} \cb{0}{'s} \cb{0}{Grac} \cb{0}{eland} \cb{0}{Wedding} \cb{0}{Chapel} \cb{0}{(} \cb{0}{artist} \cb{0}{'s} \cb{0}{rendering} \cb{0}{above} \cb{0}{)} \cb{0}{at} \cb{0}{the} \cb{0}{West} \cb{0}{gate} \cb{0}{Hotel} \cb{0}{on} \cb{0}{Thursday} \cb{0}{,} \cb{0}{April} \cb{0}{23} \cb{3}{Flash} \cb{0}{back} \cb{0}{:} \cb{1}{In} \cb{0}{this} \cb{0}{May} \cb{0}{1} \cb{0}{,} \cb{0}{1967} \cb{0}{,} \cb{0}{file} \cb{0}{photo} \cb{0}{,} \cb{0}{singer} \cb{0}{Elvis} \cb{0}{Pres} \cb{0}{ley} \cb{0}{,} \cb{0}{32} \cb{0}{,} \cb{0}{and} \cb{0}{his} \cb{0}{bride} \cb{0}{,} \cb{0}{21} \cb{0}{,} \cb{0}{the} \cb{0}{former} \cb{0}{Pr} \cb{0}{isc} \cb{0}{illa} \cb{0}{Be} \cb{0}{aul} \cb{0}{ieu} \cb{0}{,} \cb{0}{appear} \cb{0}{at} \cb{0}{the} \cb{0}{Al} \cb{0}{addin} \cb{0}{Hotel} \cb{0}{in} \cb{0}{Las} \cb{0}{Vegas} \cb{0}{,} \cb{0}{after} \cb{0}{their} \cb{0}{wedding} }

\newcommand{\elvishigh}{Priscilla Presley will serve as a witness at the first wedding to be held at Elvis Presley's Graceland Wedding Chapel in Las Vegas . The 69-year-old collaborated with NBC's Today show to launch a contest for one Elvis-obsessed couple to win the 'ultimate wedding' The winning duo will tie the knot at Elvis Presley's Graceland Wedding Chapel inside the Westgate Hotel on Thursday, April 23 . }

\newcommand{\elvislow}{The winning couple will tie the knot at Elvis Presley's Graceland Wedding Chapel inside the Westgate Hotel on Thursday, April 23 'I am so excited to be a part of this,' she said. 'It is going to be a great experience for me and my family.' The bride's mother, Doris Presley, is also a part of the group that will receive a \$1 million cash prize.}

\newcommand{\elvisfigure}{
\onecolumn
\small
\setlength\fboxsep{1pt}
\begin{longtable}{@{}p{0.48\linewidth}p{0.48\linewidth}@{}}

\multicolumn{2}{c}{\cb{0}{Continued} \cb{0}{on} \cb{0}{the} \cb{0}{next} \cb{0}{page}}\\
\endfoot
\endlastfoot

\multicolumn{2}{c}{\cb{0}{Continued} \cb{0}{from} \cb{0}{the} \cb{0}{previous} \cb{0}{page}}\\
\endhead
\endfirsthead

$I(\mathcal{D}) = 1787$ & $I(\mathcal{D}|\mathcal{D})=137$ \\
\noalign{\vskip 2mm}
\fontdimen2\font=0pt
\elvisbase & \elvisfull \\
\noalign{\vskip 1mm}
\midrule 
\midrule
\noalign{\vskip 1mm}
\fontdimen2\font=\origiwspc
$I(\mathcal{D}|\mathcal{S})=1265$ for this high quality summary: & $I(\mathcal{D}|\mathcal{S})=1426$ for this low quality summary: \\
\noalign{\vskip 1mm}
\elvishigh & \elvislow \\
\noalign{\vskip 2mm}
\fontdimen2\font=0pt
\elvisgood & \elvisbad \\
\caption{A comparison of token-wise information content within a document as estimated by GPT-2 in 4 scenarios: the document on its own, the document given the document, the document given a high quality summary, and the document given a low quality summary. Tokens with a darker background color have more information.}
\label{fig:elvis-viz}
\end{longtable}
\twocolumn
}

%% file: doc_viz_cycling.tex
\newcommand{\cyclingbase}{
\cb{18}{Sir} \cb{10}{Bradley} \cb{0}{Wiggins} \cb{6}{will} \cb{15}{bid} \cb{2}{for} \cb{14}{cycling} \cb{1}{'s} \cb{24}{hour} \cb{7}{record} \cb{6}{on} \cb{9}{June} \cb{5}{7} \cb{5}{at} \cb{7}{London} \cb{1}{'s} \cb{3}{Olympic} \cb{14}{Vel} \cb{0}{od} \cb{0}{rome} \cb{1}{.} \cb{6}{The} \cb{14}{four} \cb{3}{-} \cb{4}{time} \cb{5}{Olympic} \cb{2}{champion} \cb{5}{and} \cb{10}{2012} \cb{11}{Tour} \cb{0}{de} \cb{0}{France} \cb{1}{winner} \cb{5}{,} \cb{1}{who} \cb{4}{is} \cb{13}{35} \cb{13}{on} \cb{10}{April} \cb{7}{28} \cb{0}{,} \cb{4}{will} \cb{11}{attempt} \cb{0}{to} \cb{12}{add} \cb{2}{to} \cb{0}{his} \cb{13}{accomplishments} \cb{4}{by} \cb{8}{riding} \cb{3}{the} \cb{20}{furthe} \cb{0}{st} \cb{5}{distance} \cb{3}{in} \cb{14}{60} \cb{2}{minutes} \cb{7}{at} \cb{1}{the} \cb{18}{Lee} \cb{5}{Valley} \cb{12}{Vel} \cb{6}{o} \cb{7}{Park} \cb{3}{.} \cb{9}{'} \cb{5}{The} \cb{17}{Hour} \cb{21}{Record} \cb{9}{is} \cb{3}{a} \cb{18}{holy} \cb{4}{gra} \cb{0}{il} \cb{1}{for} \cb{13}{cyclists} \cb{4}{,'} \cb{16}{Wiggins} \cb{1}{said} \cb{0}{.} \cb{16}{Four} \cb{7}{-} \cb{2}{time} \cb{4}{Olympic} \cb{3}{champion} \cb{12}{Bradley} \cb{0}{Wiggins} \cb{5}{will} \cb{15}{bid} \cb{2}{to} \cb{9}{break} \cb{20}{cycling} \cb{0}{'s} \cb{20}{hour} \cb{7}{record} \cb{4}{in} \cb{10}{June} \cb{26}{Wiggins} \cb{9}{finished} \cb{9}{his} \cb{15}{Team} \cb{0}{Sky} \cb{3}{career} \cb{4}{in} \cb{4}{the} \cb{12}{Paris} \cb{0}{-} \cb{0}{R} \cb{0}{ou} \cb{0}{ba} \cb{0}{ix} \cb{26}{253} \cb{6}{.} \cb{6}{5} \cb{1}{km} \cb{13}{one} \cb{1}{-} \cb{0}{day} \cb{1}{race} \cb{3}{on} \cb{3}{Sunday} \cb{26}{Australian} \cb{7}{rider} \cb{10}{R} \cb{2}{ohan} \cb{0}{Dennis} \cb{23}{poses} \cb{10}{after} \cb{8}{breaking} \cb{1}{the} \cb{4}{world} \cb{15}{hour} \cb{1}{record} \cb{5}{on} \cb{13}{February} \cb{5}{8} \cb{5}{in} \cb{13}{Gren} \cb{29}{chen} \cb{19}{'} \cb{10}{It} \cb{1}{'s} \cb{4}{been} \cb{19}{fought} \cb{6}{over} \cb{24}{tooth} \cb{0}{and} \cb{0}{nail} \cb{7}{by} \cb{6}{some} \cb{3}{of} \cb{0}{the} \cb{9}{greatest} \cb{10}{names} \cb{0}{in} \cb{11}{our} \cb{0}{sport} \cb{4}{for} \cb{4}{over} \cb{3}{a} \cb{7}{hundred} \cb{0}{years} \cb{5}{and} \cb{3}{it} \cb{1}{'s} \cb{3}{time} \cb{2}{for} \cb{6}{me} \cb{0}{to} \cb{10}{have} \cb{2}{a} \cb{7}{crack} \cb{0}{at} \cb{0}{it} \cb{3}{.} \cb{9}{'} \cb{4}{I} \cb{10}{like} \cb{3}{the} \cb{4}{idea} \cb{0}{of} \cb{16}{challenging} \cb{5}{myself} \cb{4}{and} \cb{18}{want} \cb{0}{to} \cb{15}{motivate} \cb{4}{people} \cb{1}{to} \cb{2}{do} \cb{1}{the} \cb{1}{same} \cb{13}{-} \cb{7}{so} \cb{8}{why} \cb{1}{not} \cb{7}{get} \cb{9}{your} \cb{13}{bike} \cb{5}{out} \cb{3}{of} \cb{0}{the} \cb{12}{shed} \cb{2}{and} \cb{9}{see} \cb{4}{how} \cb{3}{far} \cb{1}{you} \cb{0}{can} \cb{1}{go} \cb{5}{in} \cb{8}{an} \cb{0}{hour} \cb{0}{?'} \cb{10}{W} \cb{14}{iggins} \cb{4}{,} \cb{9}{whose} \cb{13}{track} \cb{20}{pedigree} \cb{3}{includes} \cb{8}{three} \cb{6}{Olympic} \cb{0}{gold} \cb{0}{medals} \cb{2}{,} \cb{3}{is} \cb{7}{expected} \cb{0}{to} \cb{13}{set} \cb{2}{a} \cb{12}{mark} \cb{13}{which} \cb{3}{will} \cb{8}{last} \cb{3}{for} \cb{7}{some} \cb{0}{time} \cb{1}{.} \cb{10}{W} \cb{14}{iggins} \cb{8}{will} \cb{14}{hope} \cb{5}{for} \cb{2}{a} \cb{21}{capacity} \cb{19}{6} \cb{0}{,} \cb{0}{000} \cb{8}{crowd} \cb{5}{to} \cb{18}{spur} \cb{10}{on} \cb{3}{his} \cb{15}{attempt} \cb{7}{,} \cb{5}{with} \cb{10}{tickets} \cb{6}{going} \cb{0}{on} \cb{0}{sale} \cb{7}{from} \cb{9}{April} \cb{6}{19} \cb{5}{,} \cb{9}{while} \cb{3}{the} \cb{9}{event} \cb{2}{will} \cb{2}{be} \cb{5}{broadcast} \cb{1}{live} \cb{1}{on} \cb{7}{Sky} \cb{0}{Sports} \cb{2}{.} \cb{8}{In} \cb{10}{June} \cb{2}{,} \cb{22}{Wiggins} \cb{9}{will} \cb{15}{hope} \cb{0}{to} \cb{15}{race} \cb{3}{in} \cb{9}{front} \cb{0}{of} \cb{4}{a} \cb{12}{sell} \cb{1}{-} \cb{0}{out} \cb{0}{crowd} \cb{1}{at} \cb{10}{London} \cb{1}{'s} \cb{4}{Olympic} \cb{21}{Vel} \cb{0}{od} \cb{0}{rome} \cb{28}{Wiggins} \cb{11}{(} \cb{4}{left} \cb{0}{)} \cb{17}{alongside} \cb{5}{his} \cb{11}{Team} \cb{0}{Sky} \cb{9}{colleague} \cb{7}{Luke} \cb{0}{Rowe} \cb{4}{after} \cb{3}{the} \cb{7}{pair} \cb{8}{raced} \cb{6}{the} \cb{13}{Paris} \cb{0}{-} \cb{0}{R} \cb{0}{ou} \cb{0}{ba} \cb{0}{ix} \cb{20}{Wiggins} \cb{8}{will} \cb{11}{look} \cb{0}{to} \cb{7}{beat} \cb{4}{the} \cb{11}{record} \cb{3}{of} \cb{18}{Dennis} \cb{17}{(} \cb{8}{pictured} \cb{5}{),} \cb{2}{who} \cb{10}{managed} \cb{1}{to} \cb{18}{cycle} \cb{24}{ł} \cb{16}{52} \cb{3}{.} \cb{17}{491} \cb{1}{km} \cb{2}{in} \cb{8}{an} \cb{5}{hour} \cb{13}{The} \cb{10}{Brit} \cb{0}{on} \cb{9}{finished} \cb{8}{his} \cb{14}{Team} \cb{0}{Sky} \cb{1}{career} \cb{5}{at} \cb{16}{Paris} \cb{0}{-} \cb{0}{R} \cb{0}{ou} \cb{0}{ba} \cb{0}{ix} \cb{8}{last} \cb{10}{Sunday} \cb{4}{and} \cb{4}{will} \cb{7}{ride} \cb{4}{in} \cb{14}{next} \cb{3}{month} \cb{0}{'s} \cb{13}{inaugural} \cb{3}{Tour} \cb{0}{de} \cb{10}{Yorkshire} \cb{9}{for} \cb{3}{his} \cb{22}{ep} \cb{0}{onymous} \cb{1}{team} \cb{12}{before} \cb{15}{preparing} \cb{1}{for} \cb{1}{the} \cb{25}{Hour} \cb{15}{as} \cb{3}{part} \cb{0}{of} \cb{1}{his} \cb{9}{return} \cb{0}{to} \cb{2}{the} \cb{7}{track} \cb{1}{.} \cb{6}{The} \cb{10}{world} \cb{20}{time} \cb{6}{-} \cb{8}{trial} \cb{7}{champion} \cb{5}{is} \cb{14}{targeting} \cb{2}{a} \cb{12}{British} \cb{5}{record} \cb{15}{eighth} \cb{10}{Olympic} \cb{2}{medal} \cb{9}{-} \cb{10}{he} \cb{4}{has} \cb{12}{four} \cb{5}{gold} \cb{8}{,} \cb{5}{one} \cb{0}{silver} \cb{0}{and} \cb{3}{two} \cb{0}{bronze} \cb{4}{-} \cb{4}{at} \cb{1}{the} \cb{7}{2016} \cb{2}{Rio} \cb{0}{Olympics} \cb{2}{in} \cb{7}{the} \cb{12}{four} \cb{2}{-} \cb{9}{man} \cb{7}{,} \cb{2}{four} \cb{0}{-} \cb{15}{kil} \cb{3}{omet} \cb{0}{re} \cb{11}{team} \cb{12}{pursuit} \cb{1}{.} \cb{6}{The} \cb{12}{current} \cb{24}{Hour} \cb{19}{record} \cb{2}{is} \cb{10}{52} \cb{4}{.} \cb{18}{491} \cb{10}{km} \cb{3}{,} \cb{9}{set} \cb{2}{by} \cb{12}{Australian} \cb{12}{R} \cb{11}{ohan} \cb{7}{Dennis} \cb{2}{in} \cb{8}{February} \cb{13}{after} \cb{5}{the} \cb{16}{UC} \cb{0}{I} \cb{11}{,} \cb{12}{cycling} \cb{1}{'s} \cb{7}{world} \cb{0}{governing} \cb{0}{body} \cb{0}{,} \cb{19}{reformed} \cb{11}{regulations} \cb{7}{,} \cb{19}{reign} \cb{6}{iting} \cb{5}{interest} \cb{0}{in} \cb{1}{the} \cb{9}{event} \cb{0}{.} \cb{17}{German} \cb{15}{J} \cb{9}{ens} \cb{5}{Vo} \cb{0}{ig} \cb{0}{t} \cb{6}{was} \cb{6}{the} \cb{3}{first} \cb{3}{to} \cb{7}{make} \cb{5}{an} \cb{6}{attempt} \cb{13}{last} \cb{7}{September} \cb{4}{,} \cb{14}{recording} \cb{20}{51} \cb{7}{.} \cb{19}{115} \cb{8}{km} \cb{5}{,} \cb{6}{a} \cb{10}{mark} \cb{5}{which} \cb{9}{stood} \cb{2}{for} \cb{11}{six} \cb{7}{weeks} \cb{6}{before} \cb{14}{Austria} \cb{3}{'s} \cb{17}{Matth} \cb{0}{ias} \cb{11}{Brand} \cb{14}{le} \cb{16}{rode} \cb{17}{51} \cb{0}{.} \cb{7}{8} \cb{10}{52} \cb{0}{km} \cb{5}{,} \cb{9}{while} \cb{16}{Jack} \cb{14}{Bob} \cb{1}{ridge} \cb{5}{was} \cb{1}{the} \cb{2}{first} \cb{0}{to} \cb{13}{fall} \cb{4}{short} \cb{4}{in} \cb{5}{his} \cb{4}{attempt} \cb{1}{.} \cb{10}{D} \cb{7}{ennis} \cb{11}{'} \cb{19}{mark} \cb{8}{will} \cb{8}{come} \cb{11}{under} \cb{8}{threat} \cb{2}{from} \cb{19}{Brit} \cb{3}{on} \cb{9}{Alex} \cb{10}{D} \cb{5}{ows} \cb{0}{ett} \cb{2}{,} \cb{1}{who} \cb{6}{will} \cb{8}{make} \cb{0}{his} \cb{17}{attempt} \cb{6}{on} \cb{11}{May} \cb{6}{2} \cb{5}{in} \cb{8}{Manchester} \cb{17}{having} \cb{10}{had} \cb{5}{to} \cb{9}{postpone} \cb{10}{it} \cb{18}{previously} \cb{7}{after} \cb{7}{suffering} \cb{1}{a} \cb{4}{broken} \cb{4}{collar} \cb{0}{bone} \cb{2}{.} \cb{20}{Tickets} \cb{4}{to} \cb{11}{watch} \cb{17}{Sir} \cb{6}{Bradley} \cb{0}{Wiggins} \cb{14}{attempt} \cb{0}{to} \cb{6}{break} \cb{1}{the} \cb{15}{UC} \cb{0}{I} \cb{14}{Hour} \cb{2}{Record} \cb{6}{at} \cb{1}{the} \cb{19}{Lee} \cb{5}{Valley} \cb{12}{Vel} \cb{9}{o} \cb{7}{Park} \cb{4}{on} \cb{7}{June} \cb{6}{7} \cb{4}{will} \cb{3}{go} \cb{0}{on} \cb{0}{sale} \cb{4}{to} \cb{1}{the} \cb{1}{general} \cb{0}{public} \cb{7}{through} \cb{9}{Sky} \cb{11}{Tickets} \cb{8}{from} \cb{7}{Friday} \cb{0}{,} \cb{7}{April} \cb{6}{19} \cb{8}{(} \cb{5}{10} \cb{1}{am} \cb{8}{)} \cb{27}{price} \cb{4}{£} \cb{7}{49} \cb{5}{,} \cb{4}{£} \cb{5}{39} \cb{2}{and} \cb{0}{£} \cb{6}{29} \cb{4}{,} \cb{10}{on} \cb{13}{line} \cb{17}{sale} \cb{11}{only} \cb{7}{through} \cb{5}{the} \cb{3}{Sky} \cb{4}{Tickets} \cb{0}{website} \cb{2}{.} }

\newcommand{\cyclinggood}{
\cb{18}{Sir} \cb{0}{Bradley} \cb{0}{Wiggins} \cb{3}{will} \cb{4}{bid} \cb{0}{for} \cb{0}{cycling} \cb{0}{'s} \cb{0}{hour} \cb{0}{record} \cb{0}{on} \cb{0}{June} \cb{0}{7} \cb{12}{at} \cb{1}{London} \cb{12}{'s} \cb{6}{Olympic} \cb{12}{Vel} \cb{1}{od} \cb{0}{rome} \cb{6}{.} \cb{6}{The} \cb{11}{four} \cb{1}{-} \cb{0}{time} \cb{4}{Olympic} \cb{1}{champion} \cb{6}{and} \cb{10}{2012} \cb{4}{Tour} \cb{0}{de} \cb{0}{France} \cb{1}{winner} \cb{6}{,} \cb{1}{who} \cb{3}{is} \cb{12}{35} \cb{13}{on} \cb{10}{April} \cb{6}{28} \cb{1}{,} \cb{2}{will} \cb{4}{attempt} \cb{0}{to} \cb{12}{add} \cb{3}{to} \cb{0}{his} \cb{16}{accomplishments} \cb{3}{by} \cb{4}{riding} \cb{4}{the} \cb{12}{furthe} \cb{0}{st} \cb{0}{distance} \cb{1}{in} \cb{4}{60} \cb{0}{minutes} \cb{1}{at} \cb{3}{the} \cb{4}{Lee} \cb{0}{Valley} \cb{0}{Vel} \cb{0}{o} \cb{0}{Park} \cb{3}{.} \cb{9}{'} \cb{6}{The} \cb{3}{Hour} \cb{5}{Record} \cb{6}{is} \cb{3}{a} \cb{18}{holy} \cb{1}{gra} \cb{0}{il} \cb{1}{for} \cb{4}{cyclists} \cb{4}{,'} \cb{3}{Wiggins} \cb{1}{said} \cb{2}{.} \cb{16}{Four} \cb{5}{-} \cb{0}{time} \cb{3}{Olympic} \cb{2}{champion} \cb{6}{Bradley} \cb{0}{Wiggins} \cb{2}{will} \cb{4}{bid} \cb{8}{to} \cb{9}{break} \cb{16}{cycling} \cb{0}{'s} \cb{2}{hour} \cb{0}{record} \cb{4}{in} \cb{9}{June} \cb{22}{Wiggins} \cb{7}{finished} \cb{2}{his} \cb{1}{Team} \cb{0}{Sky} \cb{0}{career} \cb{7}{in} \cb{6}{the} \cb{9}{Paris} \cb{0}{-} \cb{0}{R} \cb{0}{ou} \cb{0}{ba} \cb{0}{ix} \cb{26}{253} \cb{4}{.} \cb{6}{5} \cb{0}{km} \cb{13}{one} \cb{1}{-} \cb{0}{day} \cb{2}{race} \cb{2}{on} \cb{2}{Sunday} \cb{18}{Australian} \cb{6}{rider} \cb{7}{R} \cb{0}{ohan} \cb{0}{Dennis} \cb{20}{poses} \cb{11}{after} \cb{8}{breaking} \cb{1}{the} \cb{5}{world} \cb{8}{hour} \cb{0}{record} \cb{4}{on} \cb{12}{February} \cb{6}{8} \cb{3}{in} \cb{16}{Gren} \cb{27}{chen} \cb{15}{'} \cb{12}{It} \cb{1}{'s} \cb{5}{been} \cb{19}{fought} \cb{6}{over} \cb{24}{tooth} \cb{0}{and} \cb{0}{nail} \cb{6}{by} \cb{6}{some} \cb{3}{of} \cb{0}{the} \cb{8}{greatest} \cb{9}{names} \cb{0}{in} \cb{10}{our} \cb{0}{sport} \cb{4}{for} \cb{4}{over} \cb{3}{a} \cb{7}{hundred} \cb{0}{years} \cb{5}{and} \cb{3}{it} \cb{1}{'s} \cb{3}{time} \cb{2}{for} \cb{6}{me} \cb{0}{to} \cb{10}{have} \cb{2}{a} \cb{7}{crack} \cb{0}{at} \cb{0}{it} \cb{3}{.} \cb{9}{'} \cb{9}{I} \cb{9}{like} \cb{2}{the} \cb{4}{idea} \cb{0}{of} \cb{11}{challenging} \cb{1}{myself} \cb{4}{and} \cb{15}{want} \cb{0}{to} \cb{16}{motivate} \cb{6}{people} \cb{1}{to} \cb{2}{do} \cb{2}{the} \cb{0}{same} \cb{11}{-} \cb{8}{so} \cb{9}{why} \cb{0}{not} \cb{7}{get} \cb{11}{your} \cb{6}{bike} \cb{5}{out} \cb{4}{of} \cb{0}{the} \cb{11}{shed} \cb{1}{and} \cb{10}{see} \cb{3}{how} \cb{0}{far} \cb{1}{you} \cb{0}{can} \cb{1}{go} \cb{5}{in} \cb{7}{an} \cb{0}{hour} \cb{1}{?'} \cb{10}{W} \cb{0}{iggins} \cb{6}{,} \cb{8}{whose} \cb{13}{track} \cb{17}{pedigree} \cb{3}{includes} \cb{7}{three} \cb{6}{Olympic} \cb{1}{gold} \cb{0}{medals} \cb{2}{,} \cb{3}{is} \cb{7}{expected} \cb{0}{to} \cb{9}{set} \cb{2}{a} \cb{13}{mark} \cb{11}{which} \cb{3}{will} \cb{8}{last} \cb{3}{for} \cb{9}{some} \cb{0}{time} \cb{1}{.} \cb{10}{W} \cb{0}{iggins} \cb{3}{will} \cb{11}{hope} \cb{4}{for} \cb{1}{a} \cb{21}{capacity} \cb{16}{6} \cb{2}{,} \cb{1}{000} \cb{11}{crowd} \cb{4}{to} \cb{22}{spur} \cb{7}{on} \cb{1}{his} \cb{11}{attempt} \cb{6}{,} \cb{5}{with} \cb{13}{tickets} \cb{6}{going} \cb{0}{on} \cb{0}{sale} \cb{6}{from} \cb{8}{April} \cb{6}{19} \cb{7}{,} \cb{9}{while} \cb{3}{the} \cb{7}{event} \cb{2}{will} \cb{2}{be} \cb{5}{broadcast} \cb{0}{live} \cb{1}{on} \cb{1}{Sky} \cb{0}{Sports} \cb{2}{.} \cb{8}{In} \cb{8}{June} \cb{1}{,} \cb{1}{Wiggins} \cb{2}{will} \cb{13}{hope} \cb{0}{to} \cb{7}{race} \cb{3}{in} \cb{9}{front} \cb{0}{of} \cb{3}{a} \cb{14}{sell} \cb{2}{-} \cb{0}{out} \cb{0}{crowd} \cb{2}{at} \cb{6}{London} \cb{0}{'s} \cb{4}{Olympic} \cb{17}{Vel} \cb{0}{od} \cb{0}{rome} \cb{23}{Wiggins} \cb{11}{(} \cb{9}{left} \cb{0}{)} \cb{16}{alongside} \cb{5}{his} \cb{8}{Team} \cb{0}{Sky} \cb{8}{colleague} \cb{9}{Luke} \cb{0}{Rowe} \cb{5}{after} \cb{3}{the} \cb{9}{pair} \cb{8}{raced} \cb{6}{the} \cb{12}{Paris} \cb{0}{-} \cb{0}{R} \cb{0}{ou} \cb{0}{ba} \cb{0}{ix} \cb{18}{Wiggins} \cb{11}{will} \cb{11}{look} \cb{0}{to} \cb{7}{beat} \cb{2}{the} \cb{6}{record} \cb{4}{of} \cb{18}{Dennis} \cb{3}{(} \cb{11}{pictured} \cb{5}{),} \cb{2}{who} \cb{11}{managed} \cb{2}{to} \cb{16}{cycle} \cb{20}{ł} \cb{12}{52} \cb{0}{.} \cb{1}{491} \cb{0}{km} \cb{2}{in} \cb{8}{an} \cb{2}{hour} \cb{12}{The} \cb{12}{Brit} \cb{0}{on} \cb{8}{finished} \cb{4}{his} \cb{3}{Team} \cb{0}{Sky} \cb{0}{career} \cb{1}{at} \cb{3}{Paris} \cb{0}{-} \cb{0}{R} \cb{0}{ou} \cb{0}{ba} \cb{0}{ix} \cb{11}{last} \cb{6}{Sunday} \cb{4}{and} \cb{3}{will} \cb{6}{ride} \cb{5}{in} \cb{17}{next} \cb{3}{month} \cb{0}{'s} \cb{14}{inaugural} \cb{4}{Tour} \cb{0}{de} \cb{9}{Yorkshire} \cb{9}{for} \cb{4}{his} \cb{20}{ep} \cb{0}{onymous} \cb{1}{team} \cb{12}{before} \cb{15}{preparing} \cb{1}{for} \cb{1}{the} \cb{18}{Hour} \cb{10}{as} \cb{2}{part} \cb{0}{of} \cb{1}{his} \cb{10}{return} \cb{0}{to} \cb{3}{the} \cb{7}{track} \cb{2}{.} \cb{6}{The} \cb{9}{world} \cb{11}{time} \cb{11}{-} \cb{0}{trial} \cb{0}{champion} \cb{4}{is} \cb{13}{targeting} \cb{1}{a} \cb{14}{British} \cb{2}{record} \cb{16}{eighth} \cb{10}{Olympic} \cb{3}{medal} \cb{9}{-} \cb{9}{he} \cb{4}{has} \cb{8}{four} \cb{4}{gold} \cb{8}{,} \cb{4}{one} \cb{0}{silver} \cb{0}{and} \cb{3}{two} \cb{0}{bronze} \cb{5}{-} \cb{4}{at} \cb{1}{the} \cb{8}{2016} \cb{4}{Rio} \cb{0}{Olympics} \cb{2}{in} \cb{8}{the} \cb{11}{four} \cb{3}{-} \cb{10}{man} \cb{8}{,} \cb{2}{four} \cb{0}{-} \cb{12}{kil} \cb{1}{omet} \cb{0}{re} \cb{9}{team} \cb{12}{pursuit} \cb{1}{.} \cb{6}{The} \cb{10}{current} \cb{3}{Hour} \cb{0}{record} \cb{1}{is} \cb{3}{52} \cb{0}{.} \cb{9}{491} \cb{0}{km} \cb{1}{,} \cb{0}{set} \cb{0}{by} \cb{1}{Australian} \cb{0}{R} \cb{0}{ohan} \cb{0}{Dennis} \cb{0}{in} \cb{0}{February} \cb{13}{after} \cb{6}{the} \cb{12}{UC} \cb{0}{I} \cb{11}{,} \cb{10}{cycling} \cb{1}{'s} \cb{7}{world} \cb{0}{governing} \cb{0}{body} \cb{0}{,} \cb{18}{reformed} \cb{12}{regulations} \cb{8}{,} \cb{21}{reign} \cb{6}{iting} \cb{5}{interest} \cb{0}{in} \cb{2}{the} \cb{9}{event} \cb{1}{.} \cb{17}{German} \cb{11}{J} \cb{1}{ens} \cb{1}{Vo} \cb{0}{ig} \cb{0}{t} \cb{6}{was} \cb{3}{the} \cb{3}{first} \cb{4}{to} \cb{8}{make} \cb{6}{an} \cb{4}{attempt} \cb{9}{last} \cb{8}{September} \cb{6}{,} \cb{14}{recording} \cb{10}{51} \cb{0}{.} \cb{14}{115} \cb{0}{km} \cb{4}{,} \cb{5}{a} \cb{8}{mark} \cb{6}{which} \cb{10}{stood} \cb{2}{for} \cb{8}{six} \cb{4}{weeks} \cb{4}{before} \cb{19}{Austria} \cb{2}{'s} \cb{11}{Matth} \cb{0}{ias} \cb{9}{Brand} \cb{14}{le} \cb{12}{rode} \cb{11}{51} \cb{0}{.} \cb{8}{8} \cb{9}{52} \cb{0}{km} \cb{5}{,} \cb{8}{while} \cb{14}{Jack} \cb{12}{Bob} \cb{0}{ridge} \cb{5}{was} \cb{1}{the} \cb{1}{first} \cb{0}{to} \cb{13}{fall} \cb{2}{short} \cb{3}{in} \cb{4}{his} \cb{5}{attempt} \cb{3}{.} \cb{10}{D} \cb{0}{ennis} \cb{6}{'} \cb{14}{mark} \cb{8}{will} \cb{8}{come} \cb{9}{under} \cb{8}{threat} \cb{3}{from} \cb{14}{Brit} \cb{0}{on} \cb{10}{Alex} \cb{9}{D} \cb{1}{ows} \cb{0}{ett} \cb{6}{,} \cb{0}{who} \cb{3}{will} \cb{8}{make} \cb{0}{his} \cb{13}{attempt} \cb{3}{on} \cb{6}{May} \cb{5}{2} \cb{2}{in} \cb{11}{Manchester} \cb{18}{having} \cb{9}{had} \cb{6}{to} \cb{9}{postpone} \cb{11}{it} \cb{17}{previously} \cb{7}{after} \cb{6}{suffering} \cb{1}{a} \cb{3}{broken} \cb{4}{collar} \cb{0}{bone} \cb{4}{.} \cb{20}{Tickets} \cb{5}{to} \cb{8}{watch} \cb{15}{Sir} \cb{0}{Bradley} \cb{0}{Wiggins} \cb{12}{attempt} \cb{1}{to} \cb{6}{break} \cb{0}{the} \cb{20}{UC} \cb{0}{I} \cb{1}{Hour} \cb{1}{Record} \cb{5}{at} \cb{5}{the} \cb{2}{Lee} \cb{0}{Valley} \cb{0}{Vel} \cb{0}{o} \cb{0}{Park} \cb{5}{on} \cb{0}{June} \cb{0}{7} \cb{4}{will} \cb{2}{go} \cb{0}{on} \cb{0}{sale} \cb{4}{to} \cb{2}{the} \cb{0}{general} \cb{0}{public} \cb{8}{through} \cb{5}{Sky} \cb{9}{Tickets} \cb{8}{from} \cb{8}{Friday} \cb{0}{,} \cb{6}{April} \cb{6}{19} \cb{8}{(} \cb{5}{10} \cb{1}{am} \cb{7}{)} \cb{27}{price} \cb{4}{£} \cb{7}{49} \cb{5}{,} \cb{3}{£} \cb{5}{39} \cb{3}{and} \cb{0}{£} \cb{6}{29} \cb{4}{,} \cb{10}{on} \cb{12}{line} \cb{18}{sale} \cb{12}{only} \cb{7}{through} \cb{4}{the} \cb{3}{Sky} \cb{4}{Tickets} \cb{0}{website} \cb{2}{.} }

\newcommand{\cyclingbad}{
\cb{18}{Sir} \cb{0}{Bradley} \cb{0}{Wiggins} \cb{0}{will} \cb{20}{bid} \cb{0}{for} \cb{11}{cycling} \cb{0}{'s} \cb{2}{hour} \cb{0}{record} \cb{9}{on} \cb{1}{June} \cb{0}{7} \cb{0}{at} \cb{0}{London} \cb{0}{'s} \cb{0}{Olympic} \cb{0}{Vel} \cb{0}{od} \cb{0}{rome} \cb{3}{.} \cb{6}{The} \cb{13}{four} \cb{0}{-} \cb{0}{time} \cb{4}{Olympic} \cb{1}{champion} \cb{7}{and} \cb{10}{2012} \cb{6}{Tour} \cb{0}{de} \cb{0}{France} \cb{1}{winner} \cb{6}{,} \cb{2}{who} \cb{2}{is} \cb{12}{35} \cb{13}{on} \cb{9}{April} \cb{6}{28} \cb{0}{,} \cb{3}{will} \cb{2}{attempt} \cb{1}{to} \cb{13}{add} \cb{2}{to} \cb{1}{his} \cb{15}{accomplishments} \cb{4}{by} \cb{8}{riding} \cb{4}{the} \cb{17}{furthe} \cb{0}{st} \cb{1}{distance} \cb{2}{in} \cb{13}{60} \cb{1}{minutes} \cb{7}{at} \cb{1}{the} \cb{19}{Lee} \cb{4}{Valley} \cb{8}{Vel} \cb{10}{o} \cb{8}{Park} \cb{3}{.} \cb{9}{'} \cb{17}{The} \cb{10}{Hour} \cb{4}{Record} \cb{3}{is} \cb{6}{a} \cb{18}{holy} \cb{0}{gra} \cb{0}{il} \cb{0}{for} \cb{4}{cyclists} \cb{16}{,'} \cb{7}{Wiggins} \cb{1}{said} \cb{1}{.} \cb{16}{Four} \cb{6}{-} \cb{3}{time} \cb{7}{Olympic} \cb{3}{champion} \cb{22}{Bradley} \cb{0}{Wiggins} \cb{3}{will} \cb{16}{bid} \cb{4}{to} \cb{0}{break} \cb{8}{cycling} \cb{0}{'s} \cb{0}{hour} \cb{0}{record} \cb{2}{in} \cb{1}{June} \cb{4}{Wiggins} \cb{13}{finished} \cb{8}{his} \cb{16}{Team} \cb{0}{Sky} \cb{3}{career} \cb{4}{in} \cb{4}{the} \cb{14}{Paris} \cb{0}{-} \cb{0}{R} \cb{0}{ou} \cb{0}{ba} \cb{0}{ix} \cb{24}{253} \cb{6}{.} \cb{6}{5} \cb{1}{km} \cb{13}{one} \cb{2}{-} \cb{0}{day} \cb{2}{race} \cb{4}{on} \cb{6}{Sunday} \cb{23}{Australian} \cb{6}{rider} \cb{10}{R} \cb{2}{ohan} \cb{0}{Dennis} \cb{23}{poses} \cb{11}{after} \cb{7}{breaking} \cb{1}{the} \cb{3}{world} \cb{15}{hour} \cb{0}{record} \cb{5}{on} \cb{12}{February} \cb{6}{8} \cb{5}{in} \cb{13}{Gren} \cb{30}{chen} \cb{16}{'} \cb{6}{It} \cb{1}{'s} \cb{4}{been} \cb{19}{fought} \cb{6}{over} \cb{25}{tooth} \cb{0}{and} \cb{0}{nail} \cb{6}{by} \cb{6}{some} \cb{3}{of} \cb{0}{the} \cb{8}{greatest} \cb{9}{names} \cb{0}{in} \cb{10}{our} \cb{0}{sport} \cb{4}{for} \cb{4}{over} \cb{3}{a} \cb{7}{hundred} \cb{0}{years} \cb{4}{and} \cb{3}{it} \cb{0}{'s} \cb{3}{time} \cb{1}{for} \cb{6}{me} \cb{0}{to} \cb{10}{have} \cb{2}{a} \cb{7}{crack} \cb{0}{at} \cb{1}{it} \cb{3}{.} \cb{9}{'} \cb{14}{I} \cb{17}{like} \cb{5}{the} \cb{4}{idea} \cb{0}{of} \cb{12}{challenging} \cb{15}{myself} \cb{6}{and} \cb{16}{want} \cb{0}{to} \cb{15}{motivate} \cb{6}{people} \cb{0}{to} \cb{2}{do} \cb{2}{the} \cb{0}{same} \cb{12}{-} \cb{7}{so} \cb{8}{why} \cb{0}{not} \cb{7}{get} \cb{9}{your} \cb{7}{bike} \cb{5}{out} \cb{4}{of} \cb{0}{the} \cb{11}{shed} \cb{1}{and} \cb{9}{see} \cb{3}{how} \cb{3}{far} \cb{1}{you} \cb{0}{can} \cb{2}{go} \cb{4}{in} \cb{4}{an} \cb{0}{hour} \cb{11}{?'} \cb{10}{W} \cb{0}{iggins} \cb{9}{,} \cb{8}{whose} \cb{14}{track} \cb{18}{pedigree} \cb{3}{includes} \cb{7}{three} \cb{4}{Olympic} \cb{3}{gold} \cb{0}{medals} \cb{1}{,} \cb{5}{is} \cb{9}{expected} \cb{0}{to} \cb{9}{set} \cb{2}{a} \cb{11}{mark} \cb{12}{which} \cb{3}{will} \cb{9}{last} \cb{2}{for} \cb{8}{some} \cb{0}{time} \cb{1}{.} \cb{10}{W} \cb{0}{iggins} \cb{3}{will} \cb{18}{hope} \cb{7}{for} \cb{1}{a} \cb{23}{capacity} \cb{17}{6} \cb{3}{,} \cb{0}{000} \cb{14}{crowd} \cb{2}{to} \cb{23}{spur} \cb{5}{on} \cb{3}{his} \cb{8}{attempt} \cb{4}{,} \cb{6}{with} \cb{15}{tickets} \cb{6}{going} \cb{0}{on} \cb{0}{sale} \cb{7}{from} \cb{8}{April} \cb{6}{19} \cb{5}{,} \cb{9}{while} \cb{2}{the} \cb{7}{event} \cb{2}{will} \cb{2}{be} \cb{5}{broadcast} \cb{0}{live} \cb{1}{on} \cb{2}{Sky} \cb{0}{Sports} \cb{2}{.} \cb{8}{In} \cb{18}{June} \cb{5}{,} \cb{8}{Wiggins} \cb{3}{will} \cb{14}{hope} \cb{0}{to} \cb{13}{race} \cb{2}{in} \cb{10}{front} \cb{0}{of} \cb{3}{a} \cb{16}{sell} \cb{2}{-} \cb{0}{out} \cb{0}{crowd} \cb{3}{at} \cb{3}{London} \cb{0}{'s} \cb{0}{Olympic} \cb{0}{Vel} \cb{0}{od} \cb{0}{rome} \cb{6}{Wiggins} \cb{13}{(} \cb{5}{left} \cb{0}{)} \cb{15}{alongside} \cb{5}{his} \cb{13}{Team} \cb{0}{Sky} \cb{7}{colleague} \cb{3}{Luke} \cb{0}{Rowe} \cb{9}{after} \cb{3}{the} \cb{10}{pair} \cb{7}{raced} \cb{7}{the} \cb{16}{Paris} \cb{0}{-} \cb{0}{R} \cb{0}{ou} \cb{0}{ba} \cb{0}{ix} \cb{7}{Wiggins} \cb{3}{will} \cb{11}{look} \cb{0}{to} \cb{8}{beat} \cb{2}{the} \cb{4}{record} \cb{3}{of} \cb{16}{Dennis} \cb{18}{(} \cb{10}{pictured} \cb{5}{),} \cb{1}{who} \cb{11}{managed} \cb{1}{to} \cb{16}{cycle} \cb{23}{ł} \cb{16}{52} \cb{3}{.} \cb{17}{491} \cb{1}{km} \cb{2}{in} \cb{7}{an} \cb{0}{hour} \cb{10}{The} \cb{10}{Brit} \cb{0}{on} \cb{9}{finished} \cb{7}{his} \cb{14}{Team} \cb{0}{Sky} \cb{1}{career} \cb{5}{at} \cb{15}{Paris} \cb{0}{-} \cb{0}{R} \cb{0}{ou} \cb{0}{ba} \cb{0}{ix} \cb{8}{last} \cb{12}{Sunday} \cb{4}{and} \cb{4}{will} \cb{8}{ride} \cb{4}{in} \cb{14}{next} \cb{3}{month} \cb{0}{'s} \cb{15}{inaugural} \cb{3}{Tour} \cb{0}{de} \cb{9}{Yorkshire} \cb{9}{for} \cb{4}{his} \cb{22}{ep} \cb{0}{onymous} \cb{1}{team} \cb{11}{before} \cb{14}{preparing} \cb{1}{for} \cb{1}{the} \cb{22}{Hour} \cb{10}{as} \cb{2}{part} \cb{0}{of} \cb{1}{his} \cb{10}{return} \cb{0}{to} \cb{2}{the} \cb{7}{track} \cb{2}{.} \cb{6}{The} \cb{7}{world} \cb{3}{time} \cb{11}{-} \cb{0}{trial} \cb{2}{champion} \cb{6}{is} \cb{17}{targeting} \cb{2}{a} \cb{11}{British} \cb{2}{record} \cb{18}{eighth} \cb{12}{Olympic} \cb{4}{medal} \cb{8}{-} \cb{8}{he} \cb{3}{has} \cb{11}{four} \cb{6}{gold} \cb{8}{,} \cb{4}{one} \cb{0}{silver} \cb{0}{and} \cb{3}{two} \cb{0}{bronze} \cb{5}{-} \cb{6}{at} \cb{1}{the} \cb{8}{2016} \cb{5}{Rio} \cb{0}{Olympics} \cb{4}{in} \cb{8}{the} \cb{11}{four} \cb{3}{-} \cb{10}{man} \cb{8}{,} \cb{2}{four} \cb{0}{-} \cb{9}{kil} \cb{1}{omet} \cb{0}{re} \cb{10}{team} \cb{11}{pursuit} \cb{1}{.} \cb{6}{The} \cb{13}{current} \cb{12}{Hour} \cb{2}{record} \cb{2}{is} \cb{13}{52} \cb{3}{.} \cb{16}{491} \cb{3}{km} \cb{2}{,} \cb{3}{set} \cb{1}{by} \cb{10}{Australian} \cb{12}{R} \cb{4}{ohan} \cb{0}{Dennis} \cb{1}{in} \cb{9}{February} \cb{13}{after} \cb{5}{the} \cb{12}{UC} \cb{0}{I} \cb{10}{,} \cb{10}{cycling} \cb{0}{'s} \cb{8}{world} \cb{0}{governing} \cb{0}{body} \cb{0}{,} \cb{18}{reformed} \cb{12}{regulations} \cb{7}{,} \cb{20}{reign} \cb{5}{iting} \cb{5}{interest} \cb{0}{in} \cb{2}{the} \cb{8}{event} \cb{0}{.} \cb{17}{German} \cb{21}{J} \cb{7}{ens} \cb{3}{Vo} \cb{0}{ig} \cb{0}{t} \cb{7}{was} \cb{6}{the} \cb{2}{first} \cb{2}{to} \cb{9}{make} \cb{5}{an} \cb{1}{attempt} \cb{13}{last} \cb{9}{September} \cb{4}{,} \cb{14}{recording} \cb{14}{51} \cb{1}{.} \cb{18}{115} \cb{3}{km} \cb{5}{,} \cb{5}{a} \cb{6}{mark} \cb{4}{which} \cb{10}{stood} \cb{1}{for} \cb{9}{six} \cb{4}{weeks} \cb{3}{before} \cb{19}{Austria} \cb{1}{'s} \cb{12}{Matth} \cb{0}{ias} \cb{10}{Brand} \cb{13}{le} \cb{15}{rode} \cb{13}{51} \cb{0}{.} \cb{8}{8} \cb{9}{52} \cb{0}{km} \cb{5}{,} \cb{8}{while} \cb{14}{Jack} \cb{13}{Bob} \cb{0}{ridge} \cb{5}{was} \cb{1}{the} \cb{1}{first} \cb{1}{to} \cb{13}{fall} \cb{3}{short} \cb{3}{in} \cb{5}{his} \cb{1}{attempt} \cb{2}{.} \cb{10}{D} \cb{8}{ennis} \cb{10}{'} \cb{15}{mark} \cb{7}{will} \cb{9}{come} \cb{7}{under} \cb{9}{threat} \cb{3}{from} \cb{15}{Brit} \cb{3}{on} \cb{14}{Alex} \cb{9}{D} \cb{2}{ows} \cb{0}{ett} \cb{2}{,} \cb{0}{who} \cb{4}{will} \cb{10}{make} \cb{0}{his} \cb{7}{attempt} \cb{2}{on} \cb{5}{May} \cb{6}{2} \cb{4}{in} \cb{11}{Manchester} \cb{17}{having} \cb{10}{had} \cb{5}{to} \cb{9}{postpone} \cb{7}{it} \cb{17}{previously} \cb{7}{after} \cb{7}{suffering} \cb{1}{a} \cb{3}{broken} \cb{4}{collar} \cb{0}{bone} \cb{2}{.} \cb{20}{Tickets} \cb{12}{to} \cb{6}{watch} \cb{12}{Sir} \cb{0}{Bradley} \cb{0}{Wiggins} \cb{3}{attempt} \cb{0}{to} \cb{0}{break} \cb{1}{the} \cb{16}{UC} \cb{0}{I} \cb{4}{Hour} \cb{1}{Record} \cb{9}{at} \cb{3}{the} \cb{21}{Lee} \cb{5}{Valley} \cb{5}{Vel} \cb{12}{o} \cb{8}{Park} \cb{5}{on} \cb{1}{June} \cb{0}{7} \cb{4}{will} \cb{6}{go} \cb{3}{on} \cb{0}{sale} \cb{4}{to} \cb{4}{the} \cb{1}{general} \cb{0}{public} \cb{10}{through} \cb{6}{Sky} \cb{11}{Tickets} \cb{8}{from} \cb{7}{Friday} \cb{1}{,} \cb{7}{April} \cb{6}{19} \cb{9}{(} \cb{6}{10} \cb{0}{am} \cb{7}{)} \cb{23}{price} \cb{3}{£} \cb{7}{49} \cb{6}{,} \cb{4}{£} \cb{5}{39} \cb{2}{and} \cb{0}{£} \cb{6}{29} \cb{4}{,} \cb{11}{on} \cb{12}{line} \cb{16}{sale} \cb{11}{only} \cb{7}{through} \cb{5}{the} \cb{3}{Sky} \cb{6}{Tickets} \cb{1}{website} \cb{2}{.} }

\newcommand{\cyclingfull}{
\cb{18}{Sir} \cb{0}{Bradley} \cb{0}{Wiggins} \cb{0}{will} \cb{0}{bid} \cb{0}{for} \cb{0}{cycling} \cb{0}{'s} \cb{0}{hour} \cb{0}{record} \cb{0}{on} \cb{0}{June} \cb{0}{7} \cb{0}{at} \cb{0}{London} \cb{0}{'s} \cb{0}{Olympic} \cb{0}{Vel} \cb{0}{od} \cb{0}{rome} \cb{0}{.} \cb{6}{The} \cb{6}{four} \cb{0}{-} \cb{0}{time} \cb{0}{Olympic} \cb{0}{champion} \cb{0}{and} \cb{0}{2012} \cb{0}{Tour} \cb{0}{de} \cb{0}{France} \cb{0}{winner} \cb{0}{,} \cb{0}{who} \cb{0}{is} \cb{0}{35} \cb{0}{on} \cb{0}{April} \cb{0}{28} \cb{0}{,} \cb{0}{will} \cb{0}{attempt} \cb{0}{to} \cb{0}{add} \cb{0}{to} \cb{0}{his} \cb{0}{accomplishments} \cb{0}{by} \cb{0}{riding} \cb{0}{the} \cb{0}{furthe} \cb{0}{st} \cb{0}{distance} \cb{0}{in} \cb{0}{60} \cb{0}{minutes} \cb{0}{at} \cb{0}{the} \cb{0}{Lee} \cb{0}{Valley} \cb{0}{Vel} \cb{0}{o} \cb{0}{Park} \cb{0}{.} \cb{9}{'} \cb{5}{The} \cb{3}{Hour} \cb{0}{Record} \cb{2}{is} \cb{2}{a} \cb{2}{holy} \cb{0}{gra} \cb{0}{il} \cb{0}{for} \cb{0}{cyclists} \cb{4}{,'} \cb{0}{Wiggins} \cb{0}{said} \cb{0}{.} \cb{16}{Four} \cb{0}{-} \cb{0}{time} \cb{0}{Olympic} \cb{0}{champion} \cb{0}{Bradley} \cb{0}{Wiggins} \cb{0}{will} \cb{0}{bid} \cb{0}{to} \cb{0}{break} \cb{0}{cycling} \cb{0}{'s} \cb{0}{hour} \cb{0}{record} \cb{0}{in} \cb{0}{June} \cb{0}{Wiggins} \cb{0}{finished} \cb{0}{his} \cb{0}{Team} \cb{0}{Sky} \cb{0}{career} \cb{0}{in} \cb{0}{the} \cb{0}{Paris} \cb{0}{-} \cb{0}{R} \cb{0}{ou} \cb{0}{ba} \cb{0}{ix} \cb{0}{253} \cb{0}{.} \cb{0}{5} \cb{0}{km} \cb{0}{one} \cb{0}{-} \cb{0}{day} \cb{0}{race} \cb{0}{on} \cb{0}{Sunday} \cb{0}{Australian} \cb{0}{rider} \cb{0}{R} \cb{0}{ohan} \cb{0}{Dennis} \cb{0}{poses} \cb{0}{after} \cb{0}{breaking} \cb{0}{the} \cb{0}{world} \cb{0}{hour} \cb{0}{record} \cb{0}{on} \cb{0}{February} \cb{0}{8} \cb{0}{in} \cb{0}{Gren} \cb{0}{chen} \cb{0}{'} \cb{0}{It} \cb{0}{'s} \cb{0}{been} \cb{0}{fought} \cb{0}{over} \cb{0}{tooth} \cb{0}{and} \cb{0}{nail} \cb{0}{by} \cb{0}{some} \cb{0}{of} \cb{0}{the} \cb{0}{greatest} \cb{0}{names} \cb{0}{in} \cb{0}{our} \cb{0}{sport} \cb{0}{for} \cb{0}{over} \cb{0}{a} \cb{0}{hundred} \cb{0}{years} \cb{0}{and} \cb{0}{it} \cb{0}{'s} \cb{0}{time} \cb{0}{for} \cb{0}{me} \cb{0}{to} \cb{0}{have} \cb{0}{a} \cb{0}{crack} \cb{0}{at} \cb{0}{it} \cb{0}{.} \cb{9}{'} \cb{6}{I} \cb{5}{like} \cb{1}{the} \cb{0}{idea} \cb{0}{of} \cb{0}{challenging} \cb{0}{myself} \cb{0}{and} \cb{0}{want} \cb{0}{to} \cb{0}{motivate} \cb{0}{people} \cb{0}{to} \cb{0}{do} \cb{0}{the} \cb{0}{same} \cb{0}{-} \cb{0}{so} \cb{0}{why} \cb{0}{not} \cb{0}{get} \cb{0}{your} \cb{0}{bike} \cb{0}{out} \cb{0}{of} \cb{0}{the} \cb{0}{shed} \cb{0}{and} \cb{0}{see} \cb{0}{how} \cb{0}{far} \cb{0}{you} \cb{0}{can} \cb{0}{go} \cb{0}{in} \cb{0}{an} \cb{0}{hour} \cb{0}{?'} \cb{10}{W} \cb{0}{iggins} \cb{3}{,} \cb{5}{whose} \cb{1}{track} \cb{0}{pedigree} \cb{0}{includes} \cb{0}{three} \cb{0}{Olympic} \cb{0}{gold} \cb{0}{medals} \cb{0}{,} \cb{0}{is} \cb{0}{expected} \cb{0}{to} \cb{0}{set} \cb{0}{a} \cb{0}{mark} \cb{0}{which} \cb{0}{will} \cb{0}{last} \cb{0}{for} \cb{0}{some} \cb{0}{time} \cb{0}{.} \cb{10}{W} \cb{0}{iggins} \cb{3}{will} \cb{6}{hope} \cb{0}{for} \cb{0}{a} \cb{0}{capacity} \cb{0}{6} \cb{0}{,} \cb{0}{000} \cb{0}{crowd} \cb{0}{to} \cb{0}{spur} \cb{0}{on} \cb{0}{his} \cb{0}{attempt} \cb{0}{,} \cb{0}{with} \cb{0}{tickets} \cb{0}{going} \cb{0}{on} \cb{0}{sale} \cb{0}{from} \cb{0}{April} \cb{0}{19} \cb{0}{,} \cb{0}{while} \cb{0}{the} \cb{0}{event} \cb{0}{will} \cb{0}{be} \cb{0}{broadcast} \cb{0}{live} \cb{0}{on} \cb{0}{Sky} \cb{0}{Sports} \cb{0}{.} \cb{8}{In} \cb{0}{June} \cb{0}{,} \cb{0}{Wiggins} \cb{0}{will} \cb{0}{hope} \cb{0}{to} \cb{0}{race} \cb{0}{in} \cb{0}{front} \cb{0}{of} \cb{0}{a} \cb{0}{sell} \cb{0}{-} \cb{0}{out} \cb{0}{crowd} \cb{0}{at} \cb{0}{London} \cb{0}{'s} \cb{0}{Olympic} \cb{0}{Vel} \cb{0}{od} \cb{0}{rome} \cb{2}{Wiggins} \cb{0}{(} \cb{0}{left} \cb{0}{)} \cb{0}{alongside} \cb{0}{his} \cb{0}{Team} \cb{0}{Sky} \cb{0}{colleague} \cb{0}{Luke} \cb{0}{Rowe} \cb{0}{after} \cb{0}{the} \cb{0}{pair} \cb{0}{raced} \cb{0}{the} \cb{0}{Paris} \cb{0}{-} \cb{0}{R} \cb{0}{ou} \cb{0}{ba} \cb{0}{ix} \cb{1}{Wiggins} \cb{1}{will} \cb{0}{look} \cb{0}{to} \cb{0}{beat} \cb{0}{the} \cb{0}{record} \cb{0}{of} \cb{0}{Dennis} \cb{0}{(} \cb{0}{pictured} \cb{0}{),} \cb{0}{who} \cb{0}{managed} \cb{0}{to} \cb{0}{cycle} \cb{2}{ł} \cb{0}{52} \cb{0}{.} \cb{0}{491} \cb{0}{km} \cb{0}{in} \cb{0}{an} \cb{0}{hour} \cb{5}{The} \cb{0}{Brit} \cb{0}{on} \cb{0}{finished} \cb{0}{his} \cb{0}{Team} \cb{0}{Sky} \cb{0}{career} \cb{0}{at} \cb{0}{Paris} \cb{0}{-} \cb{0}{R} \cb{0}{ou} \cb{0}{ba} \cb{0}{ix} \cb{0}{last} \cb{0}{Sunday} \cb{0}{and} \cb{0}{will} \cb{0}{ride} \cb{0}{in} \cb{0}{next} \cb{0}{month} \cb{0}{'s} \cb{0}{inaugural} \cb{0}{Tour} \cb{0}{de} \cb{0}{Yorkshire} \cb{0}{for} \cb{0}{his} \cb{0}{ep} \cb{0}{onymous} \cb{0}{team} \cb{0}{before} \cb{0}{preparing} \cb{0}{for} \cb{0}{the} \cb{0}{Hour} \cb{0}{as} \cb{0}{part} \cb{0}{of} \cb{0}{his} \cb{0}{return} \cb{0}{to} \cb{0}{the} \cb{0}{track} \cb{0}{.} \cb{6}{The} \cb{7}{world} \cb{0}{time} \cb{3}{-} \cb{0}{trial} \cb{0}{champion} \cb{0}{is} \cb{0}{targeting} \cb{0}{a} \cb{0}{British} \cb{0}{record} \cb{0}{eighth} \cb{0}{Olympic} \cb{0}{medal} \cb{0}{-} \cb{0}{he} \cb{0}{has} \cb{0}{four} \cb{0}{gold} \cb{0}{,} \cb{0}{one} \cb{0}{silver} \cb{0}{and} \cb{0}{two} \cb{0}{bronze} \cb{0}{-} \cb{0}{at} \cb{0}{the} \cb{0}{2016} \cb{0}{Rio} \cb{0}{Olympics} \cb{0}{in} \cb{0}{the} \cb{0}{four} \cb{0}{-} \cb{0}{man} \cb{0}{,} \cb{0}{four} \cb{0}{-} \cb{0}{kil} \cb{0}{omet} \cb{0}{re} \cb{0}{team} \cb{0}{pursuit} \cb{0}{.} \cb{6}{The} \cb{5}{current} \cb{1}{Hour} \cb{0}{record} \cb{0}{is} \cb{0}{52} \cb{0}{.} \cb{0}{491} \cb{0}{km} \cb{0}{,} \cb{0}{set} \cb{0}{by} \cb{0}{Australian} \cb{0}{R} \cb{0}{ohan} \cb{0}{Dennis} \cb{0}{in} \cb{0}{February} \cb{0}{after} \cb{0}{the} \cb{0}{UC} \cb{0}{I} \cb{0}{,} \cb{0}{cycling} \cb{0}{'s} \cb{0}{world} \cb{0}{governing} \cb{0}{body} \cb{0}{,} \cb{0}{reformed} \cb{0}{regulations} \cb{0}{,} \cb{0}{reign} \cb{0}{iting} \cb{0}{interest} \cb{0}{in} \cb{0}{the} \cb{0}{event} \cb{0}{.} \cb{17}{German} \cb{1}{J} \cb{0}{ens} \cb{0}{Vo} \cb{0}{ig} \cb{0}{t} \cb{1}{was} \cb{0}{the} \cb{0}{first} \cb{0}{to} \cb{0}{make} \cb{0}{an} \cb{0}{attempt} \cb{0}{last} \cb{0}{September} \cb{0}{,} \cb{0}{recording} \cb{0}{51} \cb{0}{.} \cb{0}{115} \cb{0}{km} \cb{0}{,} \cb{0}{a} \cb{0}{mark} \cb{0}{which} \cb{0}{stood} \cb{0}{for} \cb{0}{six} \cb{0}{weeks} \cb{0}{before} \cb{0}{Austria} \cb{0}{'s} \cb{0}{Matth} \cb{0}{ias} \cb{0}{Brand} \cb{0}{le} \cb{0}{rode} \cb{0}{51} \cb{0}{.} \cb{0}{8} \cb{0}{52} \cb{0}{km} \cb{0}{,} \cb{0}{while} \cb{0}{Jack} \cb{0}{Bob} \cb{0}{ridge} \cb{0}{was} \cb{0}{the} \cb{0}{first} \cb{0}{to} \cb{0}{fall} \cb{0}{short} \cb{0}{in} \cb{0}{his} \cb{0}{attempt} \cb{0}{.} \cb{10}{D} \cb{14}{ennis} \cb{2}{'} \cb{1}{mark} \cb{0}{will} \cb{0}{come} \cb{0}{under} \cb{0}{threat} \cb{0}{from} \cb{0}{Brit} \cb{0}{on} \cb{0}{Alex} \cb{0}{D} \cb{0}{ows} \cb{0}{ett} \cb{0}{,} \cb{0}{who} \cb{0}{will} \cb{0}{make} \cb{0}{his} \cb{0}{attempt} \cb{0}{on} \cb{0}{May} \cb{0}{2} \cb{0}{in} \cb{0}{Manchester} \cb{0}{having} \cb{0}{had} \cb{0}{to} \cb{0}{postpone} \cb{0}{it} \cb{0}{previously} \cb{0}{after} \cb{0}{suffering} \cb{0}{a} \cb{0}{broken} \cb{0}{collar} \cb{0}{bone} \cb{0}{.} \cb{20}{Tickets} \cb{2}{to} \cb{0}{watch} \cb{0}{Sir} \cb{0}{Bradley} \cb{0}{Wiggins} \cb{0}{attempt} \cb{0}{to} \cb{0}{break} \cb{0}{the} \cb{0}{UC} \cb{0}{I} \cb{0}{Hour} \cb{0}{Record} \cb{0}{at} \cb{0}{the} \cb{0}{Lee} \cb{0}{Valley} \cb{0}{Vel} \cb{0}{o} \cb{0}{Park} \cb{0}{on} \cb{0}{June} \cb{0}{7} \cb{0}{will} \cb{0}{go} \cb{0}{on} \cb{0}{sale} \cb{0}{to} \cb{0}{the} \cb{0}{general} \cb{0}{public} \cb{0}{through} \cb{0}{Sky} \cb{0}{Tickets} \cb{0}{from} \cb{0}{Friday} \cb{0}{,} \cb{0}{April} \cb{0}{19} \cb{0}{(} \cb{0}{10} \cb{0}{am} \cb{0}{)} \cb{0}{price} \cb{0}{£} \cb{0}{49} \cb{0}{,} \cb{0}{£} \cb{0}{39} \cb{0}{and} \cb{0}{£} \cb{0}{29} \cb{0}{,} \cb{0}{on} \cb{0}{line} \cb{0}{sale} \cb{0}{only} \cb{0}{through} \cb{0}{the} \cb{0}{Sky} \cb{0}{Tickets} \cb{0}{website} \cb{0}{.} }

\newcommand{\cyclinghigh}{Bradley Wiggins will bid for cycling's hour record on June 7 in London . Wiggins will attempt to ride furthest distance in 60 minutes at Lee Valley VeloPark . The Hour record is 52.491km, set by Australian Rohan Dennis in February . Wiggins finished his Team Sky career at Paris-Roubaix on Sunday . }

\newcommand{\cyclinglow}{Sir Bradley Wiggins will attempt to break cycling's hour record in June Wiggins' record attempt will take place on June 7 at London's Olympic Velodrome Wiggins will be joined in the attempt by fellow British record holder Luke Rowe, who is also a time trial specialist. Rowe, who is also a former world time trial champion, is the only man to have completed the distance in an hour. Wiggins' attempt will mark the first time a British}

\newcommand{\cyclingfigure}{
\onecolumn
\small
\setlength\fboxsep{1pt}
\begin{longtable}{@{}p{0.48\linewidth}p{0.48\linewidth}@{}}

\multicolumn{2}{c}{\cb{0}{Continued} \cb{0}{on} \cb{0}{the} \cb{0}{next} \cb{0}{page}}\\
\endfoot
\endlastfoot

$I(\mathcal{D}) = 2367$ & $I(\mathcal{D}|\mathcal{D})=148$ \\
\noalign{\vskip 2mm}
\fontdimen2\font=0pt
\cyclingbase & \cyclingfull \\
\noalign{\vskip 1mm}
\midrule 
\midrule
\noalign{\vskip 1mm}
\fontdimen2\font=\origiwspc
$I(\mathcal{D}|\mathcal{S})=1892$ for this high quality summary: & $I(\mathcal{D}|\mathcal{S})=2101$ for this low quality summary: \\
\noalign{\vskip 1mm}
\cyclinghigh & \cyclinglow \\
\noalign{\vskip 2mm}
\fontdimen2\font=0pt
\cyclinggood & \cyclingbad \\
\caption{A comparison of token-wise information content within a document as estimated by GPT-2 in 4 scenarios: the document on its own, the document given the document, the document given a high quality summary, and the document given a low quality summary. Tokens with a darker background color have more information.}
\label{fig:cycling-viz}
\end{longtable}
\twocolumn
}

%% file: doc_viz_italy.tex
\newcommand{\italybase}{
\cb{11}{R} \cb{9}{ome} \cb{6}{(} \cb{6}{CNN} \cb{0}{)} \cb{28}{Italy} \cb{4}{is} \cb{16}{coping} \cb{0}{with} \cb{2}{a} \cb{10}{rising} \cb{8}{wave} \cb{0}{of} \cb{16}{desperate} \cb{1}{migrants} \cb{5}{from} \cb{5}{Africa} \cb{3}{and} \cb{8}{Middle} \cb{0}{East} \cb{12}{hoping} \cb{0}{to} \cb{5}{make} \cb{1}{it} \cb{0}{to} \cb{1}{Europe} \cb{2}{.} \cb{13}{From} \cb{19}{Friday} \cb{6}{to} \cb{4}{Monday} \cb{1}{,} \cb{8}{a} \cb{9}{total} \cb{0}{of} \cb{8}{8} \cb{3}{,} \cb{14}{480} \cb{10}{migrants} \cb{3}{were} \cb{6}{rescued} \cb{6}{,} \cb{5}{according} \cb{0}{to} \cb{1}{the} \cb{6}{Italian} \cb{5}{coast} \cb{2}{guard} \cb{2}{,} \cb{4}{which} \cb{2}{said} \cb{3}{it} \cb{7}{received} \cb{17}{on} \cb{4}{Monday} \cb{21}{--} \cb{18}{alone} \cb{2}{--} \cb{28}{SOS} \cb{10}{calls} \cb{3}{from} \cb{13}{20} \cb{7}{boats} \cb{5}{in} \cb{11}{distress} \cb{0}{.} \cb{11}{On} \cb{7}{Tuesday} \cb{0}{,} \cb{6}{a} \cb{11}{spokesman} \cb{10}{with} \cb{19}{Save} \cb{1}{the} \cb{0}{Children} \cb{6}{told} \cb{6}{CNN} \cb{5}{the} \cb{3}{group} \cb{14}{fears} \cb{20}{400} \cb{11}{migrants} \cb{4}{could} \cb{2}{be} \cb{11}{missing} \cb{6}{,} \cb{8}{citing} \cb{16}{testimony} \cb{1}{from} \cb{9}{survivors} \cb{4}{who} \cb{2}{said} \cb{6}{their} \cb{14}{ship} \cb{14}{carrying} \cb{13}{550} \cb{3}{people} \cb{8}{caps} \cb{0}{ized} \cb{3}{in} \cb{1}{the} \cb{1}{Mediterranean} \cb{3}{Sea} \cb{11}{about} \cb{9}{80} \cb{1}{miles} \cb{6}{off} \cb{0}{the} \cb{6}{Libyan} \cb{0}{coast} \cb{1}{.} \cb{6}{The} \cb{16}{Italian} \cb{12}{coast} \cb{1}{guard} \cb{7}{,} \cb{8}{however} \cb{0}{,} \cb{10}{told} \cb{7}{CNN} \cb{2}{that} \cb{11}{while} \cb{3}{it} \cb{4}{is} \cb{11}{taking} \cb{5}{the} \cb{10}{report} \cb{0}{seriously} \cb{0}{,} \cb{1}{it} \cb{8}{cannot} \cb{4}{confirm} \cb{13}{such} \cb{3}{an} \cb{2}{incident} \cb{7}{and} \cb{5}{has} \cb{2}{not} \cb{4}{yet} \cb{10}{found} \cb{5}{evidence} \cb{13}{at} \cb{3}{sea} \cb{6}{to} \cb{8}{indicate} \cb{5}{a} \cb{12}{migrant} \cb{2}{boat} \cb{5}{carrying} \cb{12}{approximately} \cb{12}{550} \cb{16}{has} \cb{6}{caps} \cb{0}{ized} \cb{14}{with} \cb{21}{145} \cb{16}{rescued} \cb{2}{.} \cb{12}{An} \cb{15}{operation} \cb{6}{that} \cb{11}{included} \cb{17}{boats} \cb{2}{and} \cb{7}{planes} \cb{14}{did} \cb{0}{not} \cb{8}{find} \cb{1}{any} \cb{7}{survivors} \cb{3}{,} \cb{10}{nor} \cb{18}{bodies} \cb{2}{,} \cb{6}{nor} \cb{3}{any} \cb{5}{evidence} \cb{6}{to} \cb{5}{indicate} \cb{5}{a} \cb{14}{particular} \cb{9}{boat} \cb{17}{caps} \cb{0}{ized} \cb{3}{,} \cb{17}{Coast} \cb{0}{Guard} \cb{12}{official} \cb{15}{F} \cb{9}{ilipp} \cb{0}{o} \cb{9}{Mar} \cb{7}{ini} \cb{1}{said} \cb{1}{.} \cb{11}{There} \cb{7}{has} \cb{0}{been} \cb{1}{a} \cb{8}{recent} \cb{17}{up} \cb{0}{sur} \cb{0}{ge} \cb{0}{in} \cb{16}{migrant} \cb{12}{boats} \cb{9}{crossing} \cb{0}{the} \cb{1}{Mediterranean} \cb{7}{into} \cb{4}{Italy} \cb{4}{and} \cb{14}{an} \cb{3}{increase} \cb{0}{in} \cb{16}{resc} \cb{0}{ues} \cb{14}{performed} \cb{1}{by} \cb{3}{the} \cb{1}{Italian} \cb{3}{Coast} \cb{0}{Guard} \cb{10}{to} \cb{7}{aid} \cb{8}{migrant} \cb{1}{boats} \cb{2}{.} \cb{14}{Why} \cb{23}{migrants} \cb{3}{are} \cb{10}{dying} \cb{19}{trying} \cb{0}{to} \cb{2}{reach} \cb{9}{Italy} \cb{18}{According} \cb{0}{to} \cb{3}{the} \cb{7}{International} \cb{2}{Organization} \cb{0}{for} \cb{0}{Migration} \cb{1}{,} \cb{6}{Italy} \cb{15}{registered} \cb{5}{more} \cb{2}{than} \cb{8}{10} \cb{0}{,} \cb{0}{000} \cb{4}{migrants} \cb{9}{arriving} \cb{1}{in} \cb{2}{the} \cb{6}{first} \cb{4}{three} \cb{0}{months} \cb{0}{of} \cb{3}{2015} \cb{2}{,} \cb{6}{and} \cb{8}{about} \cb{5}{2} \cb{0}{,} \cb{2}{000} \cb{6}{were} \cb{7}{rescued} \cb{9}{at} \cb{1}{sea} \cb{7}{during} \cb{1}{the} \cb{2}{first} \cb{15}{weekend} \cb{0}{of} \cb{6}{April} \cb{8}{in} \cb{2}{the} \cb{14}{Channel} \cb{8}{of} \cb{0}{Sicily} \cb{0}{.} \cb{14}{Most} \cb{18}{migrants} \cb{23}{recorded} \cb{10}{this} \cb{4}{year} \cb{12}{come} \cb{0}{from} \cb{6}{countries} \cb{6}{in} \cb{11}{West} \cb{0}{Africa} \cb{11}{as} \cb{4}{well} \cb{0}{as} \cb{11}{Somalia} \cb{2}{and} \cb{8}{Syria} \cb{2}{,} \cb{6}{the} \cb{17}{I} \cb{2}{OM} \cb{2}{said} \cb{1}{.} \cb{13}{They} \cb{11}{use} \cb{25}{Libya} \cb{1}{as} \cb{1}{a} \cb{13}{country} \cb{2}{of} \cb{14}{transit} \cb{4}{.} \cb{12}{At} \cb{4}{least} \cb{20}{480} \cb{9}{migrants} \cb{2}{have} \cb{3}{died} \cb{9}{while} \cb{3}{crossing} \cb{0}{the} \cb{0}{Mediterranean} \cb{7}{since} \cb{2}{the} \cb{3}{beginning} \cb{0}{of} \cb{2}{the} \cb{0}{year} \cb{0}{,} \cb{15}{often} \cb{8}{because} \cb{2}{of} \cb{13}{bad} \cb{0}{weather} \cb{4}{and} \cb{5}{overcrowd} \cb{3}{ed} \cb{8}{vessels} \cb{13}{used} \cb{2}{by} \cb{1}{smugglers} \cb{3}{,} \cb{3}{the} \cb{16}{I} \cb{3}{OM} \cb{1}{said} \cb{2}{.} \cb{17}{Sometimes} \cb{5}{the} \cb{24}{captains} \cb{5}{and} \cb{10}{crews} \cb{20}{abandon} \cb{3}{the} \cb{9}{ships} \cb{4}{,} \cb{6}{leaving} \cb{9}{passengers} \cb{4}{to} \cb{5}{fend} \cb{0}{for} \cb{0}{themselves} \cb{1}{.} \cb{14}{Last} \cb{2}{week} \cb{14}{:} \cb{24}{978} \cb{29}{migrants} \cb{9}{rescued} \cb{4}{in} \cb{13}{one} \cb{1}{day} \cb{5}{in} \cb{11}{Mediterranean} \cb{2}{Sea} \cb{22}{CNN} \cb{5}{'s} \cb{17}{Ralph} \cb{12}{Ellis} \cb{9}{contributed} \cb{0}{to} \cb{0}{this} \cb{1}{report} \cb{0}{.} }

\newcommand{\italygood}{
\cb{11}{R} \cb{7}{ome} \cb{13}{(} \cb{4}{CNN} \cb{0}{)} \cb{18}{Italy} \cb{4}{is} \cb{14}{coping} \cb{0}{with} \cb{2}{a} \cb{8}{rising} \cb{8}{wave} \cb{0}{of} \cb{9}{desperate} \cb{1}{migrants} \cb{5}{from} \cb{4}{Africa} \cb{2}{and} \cb{8}{Middle} \cb{0}{East} \cb{12}{hoping} \cb{0}{to} \cb{5}{make} \cb{2}{it} \cb{1}{to} \cb{1}{Europe} \cb{2}{.} \cb{13}{From} \cb{10}{Friday} \cb{3}{to} \cb{4}{Monday} \cb{1}{,} \cb{7}{a} \cb{1}{total} \cb{0}{of} \cb{6}{8} \cb{2}{,} \cb{13}{480} \cb{1}{migrants} \cb{3}{were} \cb{3}{rescued} \cb{6}{,} \cb{4}{according} \cb{0}{to} \cb{1}{the} \cb{1}{Italian} \cb{2}{coast} \cb{0}{guard} \cb{5}{,} \cb{3}{which} \cb{2}{said} \cb{2}{it} \cb{8}{received} \cb{17}{on} \cb{5}{Monday} \cb{19}{--} \cb{17}{alone} \cb{3}{--} \cb{30}{SOS} \cb{7}{calls} \cb{4}{from} \cb{13}{20} \cb{5}{boats} \cb{6}{in} \cb{12}{distress} \cb{1}{.} \cb{11}{On} \cb{4}{Tuesday} \cb{2}{,} \cb{4}{a} \cb{9}{spokesman} \cb{10}{with} \cb{8}{Save} \cb{0}{the} \cb{0}{Children} \cb{6}{told} \cb{5}{CNN} \cb{5}{the} \cb{5}{group} \cb{13}{fears} \cb{16}{400} \cb{6}{migrants} \cb{4}{could} \cb{2}{be} \cb{10}{missing} \cb{8}{,} \cb{9}{citing} \cb{15}{testimony} \cb{1}{from} \cb{8}{survivors} \cb{4}{who} \cb{2}{said} \cb{6}{their} \cb{5}{ship} \cb{15}{carrying} \cb{7}{550} \cb{0}{people} \cb{0}{caps} \cb{0}{ized} \cb{4}{in} \cb{1}{the} \cb{1}{Mediterranean} \cb{5}{Sea} \cb{12}{about} \cb{9}{80} \cb{2}{miles} \cb{4}{off} \cb{1}{the} \cb{2}{Libyan} \cb{0}{coast} \cb{2}{.} \cb{6}{The} \cb{4}{Italian} \cb{1}{coast} \cb{0}{guard} \cb{9}{,} \cb{5}{however} \cb{0}{,} \cb{10}{told} \cb{4}{CNN} \cb{1}{that} \cb{12}{while} \cb{2}{it} \cb{3}{is} \cb{10}{taking} \cb{6}{the} \cb{11}{report} \cb{0}{seriously} \cb{1}{,} \cb{1}{it} \cb{6}{cannot} \cb{2}{confirm} \cb{6}{such} \cb{1}{an} \cb{0}{incident} \cb{7}{and} \cb{5}{has} \cb{2}{not} \cb{4}{yet} \cb{10}{found} \cb{5}{evidence} \cb{13}{at} \cb{2}{sea} \cb{4}{to} \cb{8}{indicate} \cb{4}{a} \cb{9}{migrant} \cb{1}{boat} \cb{8}{carrying} \cb{14}{approximately} \cb{3}{550} \cb{12}{has} \cb{0}{caps} \cb{0}{ized} \cb{14}{with} \cb{21}{145} \cb{13}{rescued} \cb{1}{.} \cb{12}{An} \cb{13}{operation} \cb{9}{that} \cb{8}{included} \cb{7}{boats} \cb{4}{and} \cb{7}{planes} \cb{14}{did} \cb{0}{not} \cb{5}{find} \cb{0}{any} \cb{4}{survivors} \cb{6}{,} \cb{12}{nor} \cb{17}{bodies} \cb{4}{,} \cb{7}{nor} \cb{4}{any} \cb{5}{evidence} \cb{6}{to} \cb{5}{indicate} \cb{5}{a} \cb{14}{particular} \cb{3}{boat} \cb{9}{caps} \cb{0}{ized} \cb{4}{,} \cb{15}{Coast} \cb{0}{Guard} \cb{10}{official} \cb{13}{F} \cb{0}{ilipp} \cb{0}{o} \cb{10}{Mar} \cb{6}{ini} \cb{1}{said} \cb{1}{.} \cb{11}{There} \cb{3}{has} \cb{0}{been} \cb{1}{a} \cb{3}{recent} \cb{1}{up} \cb{0}{sur} \cb{0}{ge} \cb{0}{in} \cb{0}{migrant} \cb{0}{boats} \cb{0}{crossing} \cb{0}{the} \cb{0}{Mediterranean} \cb{10}{into} \cb{2}{Italy} \cb{6}{and} \cb{11}{an} \cb{2}{increase} \cb{0}{in} \cb{14}{resc} \cb{1}{ues} \cb{16}{performed} \cb{1}{by} \cb{3}{the} \cb{1}{Italian} \cb{2}{Coast} \cb{0}{Guard} \cb{9}{to} \cb{7}{aid} \cb{9}{migrant} \cb{1}{boats} \cb{3}{.} \cb{14}{Why} \cb{12}{migrants} \cb{2}{are} \cb{11}{dying} \cb{17}{trying} \cb{0}{to} \cb{1}{reach} \cb{6}{Italy} \cb{13}{According} \cb{0}{to} \cb{2}{the} \cb{4}{International} \cb{1}{Organization} \cb{0}{for} \cb{0}{Migration} \cb{3}{,} \cb{7}{Italy} \cb{8}{registered} \cb{3}{more} \cb{0}{than} \cb{2}{10} \cb{0}{,} \cb{0}{000} \cb{1}{migrants} \cb{2}{arriving} \cb{0}{in} \cb{5}{the} \cb{0}{first} \cb{0}{three} \cb{0}{months} \cb{0}{of} \cb{0}{2015} \cb{9}{,} \cb{6}{and} \cb{6}{about} \cb{5}{2} \cb{0}{,} \cb{1}{000} \cb{7}{were} \cb{3}{rescued} \cb{9}{at} \cb{0}{sea} \cb{7}{during} \cb{1}{the} \cb{1}{first} \cb{15}{weekend} \cb{0}{of} \cb{6}{April} \cb{7}{in} \cb{2}{the} \cb{14}{Channel} \cb{7}{of} \cb{0}{Sicily} \cb{3}{.} \cb{14}{Most} \cb{5}{migrants} \cb{21}{recorded} \cb{9}{this} \cb{5}{year} \cb{9}{come} \cb{0}{from} \cb{7}{countries} \cb{6}{in} \cb{10}{West} \cb{0}{Africa} \cb{11}{as} \cb{2}{well} \cb{0}{as} \cb{8}{Somalia} \cb{2}{and} \cb{7}{Syria} \cb{10}{,} \cb{5}{the} \cb{15}{I} \cb{6}{OM} \cb{2}{said} \cb{2}{.} \cb{13}{They} \cb{15}{use} \cb{14}{Libya} \cb{1}{as} \cb{1}{a} \cb{14}{country} \cb{2}{of} \cb{10}{transit} \cb{5}{.} \cb{12}{At} \cb{0}{least} \cb{16}{480} \cb{4}{migrants} \cb{2}{have} \cb{1}{died} \cb{9}{while} \cb{3}{crossing} \cb{0}{the} \cb{0}{Mediterranean} \cb{6}{since} \cb{2}{the} \cb{2}{beginning} \cb{0}{of} \cb{2}{the} \cb{0}{year} \cb{5}{,} \cb{17}{often} \cb{7}{because} \cb{1}{of} \cb{12}{bad} \cb{0}{weather} \cb{5}{and} \cb{5}{overcrowd} \cb{4}{ed} \cb{6}{vessels} \cb{14}{used} \cb{2}{by} \cb{2}{smugglers} \cb{6}{,} \cb{4}{the} \cb{13}{I} \cb{7}{OM} \cb{1}{said} \cb{2}{.} \cb{17}{Sometimes} \cb{4}{the} \cb{19}{captains} \cb{7}{and} \cb{5}{crews} \cb{16}{abandon} \cb{3}{the} \cb{5}{ships} \cb{5}{,} \cb{6}{leaving} \cb{5}{passengers} \cb{5}{to} \cb{5}{fend} \cb{0}{for} \cb{0}{themselves} \cb{3}{.} \cb{14}{Last} \cb{1}{week} \cb{15}{:} \cb{23}{978} \cb{3}{migrants} \cb{8}{rescued} \cb{4}{in} \cb{10}{one} \cb{1}{day} \cb{5}{in} \cb{8}{Mediterranean} \cb{3}{Sea} \cb{19}{CNN} \cb{6}{'s} \cb{17}{Ralph} \cb{13}{Ellis} \cb{11}{contributed} \cb{0}{to} \cb{0}{this} \cb{0}{report} \cb{1}{.} }

\newcommand{\italybad}{
\cb{11}{R} \cb{7}{ome} \cb{8}{(} \cb{13}{CNN} \cb{0}{)} \cb{15}{Italy} \cb{8}{is} \cb{11}{coping} \cb{0}{with} \cb{2}{a} \cb{10}{rising} \cb{8}{wave} \cb{0}{of} \cb{9}{desperate} \cb{1}{migrants} \cb{5}{from} \cb{5}{Africa} \cb{2}{and} \cb{8}{Middle} \cb{0}{East} \cb{13}{hoping} \cb{0}{to} \cb{5}{make} \cb{1}{it} \cb{0}{to} \cb{1}{Europe} \cb{1}{.} \cb{13}{From} \cb{8}{Friday} \cb{6}{to} \cb{4}{Monday} \cb{0}{,} \cb{8}{a} \cb{0}{total} \cb{0}{of} \cb{5}{8} \cb{1}{,} \cb{7}{480} \cb{0}{migrants} \cb{3}{were} \cb{1}{rescued} \cb{3}{,} \cb{6}{according} \cb{0}{to} \cb{1}{the} \cb{2}{Italian} \cb{1}{coast} \cb{0}{guard} \cb{3}{,} \cb{3}{which} \cb{2}{said} \cb{1}{it} \cb{8}{received} \cb{17}{on} \cb{5}{Monday} \cb{21}{--} \cb{17}{alone} \cb{2}{--} \cb{26}{SOS} \cb{5}{calls} \cb{3}{from} \cb{13}{20} \cb{8}{boats} \cb{5}{in} \cb{11}{distress} \cb{1}{.} \cb{11}{On} \cb{6}{Tuesday} \cb{0}{,} \cb{5}{a} \cb{7}{spokesman} \cb{10}{with} \cb{17}{Save} \cb{0}{the} \cb{0}{Children} \cb{5}{told} \cb{8}{CNN} \cb{5}{the} \cb{4}{group} \cb{16}{fears} \cb{16}{400} \cb{4}{migrants} \cb{3}{could} \cb{2}{be} \cb{7}{missing} \cb{5}{,} \cb{8}{citing} \cb{13}{testimony} \cb{1}{from} \cb{7}{survivors} \cb{4}{who} \cb{2}{said} \cb{6}{their} \cb{10}{ship} \cb{15}{carrying} \cb{15}{550} \cb{2}{people} \cb{8}{caps} \cb{0}{ized} \cb{3}{in} \cb{0}{the} \cb{1}{Mediterranean} \cb{2}{Sea} \cb{11}{about} \cb{9}{80} \cb{2}{miles} \cb{6}{off} \cb{1}{the} \cb{6}{Libyan} \cb{0}{coast} \cb{1}{.} \cb{6}{The} \cb{3}{Italian} \cb{0}{coast} \cb{0}{guard} \cb{8}{,} \cb{7}{however} \cb{0}{,} \cb{10}{told} \cb{7}{CNN} \cb{1}{that} \cb{10}{while} \cb{2}{it} \cb{3}{is} \cb{9}{taking} \cb{4}{the} \cb{13}{report} \cb{0}{seriously} \cb{0}{,} \cb{0}{it} \cb{7}{cannot} \cb{4}{confirm} \cb{13}{such} \cb{4}{an} \cb{5}{incident} \cb{8}{and} \cb{5}{has} \cb{3}{not} \cb{4}{yet} \cb{10}{found} \cb{5}{evidence} \cb{13}{at} \cb{4}{sea} \cb{4}{to} \cb{9}{indicate} \cb{5}{a} \cb{7}{migrant} \cb{4}{boat} \cb{5}{carrying} \cb{14}{approximately} \cb{11}{550} \cb{14}{has} \cb{5}{caps} \cb{0}{ized} \cb{13}{with} \cb{20}{145} \cb{14}{rescued} \cb{1}{.} \cb{12}{An} \cb{12}{operation} \cb{8}{that} \cb{10}{included} \cb{8}{boats} \cb{3}{and} \cb{8}{planes} \cb{14}{did} \cb{0}{not} \cb{5}{find} \cb{0}{any} \cb{3}{survivors} \cb{3}{,} \cb{10}{nor} \cb{17}{bodies} \cb{4}{,} \cb{7}{nor} \cb{4}{any} \cb{5}{evidence} \cb{6}{to} \cb{6}{indicate} \cb{6}{a} \cb{13}{particular} \cb{7}{boat} \cb{15}{caps} \cb{0}{ized} \cb{4}{,} \cb{18}{Coast} \cb{0}{Guard} \cb{9}{official} \cb{13}{F} \cb{0}{ilipp} \cb{0}{o} \cb{10}{Mar} \cb{6}{ini} \cb{1}{said} \cb{1}{.} \cb{11}{There} \cb{5}{has} \cb{0}{been} \cb{1}{a} \cb{9}{recent} \cb{15}{up} \cb{0}{sur} \cb{0}{ge} \cb{0}{in} \cb{6}{migrant} \cb{6}{boats} \cb{8}{crossing} \cb{0}{the} \cb{1}{Mediterranean} \cb{8}{into} \cb{1}{Italy} \cb{5}{and} \cb{13}{an} \cb{2}{increase} \cb{0}{in} \cb{14}{resc} \cb{0}{ues} \cb{18}{performed} \cb{1}{by} \cb{2}{the} \cb{1}{Italian} \cb{4}{Coast} \cb{0}{Guard} \cb{10}{to} \cb{7}{aid} \cb{9}{migrant} \cb{2}{boats} \cb{2}{.} \cb{14}{Why} \cb{7}{migrants} \cb{2}{are} \cb{4}{dying} \cb{17}{trying} \cb{0}{to} \cb{1}{reach} \cb{2}{Italy} \cb{12}{According} \cb{0}{to} \cb{1}{the} \cb{8}{International} \cb{1}{Organization} \cb{0}{for} \cb{0}{Migration} \cb{1}{,} \cb{5}{Italy} \cb{14}{registered} \cb{6}{more} \cb{0}{than} \cb{8}{10} \cb{0}{,} \cb{0}{000} \cb{4}{migrants} \cb{9}{arriving} \cb{1}{in} \cb{2}{the} \cb{5}{first} \cb{4}{three} \cb{0}{months} \cb{0}{of} \cb{3}{2015} \cb{1}{,} \cb{5}{and} \cb{7}{about} \cb{5}{2} \cb{0}{,} \cb{2}{000} \cb{5}{were} \cb{5}{rescued} \cb{9}{at} \cb{1}{sea} \cb{8}{during} \cb{1}{the} \cb{2}{first} \cb{16}{weekend} \cb{0}{of} \cb{7}{April} \cb{9}{in} \cb{2}{the} \cb{14}{Channel} \cb{6}{of} \cb{0}{Sicily} \cb{0}{.} \cb{14}{Most} \cb{4}{migrants} \cb{20}{recorded} \cb{8}{this} \cb{3}{year} \cb{11}{come} \cb{0}{from} \cb{8}{countries} \cb{6}{in} \cb{10}{West} \cb{0}{Africa} \cb{10}{as} \cb{3}{well} \cb{0}{as} \cb{7}{Somalia} \cb{2}{and} \cb{6}{Syria} \cb{3}{,} \cb{4}{the} \cb{1}{I} \cb{0}{OM} \cb{3}{said} \cb{0}{.} \cb{13}{They} \cb{15}{use} \cb{18}{Libya} \cb{1}{as} \cb{1}{a} \cb{13}{country} \cb{1}{of} \cb{10}{transit} \cb{4}{.} \cb{12}{At} \cb{3}{least} \cb{15}{480} \cb{1}{migrants} \cb{2}{have} \cb{1}{died} \cb{8}{while} \cb{4}{crossing} \cb{0}{the} \cb{1}{Mediterranean} \cb{6}{since} \cb{4}{the} \cb{3}{beginning} \cb{0}{of} \cb{2}{the} \cb{1}{year} \cb{1}{,} \cb{16}{often} \cb{6}{because} \cb{1}{of} \cb{12}{bad} \cb{0}{weather} \cb{5}{and} \cb{6}{overcrowd} \cb{4}{ed} \cb{7}{vessels} \cb{14}{used} \cb{2}{by} \cb{2}{smugglers} \cb{4}{,} \cb{3}{the} \cb{1}{I} \cb{0}{OM} \cb{2}{said} \cb{1}{.} \cb{17}{Sometimes} \cb{3}{the} \cb{21}{captains} \cb{6}{and} \cb{5}{crews} \cb{18}{abandon} \cb{3}{the} \cb{6}{ships} \cb{3}{,} \cb{7}{leaving} \cb{7}{passengers} \cb{4}{to} \cb{5}{fend} \cb{0}{for} \cb{0}{themselves} \cb{2}{.} \cb{14}{Last} \cb{3}{week} \cb{10}{:} \cb{22}{978} \cb{2}{migrants} \cb{6}{rescued} \cb{4}{in} \cb{11}{one} \cb{1}{day} \cb{6}{in} \cb{8}{Mediterranean} \cb{2}{Sea} \cb{19}{CNN} \cb{6}{'s} \cb{17}{Ralph} \cb{13}{Ellis} \cb{11}{contributed} \cb{0}{to} \cb{0}{this} \cb{0}{report} \cb{1}{.} }

\newcommand{\italyfull}{
\cb{11}{R} \cb{1}{ome} \cb{5}{(} \cb{0}{CNN} \cb{0}{)} \cb{0}{Italy} \cb{0}{is} \cb{0}{coping} \cb{0}{with} \cb{0}{a} \cb{0}{rising} \cb{0}{wave} \cb{0}{of} \cb{0}{desperate} \cb{0}{migrants} \cb{0}{from} \cb{0}{Africa} \cb{0}{and} \cb{0}{Middle} \cb{0}{East} \cb{0}{hoping} \cb{0}{to} \cb{0}{make} \cb{0}{it} \cb{0}{to} \cb{0}{Europe} \cb{0}{.} \cb{13}{From} \cb{2}{Friday} \cb{0}{to} \cb{0}{Monday} \cb{0}{,} \cb{0}{a} \cb{0}{total} \cb{0}{of} \cb{0}{8} \cb{0}{,} \cb{0}{480} \cb{0}{migrants} \cb{0}{were} \cb{0}{rescued} \cb{0}{,} \cb{0}{according} \cb{0}{to} \cb{0}{the} \cb{0}{Italian} \cb{0}{coast} \cb{0}{guard} \cb{0}{,} \cb{0}{which} \cb{0}{said} \cb{0}{it} \cb{0}{received} \cb{0}{on} \cb{0}{Monday} \cb{0}{--} \cb{0}{alone} \cb{0}{--} \cb{1}{SOS} \cb{0}{calls} \cb{0}{from} \cb{0}{20} \cb{0}{boats} \cb{0}{in} \cb{0}{distress} \cb{0}{.} \cb{11}{On} \cb{1}{Tuesday} \cb{0}{,} \cb{0}{a} \cb{0}{spokesman} \cb{0}{with} \cb{0}{Save} \cb{0}{the} \cb{0}{Children} \cb{0}{told} \cb{0}{CNN} \cb{0}{the} \cb{0}{group} \cb{0}{fears} \cb{0}{400} \cb{0}{migrants} \cb{0}{could} \cb{0}{be} \cb{0}{missing} \cb{0}{,} \cb{0}{citing} \cb{0}{testimony} \cb{0}{from} \cb{0}{survivors} \cb{0}{who} \cb{0}{said} \cb{0}{their} \cb{0}{ship} \cb{0}{carrying} \cb{0}{550} \cb{0}{people} \cb{0}{caps} \cb{0}{ized} \cb{0}{in} \cb{0}{the} \cb{0}{Mediterranean} \cb{0}{Sea} \cb{0}{about} \cb{0}{80} \cb{0}{miles} \cb{0}{off} \cb{0}{the} \cb{0}{Libyan} \cb{0}{coast} \cb{0}{.} \cb{6}{The} \cb{3}{Italian} \cb{0}{coast} \cb{0}{guard} \cb{3}{,} \cb{0}{however} \cb{0}{,} \cb{0}{told} \cb{0}{CNN} \cb{0}{that} \cb{0}{while} \cb{0}{it} \cb{0}{is} \cb{0}{taking} \cb{0}{the} \cb{0}{report} \cb{0}{seriously} \cb{0}{,} \cb{0}{it} \cb{0}{cannot} \cb{0}{confirm} \cb{0}{such} \cb{0}{an} \cb{0}{incident} \cb{0}{and} \cb{0}{has} \cb{0}{not} \cb{0}{yet} \cb{0}{found} \cb{0}{evidence} \cb{0}{at} \cb{0}{sea} \cb{0}{to} \cb{0}{indicate} \cb{0}{a} \cb{0}{migrant} \cb{0}{boat} \cb{0}{carrying} \cb{0}{approximately} \cb{0}{550} \cb{0}{has} \cb{0}{caps} \cb{0}{ized} \cb{0}{with} \cb{0}{145} \cb{0}{rescued} \cb{0}{.} \cb{12}{An} \cb{5}{operation} \cb{0}{that} \cb{0}{included} \cb{0}{boats} \cb{0}{and} \cb{0}{planes} \cb{0}{did} \cb{0}{not} \cb{0}{find} \cb{0}{any} \cb{0}{survivors} \cb{0}{,} \cb{0}{nor} \cb{0}{bodies} \cb{0}{,} \cb{0}{nor} \cb{0}{any} \cb{0}{evidence} \cb{0}{to} \cb{0}{indicate} \cb{0}{a} \cb{0}{particular} \cb{0}{boat} \cb{0}{caps} \cb{0}{ized} \cb{0}{,} \cb{1}{Coast} \cb{0}{Guard} \cb{0}{official} \cb{2}{F} \cb{0}{ilipp} \cb{0}{o} \cb{0}{Mar} \cb{0}{ini} \cb{0}{said} \cb{0}{.} \cb{11}{There} \cb{3}{has} \cb{0}{been} \cb{0}{a} \cb{0}{recent} \cb{0}{up} \cb{0}{sur} \cb{0}{ge} \cb{0}{in} \cb{0}{migrant} \cb{0}{boats} \cb{0}{crossing} \cb{0}{the} \cb{0}{Mediterranean} \cb{0}{into} \cb{0}{Italy} \cb{0}{and} \cb{0}{an} \cb{0}{increase} \cb{0}{in} \cb{0}{resc} \cb{0}{ues} \cb{0}{performed} \cb{0}{by} \cb{0}{the} \cb{0}{Italian} \cb{0}{Coast} \cb{0}{Guard} \cb{0}{to} \cb{0}{aid} \cb{0}{migrant} \cb{0}{boats} \cb{0}{.} \cb{14}{Why} \cb{0}{migrants} \cb{0}{are} \cb{0}{dying} \cb{0}{trying} \cb{0}{to} \cb{0}{reach} \cb{0}{Italy} \cb{3}{According} \cb{0}{to} \cb{0}{the} \cb{0}{International} \cb{0}{Organization} \cb{0}{for} \cb{0}{Migration} \cb{0}{,} \cb{0}{Italy} \cb{0}{registered} \cb{0}{more} \cb{0}{than} \cb{0}{10} \cb{0}{,} \cb{0}{000} \cb{0}{migrants} \cb{0}{arriving} \cb{0}{in} \cb{0}{the} \cb{0}{first} \cb{0}{three} \cb{0}{months} \cb{0}{of} \cb{0}{2015} \cb{0}{,} \cb{0}{and} \cb{0}{about} \cb{0}{2} \cb{0}{,} \cb{0}{000} \cb{0}{were} \cb{0}{rescued} \cb{0}{at} \cb{0}{sea} \cb{0}{during} \cb{0}{the} \cb{0}{first} \cb{0}{weekend} \cb{0}{of} \cb{0}{April} \cb{0}{in} \cb{0}{the} \cb{0}{Channel} \cb{0}{of} \cb{0}{Sicily} \cb{0}{.} \cb{14}{Most} \cb{1}{migrants} \cb{9}{recorded} \cb{0}{this} \cb{0}{year} \cb{0}{come} \cb{0}{from} \cb{0}{countries} \cb{0}{in} \cb{0}{West} \cb{0}{Africa} \cb{0}{as} \cb{0}{well} \cb{0}{as} \cb{0}{Somalia} \cb{0}{and} \cb{0}{Syria} \cb{0}{,} \cb{0}{the} \cb{1}{I} \cb{0}{OM} \cb{0}{said} \cb{0}{.} \cb{13}{They} \cb{3}{use} \cb{3}{Libya} \cb{0}{as} \cb{0}{a} \cb{0}{country} \cb{0}{of} \cb{0}{transit} \cb{0}{.} \cb{12}{At} \cb{0}{least} \cb{4}{480} \cb{0}{migrants} \cb{0}{have} \cb{0}{died} \cb{0}{while} \cb{0}{crossing} \cb{0}{the} \cb{0}{Mediterranean} \cb{0}{since} \cb{0}{the} \cb{0}{beginning} \cb{0}{of} \cb{0}{the} \cb{0}{year} \cb{0}{,} \cb{0}{often} \cb{0}{because} \cb{0}{of} \cb{0}{bad} \cb{0}{weather} \cb{0}{and} \cb{0}{overcrowd} \cb{0}{ed} \cb{0}{vessels} \cb{0}{used} \cb{0}{by} \cb{0}{smugglers} \cb{0}{,} \cb{0}{the} \cb{1}{I} \cb{0}{OM} \cb{0}{said} \cb{0}{.} \cb{17}{Sometimes} \cb{1}{the} \cb{2}{captains} \cb{0}{and} \cb{0}{crews} \cb{0}{abandon} \cb{0}{the} \cb{0}{ships} \cb{0}{,} \cb{0}{leaving} \cb{0}{passengers} \cb{0}{to} \cb{0}{fend} \cb{0}{for} \cb{0}{themselves} \cb{0}{.} \cb{14}{Last} \cb{0}{week} \cb{0}{:} \cb{2}{978} \cb{0}{migrants} \cb{0}{rescued} \cb{0}{in} \cb{0}{one} \cb{0}{day} \cb{0}{in} \cb{0}{Mediterranean} \cb{0}{Sea} \cb{7}{CNN} \cb{0}{'s} \cb{0}{Ralph} \cb{0}{Ellis} \cb{0}{contributed} \cb{0}{to} \cb{0}{this} \cb{0}{report} \cb{0}{.} }

\newcommand{\italyhigh}{A Save the Children spokesman says a ship carrying 550 people capsized off the Libyan coast . The Italian coast guard says it can not confirm such an incident . There has been a recent upsurge in migrant boats crossing the Mediterranean . Italy registered more than 10,000 migrants arriving in first three months of 2015 .}

\newcommand{\italylow}{Italy's coast guard says it has rescued 8,480 migrants since Friday, more than 1,000 of whom are believed to have died TIMELINE: Migrant crisis in the Mediterranean The IOM says that of the 8,480 migrants rescued by Italian coast guard vessels, 5,943 were women and children, and 1,852 were men. The IOM says that of the 1,852 women and children, 781 were from Eritrea}

\newcommand{\italyfigure}{
\onecolumn
\small
\setlength\fboxsep{1pt}
\begin{longtable}{@{}p{0.48\linewidth}p{0.48\linewidth}@{}}

\multicolumn{2}{c}{\cb{0}{Continued} \cb{0}{on} \cb{0}{the} \cb{0}{next} \cb{0}{page}}\\
\endfoot
\endlastfoot

\multicolumn{2}{c}{\cb{0}{Continued} \cb{0}{from} \cb{0}{the} \cb{0}{previous} \cb{0}{page}}\\
\endhead
\endfirsthead

$I(\mathcal{D}) = 1352$ & $I(\mathcal{D}|\mathcal{D})=118$ \\
\noalign{\vskip 2mm}
\fontdimen2\font=0pt
\italybase & \italyfull \\
\noalign{\vskip 1mm}
\midrule 
\midrule
\noalign{\vskip 1mm}
\fontdimen2\font=\origiwspc
$I(\mathcal{D}|\mathcal{S})=1137$ for this high quality summary: & $I(\mathcal{D}|\mathcal{S})=1168$ for this low quality summary: \\
\noalign{\vskip 1mm}
\italyhigh & \italylow \\
\noalign{\vskip 2mm}
\fontdimen2\font=0pt
\italygood & \italybad \\
\caption{A comparison of token-wise information content within a document as estimated by GPT-2 in 4 scenarios: the document on its own, the document given the document, the document given a high quality summary, and the document given a low quality summary. Tokens with a darker background color have more information.}
\label{fig:italy-viz}
\end{longtable}
\twocolumn
}

%% file: appendix.tex
\section{Experimental Setup}
\lmcomputetable
In our experiments, we used the Transformers library\footnote{https://github.com/huggingface/transformers} \citep{Wolf2019HuggingFacesTS} for implementations of GPT-2, GPT, XLNet, and Transformer-XL. Sentence tokenization was performed with the NLTK Punkt sentence tokenizer\footnote{https://www.nltk.org/api/nltk.tokenize.html}. We computed our metrics using a single NVIDIA V100 GPU.  Our code will be published on GitHub and linked in this paper after blind review.

Table \ref{tab:lm-compute} shows information about each language model we used in our experiments: the HuggingFace model hub\footnote{https://huggingface.co/models} name for the model, number of parameters in the model, and the number of seconds it takes to compute the Shannon Score for a given document-summary pair on average for SummEval when using the model. Our implementation did not include any batching since we had no need for it, but if we batched together document sentences for model inference we could greatly improve our runtime.

\section{Full Doc Information Visualizations}
Tables \ref{fig:elvis-viz}, \ref{fig:italy-viz}, and \ref{fig:cycling-viz} show full document information visualizations for 3 different CNN/DailyMail documents. Each of these documents were paired with 2 SummEval system summaries: one that humans judged to be of high quality, and one that humans judged to be of low quality. The visualizations show the information content of each token in the document as estimated by GPT-2 in 4 scenarios: $I(\mathcal{D})$ (the document on its own), $I(\mathcal{D}|\mathcal{D})$ (the document given the document), $I(\mathcal{D}|\mathcal{S})$ (the document given a summary) for the high quality summary, and $I(\mathcal{D}|\mathcal{S})$ for the low quality summary. A darker background color denotes higher information according to the model.

\elvisfigure
\italyfigure
\cyclingfigure